\documentclass[preprint, 12pt, authoryear]{elsarticle}

\usepackage{amssymb}
\usepackage{amsmath}

\usepackage{booktabs}
\usepackage{multirow}
\usepackage{tikz}
\usepackage{subcaption}
\usepackage{pifont}
\usepackage{xspace}
\usepackage{fontawesome5}
\usepackage{wrapfig}
\usepackage[most]{tcolorbox}

\newcommand{\cmark}{\ding{51}}
\newcommand{\xmark}{\ding{55}}

\newcommand{\datar}{$\mathcal{D}_r$\xspace}
\newcommand{\dataf}{$\mathcal{D}_f$\xspace}

\newcommand{\thetar}{$\theta_r$\xspace}
\newcommand{\thetao}{$\theta_o$\xspace}
\newcommand{\thetau}{$\theta_u$\xspace}
\newcommand{\thetaulast}{$\theta_u^{last}$\xspace}

\newcommand{\eg}{\textit{e.g.}}
\newcommand{\ie}{\textit{i.e.}}
\newcommand{\vs}{\textit{vs}.\xspace}
\newcommand{\qq}[1]{``#1''}

\journal{Engineering Applications of Artificial Intelligence}

\begin{document}

\begin{frontmatter}



\title{Are We Truly Forgetting?\\A Critical Re-examination of Machine Unlearning Evaluation Protocols} 



\author[inst1]{Yongwoo Kim\fnref{eq}}
\author[inst2]{Sungmin Cha\fnref{eq}}
\author[inst1]{Donghyun Kim\fnref{ca}}\ead{d_kim@korea.ac.kr}

\fntext[eq]{First authors}
\fntext[ca]{Corresponding author}


\address[inst1]{Department of Artificial Intelligence, Korea University, South Korea}
\address[inst2]{Computer Science Dept. at Courant Institute of Mathemtical Sciences, New York University, United States}

\begin{abstract}
Machine unlearning is a process to remove specific data points from a trained model while maintaining the performance on the retain data, addressing privacy or legal requirements. Despite its importance, existing unlearning evaluations tend to focus on logit-based metrics under small-scale scenarios. We observe that this could lead to a false sense of security in unlearning approaches under real-world scenarios. In this paper, we conduct a comprehensive evaluation that employs representation-based evaluations of the unlearned model under large-scale scenarios to verify whether the unlearning approaches truly eliminate the targeted data from the model's representation perspective. Our analysis reveals that current state-of-the-art unlearning approaches either completely degrade the representational quality of the unlearned model or merely modify the classifier, thereby achieving superior logit-based performance while maintaining representational similarity to the original model. Furthermore, we introduce a novel unlearning evaluation scenario in which the forgetting classes exhibit semantic similarity to downstream task classes, necessitating that feature representations diverge significantly from those of the original model, thus enabling a more thorough evaluation from a representation perspective. We hope our benchmark will serve as a standardized protocol for evaluating unlearning algorithms under realistic conditions.
\end{abstract}




\end{frontmatter}



\section{Introduction}
\label{sec:intro}
Neural networks have driven transformative advancements across a multitude of domains, such as image classification~\citep{deng2009imagenet, he2016deep, liu2022convnext}. A key contributor to this progress has been the substantial enhancement of the model's capabilities, facilitated by the training of sophisticated models on vast datasets~\citep{bengio2021deep}. However, these successes have also raised significant privacy concerns about these datasets, particularly related to the unchecked use of sensitive data~\citep{carlini2021extracting}. As privacy rights, including the `right to be forgotten' (\eg, the GDPR legislation~\citep{cao2015towards, MANTELERO2013229}), have gained increasing recognition, the need for effective solutions has become urgent. As a result, this has led to the emergence of machine unlearning as a critical research area~\citep{nguyen2022survey}.

\begin{figure*}[t]
    \centering
    \begin{subfigure}[b]{0.45\textwidth}
        \centering
        \includegraphics[width=\textwidth]{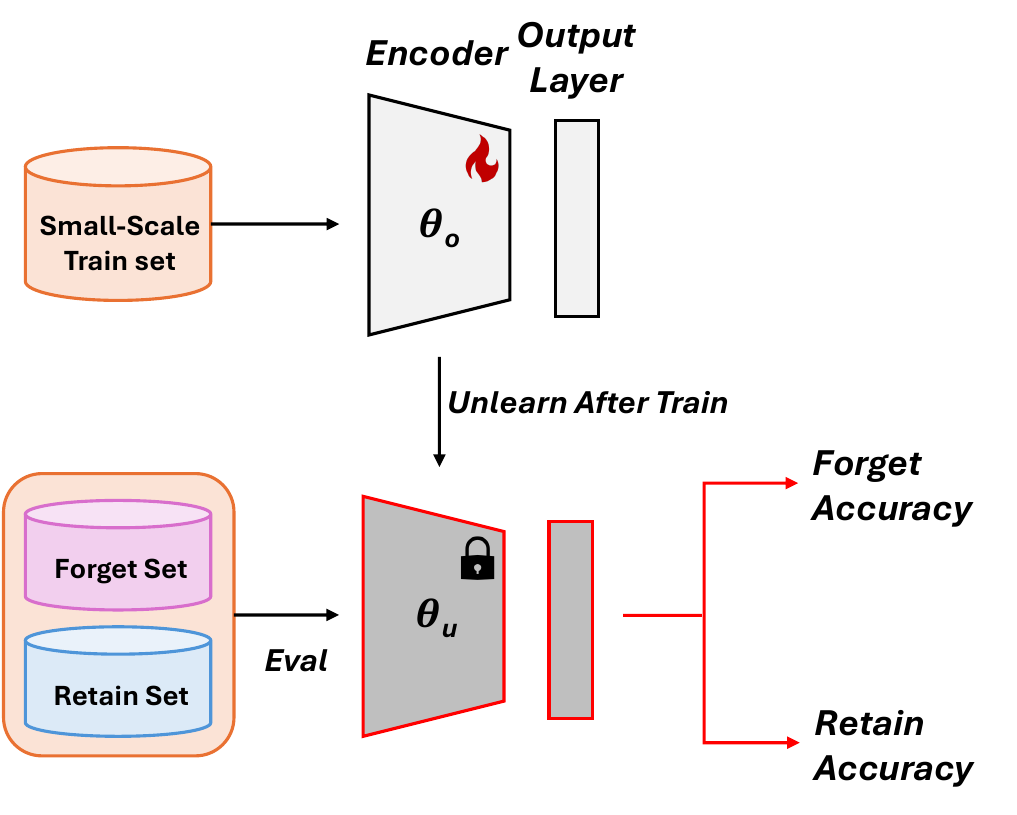}
        \caption{Traditional evaluation framework.}
        \label{fig:traditional_eval_framework}
    \end{subfigure}
    \hspace{4mm}
    \begin{subfigure}[b]{0.45\textwidth}
        \centering
        \includegraphics[width=\textwidth]{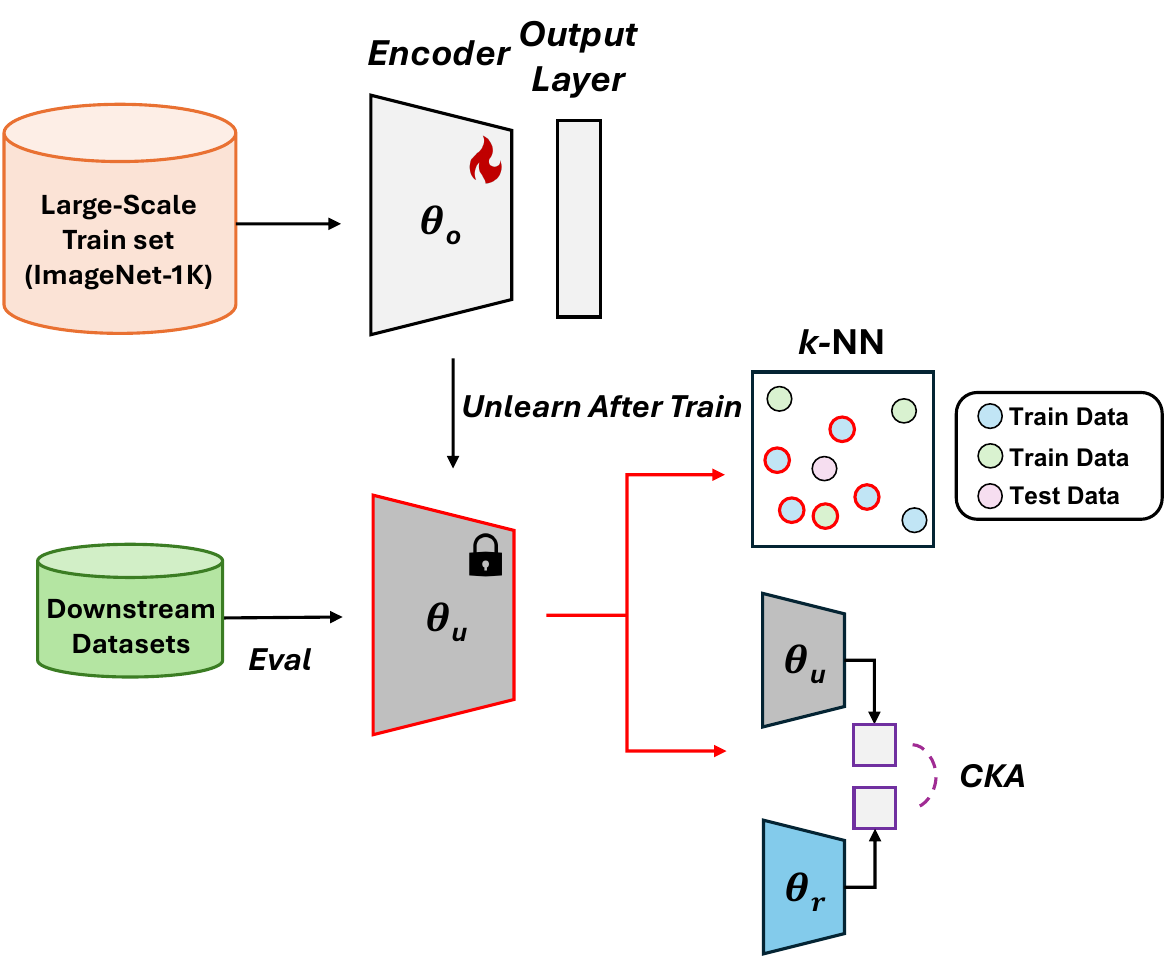}
        \caption{Our proposed evaluation framework.}
        \label{fig:downstream_eval_framework}
    \end{subfigure}
    \caption{A comparison of (a) the traditional evaluation framework and (b) our proposed evaluation framework. Note that \thetau and \thetar refer to the unlearned model and the model trained on the retain set, respectively. Traditional unlearning evaluation methods primarily focus on analyzing the unlearned model's output logits to assess the effectiveness of unlearning under small-scale scenarios such as CIFAR-10. In contrast, our framework introduces additional evaluation factors by examining the unlearned model's feature representation similarity with \thetar in terms of transferability and representational similarity under the large-scale unlearning scenario, such as using ImageNet-1K.}
    \vspace{-4mm}
    \label{fig:evaluation_framework}
\end{figure*}

The primary objective of machine unlearning is to effectively eliminate the influence of specific data from a pre-trained model while preserving the knowledge derived from other data~\citep{nguyen2022survey}. The most straightforward approach, known as \textit{exact unlearning}, involves retraining the model from scratch using a refined dataset~\citep{Mahadevan2021CertifiableMU} (hereafter referred to as the \textit{``retrained model''}).  This approach is regarded as the gold standard as it guarantees the complete elimination of the targeted data. However, it is computationally expensive, requiring retraining for each unlearning request. To overcome these limitations, several \textit{approximate unlearning} algorithms have been proposed. They offer more efficient alternatives while maintaining the goal of machine unlearning. In image classification tasks, approximate unlearning algorithms are typically evaluated based on their ability to produce models that achieve classification accuracy similar to that of the retrained model on the given dataset (\eg, retain and forget sets)~\citep{nguyen2022survey}, as illustrated on the left side of Fig.~\ref{fig:evaluation_framework}. Following this evaluation method, several techniques have induced unlearning through methods such as Gradient Ascent \citep{thudi2022unrolling}, Random Labeling \citep{golatkar2020eternal}, and metric learning~\citep{bonato2024retain}, while mitigating over-forgetting by incorporating additional regularization terms~\citep{kurmanji2023towards, fan2024salun, cha2024learning, bonato2024retain}.

Despite recent advancements, current unlearning evaluations remain narrow. They primarily focus on logit-based metrics (\eg, accuracy, Membership Inference Attack) and small-scale tasks, such as unlearning one class or a few classes in CIFAR-10, as shown in Fig.~\ref{fig:evaluation_framework} and Table~\ref{tab:revision_1}. Such narrow evaluation scopes raise significant concerns about the scalability of these methods to larger datasets and real-world scenarios. Privacy considerations assume heightened significance in larger model architectures (\eg, foundation models), as these systems are trained on extensive datasets, frequently sourced from web-crawled images, and are subsequently employed across diverse downstream applications. This limitation overlooks the broader implications of unlearning algorithms on the model's foundational component (\ie, encoder), potentially leading to incomplete assessments of its effectiveness.

\begin{table*}[t]
\centering
\caption{Comparison of evaluation protocols in recent machine unlearning literature. While current unlearning methods are evaluated on small-scale datasets for unlearning, our work utilizes ImageNet-1K for unlearning, which consists of about 1.28 million images.}
\resizebox{\textwidth}{!}{
\begin{tabular}{l l l c c}
\toprule
\textbf{Work} & \textbf{Backbone Model} & \textbf{Unlearning Dataset} & \textbf{Unlearning Classes} & \textbf{Downstream Dataset} \\
\midrule
\cite{thudi2022unrolling} & ResNet-18 & CIFAR-10/100 (\#60K) & - & \xmark \\
\cite{golatkar2020eternal} & ResNet-18 & MNIST (\#70K), etc. & 1 & \xmark \\
\cite{chen2024unsc} & ResNet-18, VGG-11, AllCNN & SVHN (\#99K), etc. & 1-10 & \xmark \\
\cite{fan2024salun} & ResNet-18 & CIFAR-10 (\#60K) & 1 & \xmark \\
\cite{cotogni2023duck} & ResNet-18, VGG & Tiny ImageNet (\#110K), etc. & 1 & \xmark \\
\cite{zhang2024contrastive} & ResNet-18/34/50/101 & SVHN (\#99K), etc. & 1 & \xmark \\
\cite{kurmanji2023towards} & ResNet-18, AllCNN & CIFAR-10 (\#60K), etc. & 1 & \xmark \\
\cite{bonato2024retain} & ResNet-18/34/50, AllCNN, ViT-B16 & Tiny ImageNet (\#110K), etc. & 1 & \xmark \\
\midrule
\textbf{Ours} & \textbf{ResNet-50, Swin-T, ConvNeXt} & \textbf{ImageNet-1K} (\#1.28M) & \textbf{100-300} & \checkmark \\
\bottomrule
\end{tabular}
}
\label{tab:revision_1}
\end{table*}

\begin{figure*}[t]
    \centering
    \includegraphics[width=0.9\linewidth]{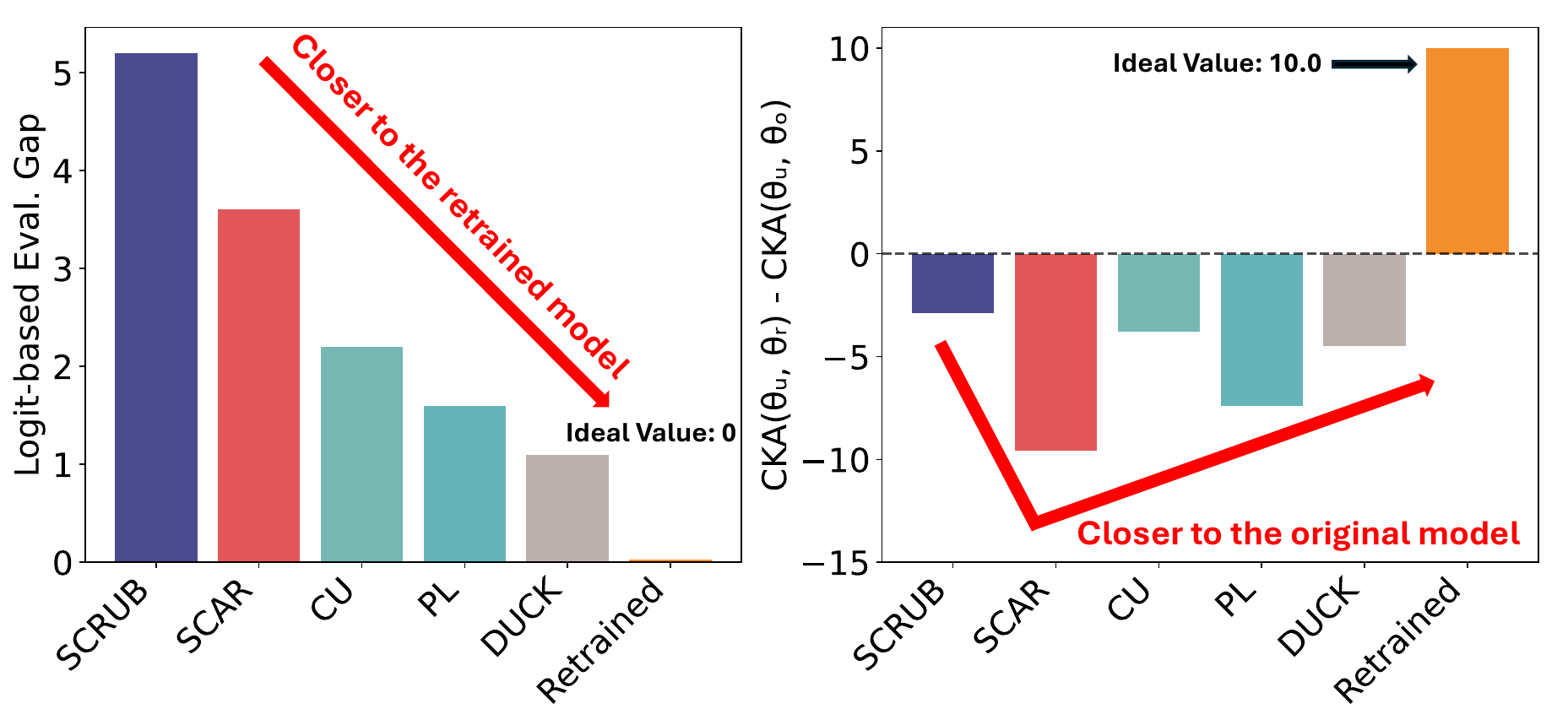}
    \vspace{-2mm}
    \caption{Performance comparison between logit-based (left) and representation-based evaluation (right) reveals contrasting findings.}
    \label{fig:logitvsrep}
    \vspace{-2mm}
\end{figure*}

In this paper, we propose a holistic unlearning evaluation framework that addresses the limitations of small-scale, logit-based evaluations. We incorporate assessments of unlearned models from the perspective of feature representations in large-scale settings. As illustrated in Fig.~\ref{fig:evaluation_framework}, our framework evaluates the representations of the unlearned models using Centered Kernel Alignment (CKA)~\citep{kornblith2019similarity} to compare the feature representations of the unlearned models with those of the retrained model (\thetar). Additionally, we apply $k$-Nearest Neighbors ($k$-NN) analysis on the feature space across various downstream datasets to assess the quality of the unlearned model's feature representations and compare the results with those of the retrained model.
Through these evaluation methods, we re-evaluate multiple state-of-the-art unlearning algorithms in large-scale scenarios. Fig.~\ref{fig:logitvsrep} illustrates key findings from experiments on unlearning 100 random classes in ImageNet-1K using ResNet-50. While logit-based evaluation (\eg, classification accuracy) demonstrates successful unlearning by showing a decreasing gap between the unlearned models (\thetau) and the retrained models (\thetar) (Fig.~\ref{fig:logitvsrep}, left), a representation-based evaluation using CKA reveals a critical limitation.
If unlearning is truly effective, the unlearned model should be indistinguishable from the retrained model, with the ideal value being 10. However, as shown in Fig.~\ref{fig:logitvsrep} (right), all unlearned models exhibit greater similarity to the original model than to the retrained model. Furthermore, the order of effectiveness of these algorithms varies depending on whether a logit-based or representation-based evaluation is used. These findings highlight a fundamental issue: existing unlearning methods fail to sufficiently erase learned information at the representation level, and they were not evaluated from multifaceted perspectives. These findings reinforce the need for representation-based evaluation to rigorously assess whether unlearning truly eliminates the influence of the targeted data.

Furthermore, random class forgetting has been widely adopted as a standard evaluation scenario in unlearning studies~\citep{cha2024learning, fan2024salun, bonato2024retain, kurmanji2023towards}. However, we highlight a critical limitation of this approach: the forgetting process may be inadequately assessed, as similar representational knowledge can persist across multiple alternative classes, making it difficult to verify whether the targeted data has truly been unlearned. To address this issue, we propose a novel unlearning evaluation paradigm, \textit{Top Class-wise Forgetting}, in which the classes selected for unlearning are semantically similar to those in downstream tasks. This setup requires a substantial divergence of feature representations from the original model, offering a more rigorous assessment of the unlearning algorithm's effectiveness. We then evaluate existing unlearning methods using both logit-based and representation-based metrics across diverse downstream datasets. Our goal is to establish a rigorous benchmark for the community, enabling a holistic evaluation of future methods.
In summary, our contributions are as follows:
\begin{itemize}
    \item We identify fundamental limitations in current evaluations, which rely on logit-based metrics under small-scale scenarios.
    \item To address this, we propose a unified benchmark that incorporates representation-based metrics and reveals that existing methods fail to alter internal representations. Furthermore, we conduct various experiments in large-scale settings (\textit{i.e.}, evaluations on ImageNet-1K using ResNet-50, Swin-T, and ConvNeXt) to determine whether current baselines can truly unlearn the model's knowledge.
    \item We further introduce Top Class-wise Forgetting, a more rigorous evaluation scenario that emphasizes semantic overlap with downstream tasks and better reflects real-world unlearning challenges.
\end{itemize}

\section{Related Works}
\label{sec:relatedworks}
\noindent\textbf{Machine unlearning in image classification.}
Early research on machine unlearning in neural networks has primarily focused on image classification, leading to the development of various unlearning algorithms with distinct strategies~\citep{nguyen2022survey}. A widely used approach, Gradient Ascent (GA), maximizes the loss on the forget set to erase its influence, making it one of the most fundamental yet widely adopted techniques in recent studies~\citep{golatkar2020eternal, tarun2023fast, graves2020amnesiacmachinelearning, recoverable, cha2024learning}. Similarly, Fine-tuning achieves unlearning by inducing catastrophic forgetting on the forget set, updating the model solely on the retain set~\citep{golatkar2020eternal,warnecke2021machine}. More sophisticated unlearning techniques have been developed: SCRUB~\citep{kurmanji2023towards} employs a knowledge distillation framework to selectively forget knowledge from the forget set while retaining other knowledge. L2UL~\citep{cha2024learning} enhances unlearning by incorporating two regularization terms that mitigate forgetting at both the feature and weight levels when combined with GA. SalUn~\citep{fan2024salun} modifies the most influential parameters through weight saliency, enabling selective unlearning. SCAR~\citep{bonato2024retain} leverages metric learning alongside a distillation-based strategy, aligning the forget set representations to the nearest incorrect class using the Mahalanobis distance. Beyond image classification, recent works have begun to address the complexities of unlearning in generative models, specifically those pre-trained on large-scale datasets such as ImageNet-1K~\citep{li2024machine, feng2025controllable}.

\noindent\textbf{Evaluation and analysis of representations.}
Transfer learning has become a foundational technique in machine learning, facilitating the adaptation of pre-trained models to diverse downstream tasks~\citep{pan2010transfer}. Beyond its practical applications, it is widely used to assess the quality of the learned representations.
For instance, it has been employed to compare the learned representations of different model architectures~\citep{kornblith2019better} and algorithms across various domains, such as self-supervised learning~\citep{chen2020simple, he2020momentum} and continual learning~\citep{cha2022towards}.
This evaluation is typically performed using metrics such as $k$-Nearest Neighbors ($k$-NN), which measures how well the learned representations generalize in various downstream tasks. Centered Kernel Alignment (CKA)~\citep{kornblith2019similarity} has been widely used to compare feature representations across models, providing insights into their structural similarities and differences. Furthermore, t-SNE~\citep{JMLR:v9:vandermaaten08a} has been extensively utilized to visualize feature representations, offering an intuitive means to compare and analyze how different models organize learned representations.

In contrast to prior work, we revisit the foundations of unlearning evaluation by exposing key limitations in conventional protocols and proposing a unified, representation-aware benchmark that enables a more realistic and rigorous assessment.

\noindent\textbf{Traditional evaluation framework in machine unlearning.}
In the literature on machine unlearning, particularly for image classification tasks, logit-based evaluation metrics have primarily been used to quantify the effectiveness of unlearning algorithms on small-scale scenarios \citep{golatkar2020eternal, thudi2022unrolling, cotogni2023duck, kurmanji2023towards, chen2023boundary, jia2023model, fan2024salun, cha2024learning, bonato2024retain, chen2024unsc, zhang2024contrastive, zhao2024makes}. The most common metrics are Forget Accuracy and Retain Accuracy. Forget Accuracy measures how effectively the model removes knowledge related to the data intended to be forgotten, typically by evaluating the model’s performance on the forget set. In contrast, Retain Accuracy assesses how well the model preserves its performance on the retain set, thus ensuring that unlearning does not degrade the model’s utility on the retain set.
Additionally, Membership Inference Attacks (MIA) have become a standard evaluation metric in machine unlearning to quantify the privacy risks associated with residual information about the forget data in models \citep{carlini2021extracting, song2021systematic, golatkar2021mixed, cho2022selective}. Introduced in early privacy research \citep{shokri2017membership}, MIA simulates an adversary’s attempt to determine whether a given data sample was included in the training set by analyzing the model’s outputs. In the context of machine unlearning, MIA is used to assess the effectiveness of unlearning algorithms by measuring whether an attacker can still infer membership in the forget set after the unlearning process. Recent unlearning work \citep{jia2023model} employs MIA to rigorously evaluate their unlearning method, demonstrating that their method significantly reduces the success rate of MIA on the forget set compared to prior methods. While these metrics have served as fundamental tools for evaluation, they predominantly focus on output-level behaviors and may not fully capture changes in the internal representation or feature space of the model. Consequently, recent works have begun to emphasize the need for representation-based evaluation frameworks \citep{jeon2024information, lee2025esc, siddiqui2025dormant, sepahvand2025selective}.

\noindent\textbf{Representation-based evaluation in machine unlearning.} \ \
Recent research in machine unlearning has begun to acknowledge the limitations of conventional logit-based evaluation, shifting towards methods that analyze the model's internal representations \citep{jeon2024information, lee2025esc, siddiqui2025dormant, sepahvand2025selective}. \cite{jeon2024information} proposed an approach based on information theory. They introduced the Information Difference Index (IDI), a white-box metric that quantifies residual information by measuring the mutual information between intermediate features and the forgetting labels. IDI is a powerful tool for detecting subtle information leakage that black-box metrics might miss by tracking the information within the model. \cite{lee2025esc} introduced the concept of Knowledge Deletion (KD), pointing out that existing unlearning methods often modify only the classification head while leaving the feature extractor largely unchanged. To cope with this, they proposed the Knowledge Retention (KR) score, a metric that measures the accuracy of a linear classifier retrained on the frozen feature extractor of an unlearned model.
\\
In contrast to these prior works, our evaluation framework offers two key advantages. First, it does not require any additional training or label supervision, enabling a training-free and computationally efficient evaluation that can be applied across diverse downstream datasets (\eg, Office-Home). Second, unlike IDI or KR, which still rely on accuracy (\eg, linear head retraining with binary labels in IDI or classification in KR), our approach directly measures the representational distance between the unlearned model, the original model, and the gold standard, allowing us to quantitatively assess how unlearning impacts the final model. Furthermore, our framework goes beyond proposing a representation-based metric. It also introduces large-scale evaluation settings, practical transferability assessment across diverse downstream datasets, and the Top Class-wise Forgetting scenario. While these prior works evaluate small-scale scenarios such as CIFAR-10, Tiny ImageNet, or ImageNet-1K with only 5-class forgetting, our framework targets large-scale unlearning involving 100–300 classes and proposes the Top Class-wise Forgetting setup to address the challenges of evaluating unlearning performance under random class forgetting. In summary, our evaluation framework advances beyond previous evaluation metrics by incorporating the training-free, computationally efficient evaluation metric from a transfer learning perspective and a novel, large-scale scenario.

\section{Why We Need Better Benchmarks: An Empirical Study}
\label{sec:emprical_study}
In this section, we formulate the problem and present a comprehensive analysis of existing machine unlearning methods, focusing on both logit-based and representation-based evaluations in large-scale settings. We review classical unlearning evaluation protocols and introduce novel evaluation metrics and scenarios. Next, we motivate our approach by highlighting the limitations of current evaluation frameworks. Then, in Sec.~\ref{sec:evaluation_framework}, we introduce our proposed evaluation metric, present experimental results under large-scale settings from a representation perspective, and report findings from Top Class-wise Forgetting scenarios from a transfer learning perspective.

\subsection{Problem Setup}
\label{sec:problem_setup}
We define the entire dataset as $\mathcal{D} = \{ (x_i, y_i) \}_{i=1}^N$, where $x_i$ represents the input image, $y_i$ denotes its corresponding label, and $N$ is the total number of samples in the dataset. The subset of data targeted for unlearning, referred to as the forget set, is denoted by $\mathcal{D}_f \subset \mathcal{D}$. The retain set, denoted by $\mathcal{D}_r = \mathcal{D} \setminus \mathcal{D}_f$, includes all samples in $\mathcal{D}$ that are not part of the forget set.
Additionally, we define the test forget set, denoted as $\mathcal{D}^{\text{te}}_f$, which consists of data points from the same classes as \dataf but are not included in the train set. Similarly, the test retain set, denoted as $\mathcal{D}^{\text{te}}_r$, contains data points from the same classes as \datar. We assume that the original model \thetao is trained on the entire dataset $\mathcal{D}$ and subsequently unlearns the knowledge related to the forget set $\mathcal{D}_f$. Note that the goal of machine unlearning is to produce an unlearned model \thetau that retains the knowledge of \datar while achieving the unlearning of \dataf. Additionally, we consider a retrained model \thetar, which is trained on \datar from scratch, as an oracle for comparison in the unlearning process.

\subsection{Logit-based Evaluation Metrics}
\label{logit_based_evaluation_metrics}
As a classical evaluation protocol, unlearning models are evaluated based on the following metrics: \textbf{Forget Accuracy (FA):} Measures the accuracy of the unlearned model \thetau on \dataf. \textbf{Retain Accuracy (RA):} Measures the accuracy of the unlearned model \thetau on \datar. \textbf{Test Forget Accuracy (TFA):} Classification accuracy on the test forget set $ \mathcal{D}^{\text{te}}_f $. Similar to FA, this evaluates the model’s ability to forget removed classes beyond the training data. \textbf{Test Retain Accuracy (TRA):} Classification accuracy on the test retain set $ \mathcal{D}^{\text{te}}_r $. It indicates how effectively the model’s knowledge generalizes to unseen data from the remaining classes. Following \citep{zhao2024makes}, we introduce the \textbf{Average GAP of Logit-based Metrics (AGL)}, which quantifies the discrepancy between the given model (\eg, the unlearned model \thetau) and the gold standard \thetar. This metric aggregates the logit-based evaluations:
\begin{equation}
\begin{aligned}
AGL = & \Bigl(1 - \mathrm{G}(\theta_u,  \mathcal{D}_f)\Bigr) \times \Bigl(1 - \mathrm{G}(\theta_u,  \mathcal{D}_r)\Bigr) \\
& \times \Bigl(1 - \mathrm{G}(\theta_u, \mathcal{D}^{\text{te}}_{f})\Bigr) \times \Bigl(1 - \mathrm{G}(\theta_u,  \mathcal{D}^{\text{te}}_{r})\Bigr),
\end{aligned}
\label{eq:agl}
\end{equation}
where \dataf, \datar, \( \mathcal{D}^{\text{te}}_{f} \), and \( \mathcal{D}^{\text{te}}_{r} \) represent the source training datasets and their corresponding test sets, and $\mathrm{G_{}}$ computes the classification accuracy differences, 
$\mathrm{G_{}}(\theta_{u}, \mathcal{D}_{\text{}}) = \bigl| \, \text{A}(\theta_{u}, \mathcal{D}_{\text{}}) - \text{A}(\theta_{r}, \mathcal{D}_{\text{}}) \bigr|$,
where $\mathrm{A}$ represents the classification accuracy on $\mathcal{D}$.

\noindent\textbf{Membership Inference Attack (MIA).} \ \
Membership Inference Attack efficacy is measured using a prediction confidence-based attack method \citep{song2021systematic}. The process includes two phases: (1) training, where a balanced dataset from \datar and unseen test data is used to train the Support Vector Machine (SVM)-based MIA predictor; and (2) testing, where the trained MIA model evaluates the unlearning performance. It measures the unlearning effectiveness by applying the MIA binary classifier to the unlearned model \thetau on the forget set \dataf. The more the model predicts the forget set as test data, the higher the resulting MIA-efficacy. MIA-Efficacy evaluates how effectively unlearning removes information by using privacy attack methods. We report the MIA results in Table~\ref{tab:mia}.

\subsection{Representation-based Evaluation Metrics}
To overcome the limitations of logit-based evaluation, we propose an additional representation-based approach that assesses unlearning effectiveness by comparing features extracted from the encoder (\eg, after the average pooling layer in ResNet) between the unlearned and retrained models. This enables a fine-grained analysis of representational changes beyond what output-level metrics can capture. We use the following two metrics for the representation-based evaluation: First, we use \textbf{Centered Kernel Alignment (CKA)}~\citep{kornblith2019similarity} to measure the similarity between the feature representations of \thetau and \thetar, as well as \thetau and \thetao. Second, we measure \textbf{$k$-Nearest Neighbors Accuracy ($k$-NN)}~\citep{cover1967nearest} to assess the representational quality of \thetau. We employ a $k$-NN classifier, which takes features of data from a targeted model and performs classification based on the neighborhood features. Additionally, we also employ t-SNE~\citep{JMLR:v9:vandermaaten08a} to visualize feature distributions.

\noindent\textbf{Centered Kernel Alignment (CKA).} \ \
A successful unlearning algorithm can be identified if the feature representations of the unlearned model align closely with those of the retrained model, which serves as the oracle. To quantitatively evaluate this alignment, we employ Centered Kernel Alignment (CKA)~\citep{kornblith2019similarity}, a robust metric for assessing the similarity between feature representations. CKA operates by comparing the normalized Gram matrices of two sets of feature embeddings. Specifically, the CKA between two Gram matrices $ K $ and $ L $ is defined as:
\begin{equation}
\text{CKA}(K, L) = \frac{\| \text{HSIC}(K, L) \|_F^2}{\| \text{HSIC}(K, K) \|_F^2 \cdot \| \text{HSIC}(L, L) \|_F^2},
\label{eqn:cka}
\end{equation}
where $ K $ and $ L $ represent the Gram matrices of the two feature representations, and HSIC denotes the Hilbert-Schmidt independence criterion. By applying CKA, we measure the similarity between the feature representations of \thetau (the unlearned model), \thetao (the original model), and \thetar (the retrained model) across various downstream datasets. Note that higher CKA values indicate a greater degree of alignment, suggesting that the unlearned model retains the desired feature representations after unlearning.

\noindent\textbf{$k$-Nearest Neighbors ($k$-NN).} \ \
To evaluate the transferability of the unlearned model, we employ the $k$-NN method. Feature embeddings are extracted from the unlearned model, \thetau, by feeding it the downstream data. These embeddings are then split, with 80\% used to train a $k$-NN classifier and the remaining 20\% used for testing. For each test sample, the classifier assigns a label based on the majority vote of its $k$ nearest neighbors from the training set. The $k$-NN Accuracy is computed as the proportion of correctly classified test samples. This evaluation provides insights into whether \thetau preserves meaningful, transferable feature embeddings that are effective for downstream tasks. We set $k=5$ in the $k$-Nearest Neighbors ($k$-NN) classifier in experiments, using standard cosine distance. Ties in the majority vote are broken by selecting the class with the lowest numerical index. However, we consider its impact negligible, as such occurrences were rare in our evaluation.

\noindent\textbf{t-SNE visualization.} \ \
To understand how unlearning affects a model’s internal features, we visualized the features using t-SNE. Instead of visualizing the full dataset, we selected a consistent, randomly chosen subset of classes from both the forget set and the retain set due to computational costs. This subset is identical across all unlearning algorithms, ensuring fair comparison. The t-SNE visualization of all unlearning baselines is shown in Fig.~\ref{fig:full_visualization}.

\noindent\textbf{Average GAP of Representation-based metrics (AGR).} \ \
AGR is designed to holistically assess representation-level unlearning by measuring two key aspects: the feature representational alignment with the retrained model, measured by $\mathrm{CKA}_{\text{AGR}}$, and the preservation of transferable knowledge, evaluated by $\mathrm{G_{kNN}}$. Details on these metrics ($\mathrm{CKA}_{\text{AGR}}$, $\mathrm{G_{kNN}}$) are introduced in Sec.~\ref{subsec:unified_metric}.
The AGR score is computed by combining these two components as follows:
\begin{equation}
    AGR = \Bigl(1 - \mathrm{G_{kNN}}(\theta_u, \mathcal{D}_{\text{down}})\Bigr) \times \mathrm{CKA}_{\text{AGR}}(\theta_u, \mathcal{D}_{\text{down}}).
\end{equation}
The calculation of these components depends on the unlearning scenario. For the Random Class-wise Forgetting scenario, we measure $\mathrm{G_{kNN}}$ and $\mathrm{CKA}_{\text{AGR}}$ across all downstream datasets and use their average values.
However, for the Top Class-wise Forgetting scenario, the metrics are measured only on the semantically related downstream dataset. For instance, when evaluating a model under the CUB Top-100 Class-wise Forgetting scenario, we calculate $\mathrm{G_{kNN}}$ and $\mathrm{CKA}_{\text{AGR}}$ using only the CUB dataset.


\subsection{Models and Datasets}
\noindent\textbf{Models and datasets.} \ \
To ensure a more meaningful evaluation of unlearning algorithms considering real-world scenarios, we employ the ResNet-50~\citep{he2016deep} model pre-trained on ImageNet-1K~\citep{deng2009imagenet} as the large-scale setting. We further extend our experiments to include the Swin-T \citep{liu2021swin} and ConvNeXt \citep{liu2022convnext} architectures.

\noindent\textbf{Downstream datasets.} \ \
\label{subsec:downstream_datasets}
For our evaluation framework, we employ three diverse downstream datasets: \textbf{Office-Home} \citep{venkateswara2017deep}: A domain adaptation benchmark comprising distinct domains. We use Real World, Art, Clipart, and Product domains. It includes 65 classes with over 15,500 images, providing a robust basis for cross-domain transfer learning. \textbf{CUB-200-2011 (CUB)} \citep{wah2011caltech}: The Caltech-UCSD Birds dataset features fine-grained classification with 200 bird species and 11,788 images, suitable for evaluating detailed representational quality. \textbf{DomainNet-126} \citep{peng2019moment}: A multi-source domain dataset containing several domains. We use Clipart, Painting, Real, and Sketch domains. This dataset includes 126 classes.

\subsection{Unlearning Baselines}
We divide unlearning baselines into two categories based on whether they use the retain set (\datar) or not: (i) baselines without \datar, and (ii) baselines with \datar. The details of these algorithms are as follows.

\noindent\textbf{Unlearning w/o \datar.} \ \
In this section, baselines aim to remove information using only \dataf without the retain set. While this reduces time complexity and the privacy risk, it often sacrifices the accuracy on the retain set.

\noindent\textbf{Gradient Ascent (GA)~\citep{thudi2022unrolling}.} \ \
GA forces the model to misclassify \dataf by maximizing the loss on forget samples. For a set of forget tuples $\{(x_i,y_i)\} \subset \mathcal{D}_f$, GA updates model parameters \thetao with the cross-entropy loss $\ell$:
\begin{equation}
\centering
\label{eq:ga}
    \theta \;\leftarrow\; \theta \;+\; \eta\,\nabla_{\theta} 
    \Bigl[
      \sum_{(x_i,y_i)\in \mathcal{D}_f}
      \ell\bigl(f_{\theta}(x_i),\,y_i\bigr)
    \Bigr].
\end{equation}

\noindent\textbf{Random Labeling (RL)~\citep{golatkar2020eternal}.} \ \
RL discards ground-truth labels of the forget set and replaces each $y_i$ with a label $y_i^\mathrm{rand}$ sampled from other classes:
\begin{equation}
\label{eq:rl}
    \min_{\theta}\;
    \sum_{(x_i,\,y_i^\mathrm{rand})\in \mathcal{D}_f}
    \ell\bigl(f_{\theta}(x_i),\,y_i^\mathrm{rand}\bigr).
\end{equation}
By training on random labels for \dataf, the network forgets the original associations.

\noindent\textbf{Pseudo Labeling (PL)~\citep{chen2024unsc}.} \ \
PL ensures that \dataf is unlearned in a manner that minimizes noise. For each sample \( x_i \in \mathcal{D}_f \), the original model \thetao assigns a pseudo-label \( y_i' \), representing the class with the highest predicted probability after excluding the forget set. The unlearning objective is defined as:
\begin{equation}
\label{eq:pl}
    \min_{\theta}\;
    \sum_{(x_i,\,y_i')\in \mathcal{D}_f}
    \ell\bigl(f_{\theta}(x_i),\,y_i'\bigr).
\end{equation}
where \( y_i' \) is the pseudo-label and \( \ell \) denotes the cross-entropy loss. Unlike random or heuristic labeling methods, Pseudo-Labeling ensures that \dataf is mapped to semantically similar classes, preserving the decision boundary of the model.

\noindent\textbf{SalUn~\citep{fan2024salun}.} \ \
SalUn restricts unlearning updates to only those parameters deemed salient for classifying \dataf. Let $\theta_{\mathrm{s}} \subseteq \theta$ be the parameter subset. Then the update is
\begin{equation}
\label{eq:salun}
    \theta_{\mathrm{s}} \;\leftarrow\;
    \theta_{\mathrm{s}} \;-\; \eta\,\nabla_{\theta_{\mathrm{s}}}
    \sum_{(x_i,y_i)\in \mathcal{D}_f}
    \ell\bigl(f_{\theta}(x_i),\,y_i\bigr).
\end{equation}
This targets \dataf knowledge at the parameter level while minimally disturbing the rest of the network.

\noindent\textbf{Unlearning w/ \datar.} \ \
Several baselines utilize \datar, as using only \dataf often fails to preserve the knowledge of \datar. Baselines in this category incorporate training or distillation on the retain set, thereby improving utility for the retain set.

\noindent\textbf{DUCK~\citep{cotogni2023duck}.} \ \
DUCK employs a distance-based metric learning objective to destroy class cues for $\mathcal{D}_f.$ Suppose each valid class $c\neq y_i$ has a centroid $\mathbf{m}_c.$ For each forget set $(x_i,y_i),$ DUCK shifts $f_{\theta}(x_i)$ (the feature embedding) to the nearest incorrect centroid:
\begin{equation}
\label{eq:duck}
    \begin{aligned}
        \min_{\theta} \; &\sum_{(x_i,y_i) \in \mathcal{D}_f} 
        \mathrm{dist}\Bigl(f_{\theta}(x_i), \mathbf{m}_{c^*} \Bigr), \ \ c^* = \; \arg\min_{c\neq y_i} 
        \mathrm{dist}\bigl(f_{\theta}(x_i), \mathbf{m}_c\bigr).
    \end{aligned}
\end{equation}
In doing so, the feature space for forget classes is pushed into the incorrect boundary.

\noindent\textbf{Contrastive Unlearning (CU)~\citep{zhang2024contrastive}.} \ \
CU uses a contrastive objective to decouple the forget set from the retain set in the embedding space.
Though one can implement CU without \datar explicitly in every step (\textit{e.g.}, by approximating distances), it is often more effective when \datar is used to preserve the retain set.

\noindent\textbf{SCRUB~\citep{kurmanji2023towards}.} \ \
SCRUB uses a teacher–student paradigm. The original model (teacher) provides knowledge about \datar, but its information on \dataf is filtered. If \( f_{\theta}(x) \) is the teacher’s feature, and $f_{\theta_u}(x)$ is the student’s, SCRUB trains \thetau to match the teacher’s logits on \datar, while removing any relevant signals from \dataf. One can write a selective distillation objective:
\begin{equation}
    \begin{aligned}
        \max_{\theta_u} \;
        &\sum_{(x_j,y_j) \in \mathcal{D}_f} 
        d\bigl(f_{\theta_u}(x_j), f_{\theta}(x_j)\bigr)
        + \min_{\theta_u}
        \sum_{(x_j,y_j) \in \mathcal{D}_r} 
        d\bigl(f_{\theta_u}(x_j), f_{\theta}(x_j)\bigr).
    \end{aligned}
\end{equation}
with an additional mechanism that blocks or penalizes the student's copying of the teacher’s information for classes in \dataf.

\noindent\textbf{SCAR~\citep{bonato2024retain}.} \ \
SCAR exploits a distance-based realignment for \dataf, pushing their features toward incorrect classes to erase the original boundary, which preserves the representation for \datar via distillation or cross-entropy, minimizing these terms:
\begin{equation}
    \begin{aligned}
        &\sum_{(x_i,y_i) \in \mathcal{D}_f} \mathrm{dist}\bigl(f_{\theta}(x_i), \mathbf{m}_{c^*}\bigr) \quad + \sum_{(x_j,y_j) \in \mathcal{D}_r} \ell\bigl(f_{\theta}(x_j), y_j\bigr).
    \end{aligned}
\end{equation}
where $\mathbf{m}_{c^*}$ is again an incorrect-class centroid. We use retained classes for \datar in this paper.

\subsection{Unlearning Scenarios}
In our experiments and analysis, we consider two unlearning scenarios. The first is Random-$N$ Class-wise Forgetting, where we randomly select $N$ classes to remove. This scenario is a standard approach to evaluate unlearning algorithms. To provide a more comprehensive assessment, we introduce a second scenario, Top-$N$ Class-wise Forgetting, in which we specifically target classes that exhibit semantic similarity to the downstream task classes. The selection of $N$ classes is based on their relevance to the downstream task.

\subsection{Motivation: Why Do We Need Representation-Based Metrics?}
\label{subsec:motivation}
Current machine unlearning algorithms rely on logit-based evaluation metrics to assess their effectiveness. These algorithms aim to achieve classification accuracy similar to a model trained only on the retain set (\thetar), and based on these metrics, they are considered to have successfully accomplished unlearning. However, this evaluation raises a critical research question:
\begin{tcolorbox}[colback=gray!10, boxrule=1pt]
\textit{Do existing unlearning evaluations adequately reflect successful unlearning from a representation perspective?}
\end{tcolorbox}

To explore this question, we investigate whether unlearned models (\thetau) exhibit feature representations that are more similar to those of the retrained model (\thetar) while remaining significantly different from the original model (\thetao). To do this, we randomly select 100 forget classes from the 1,000 classes in ImageNet-1K and perform unlearning using baseline algorithms.

\begin{figure*}[t!]
    \vspace{-4mm}
    \begin{center}
    \end{center}
    \hfill
    \begin{subfigure}[b]{0.28\linewidth}
        \centering
        \includegraphics[width=\linewidth]{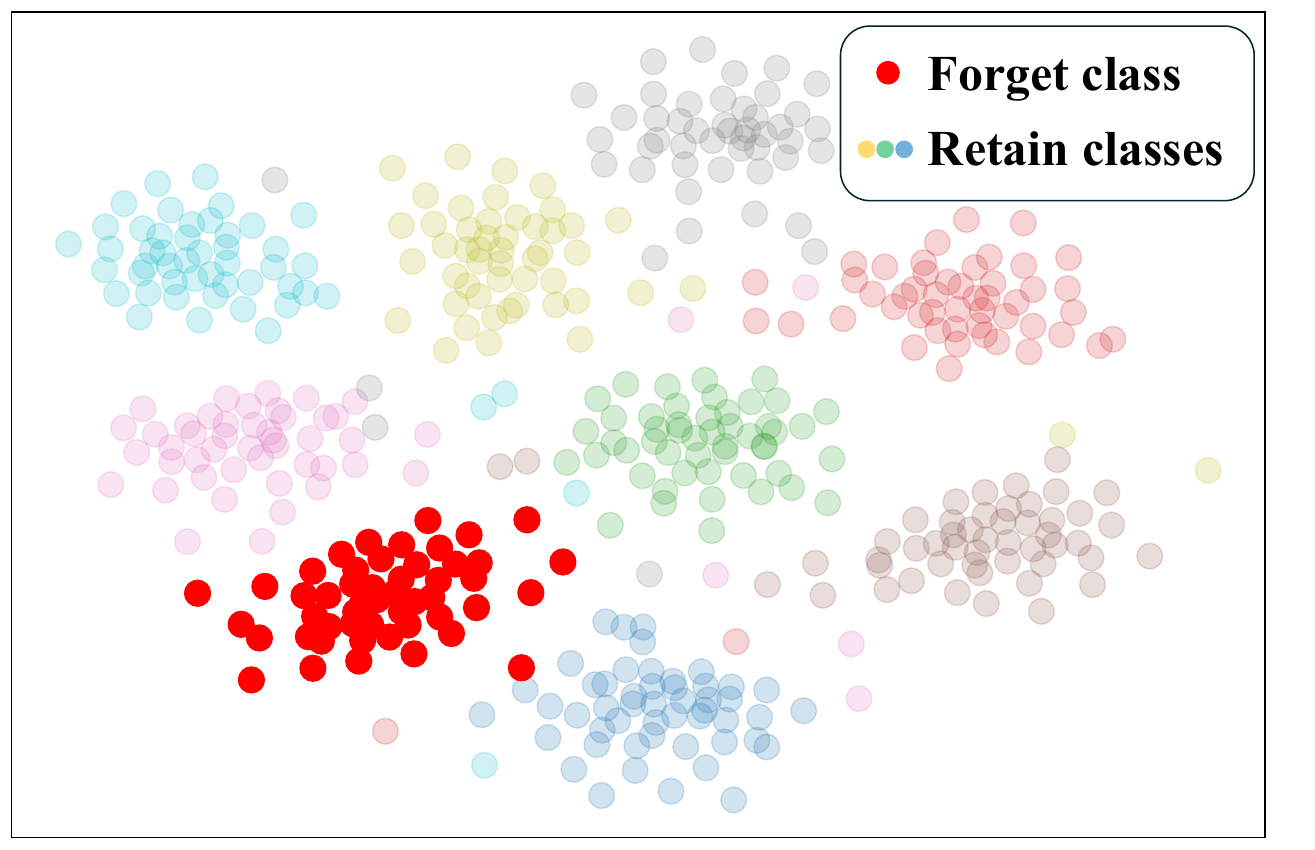}
        \caption{Original (\thetao)}
    \end{subfigure}
    \hspace{1em}
    \begin{subfigure}[b]{0.28\linewidth}
        \centering
        \includegraphics[width=\linewidth]{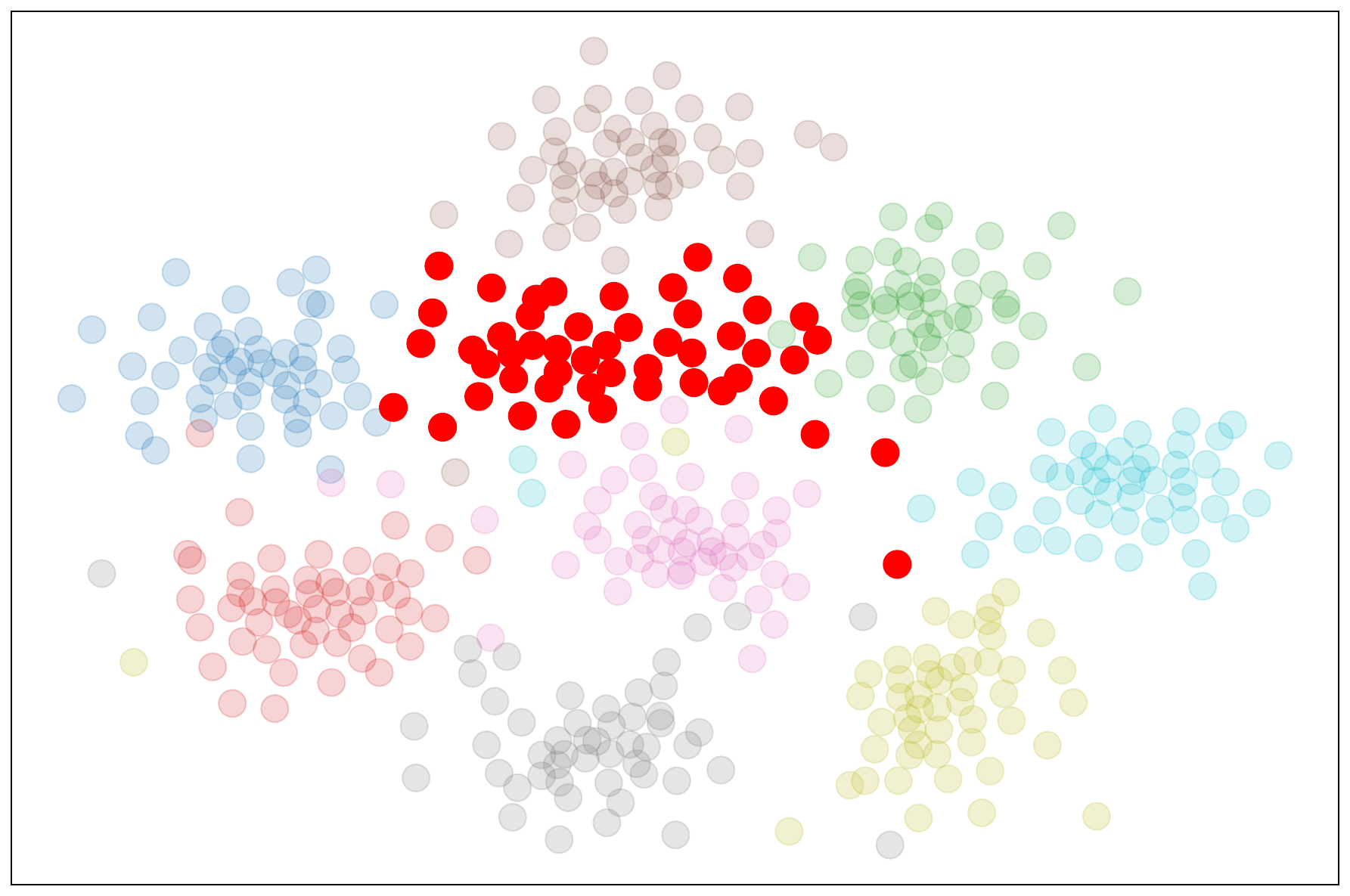}
        \caption{Retrained (\thetar)}
    \end{subfigure}
    \hspace{1em}
    \begin{subfigure}[b]{0.28\linewidth}
        \centering
        \includegraphics[width=\linewidth]{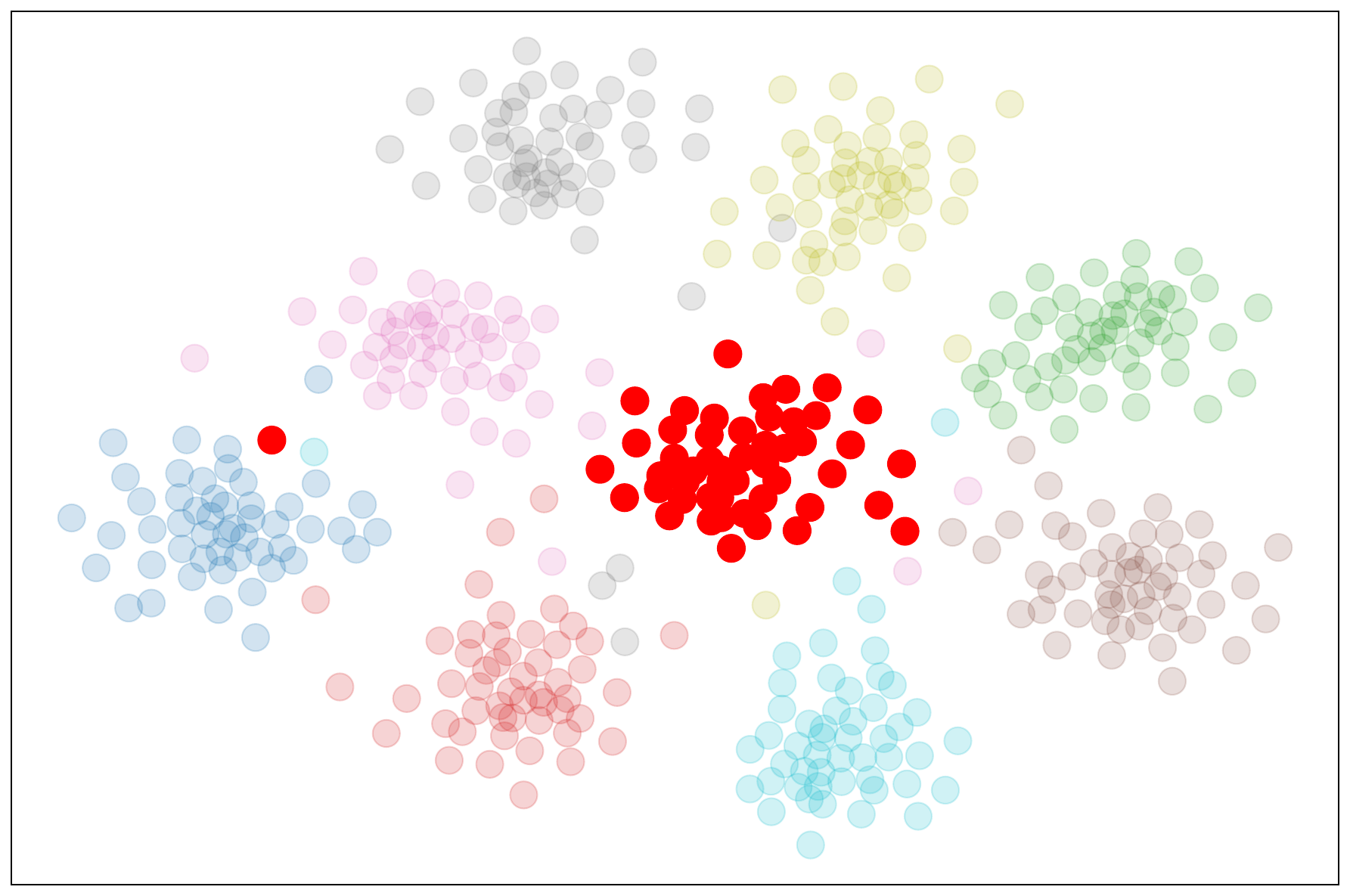}
        \caption{FT}
    \end{subfigure}
    \hfill
    \vspace{1em}
    \vspace{1em}
    \hfill
    \begin{subfigure}[b]{0.28\linewidth}
        \centering
        \includegraphics[width=\linewidth]{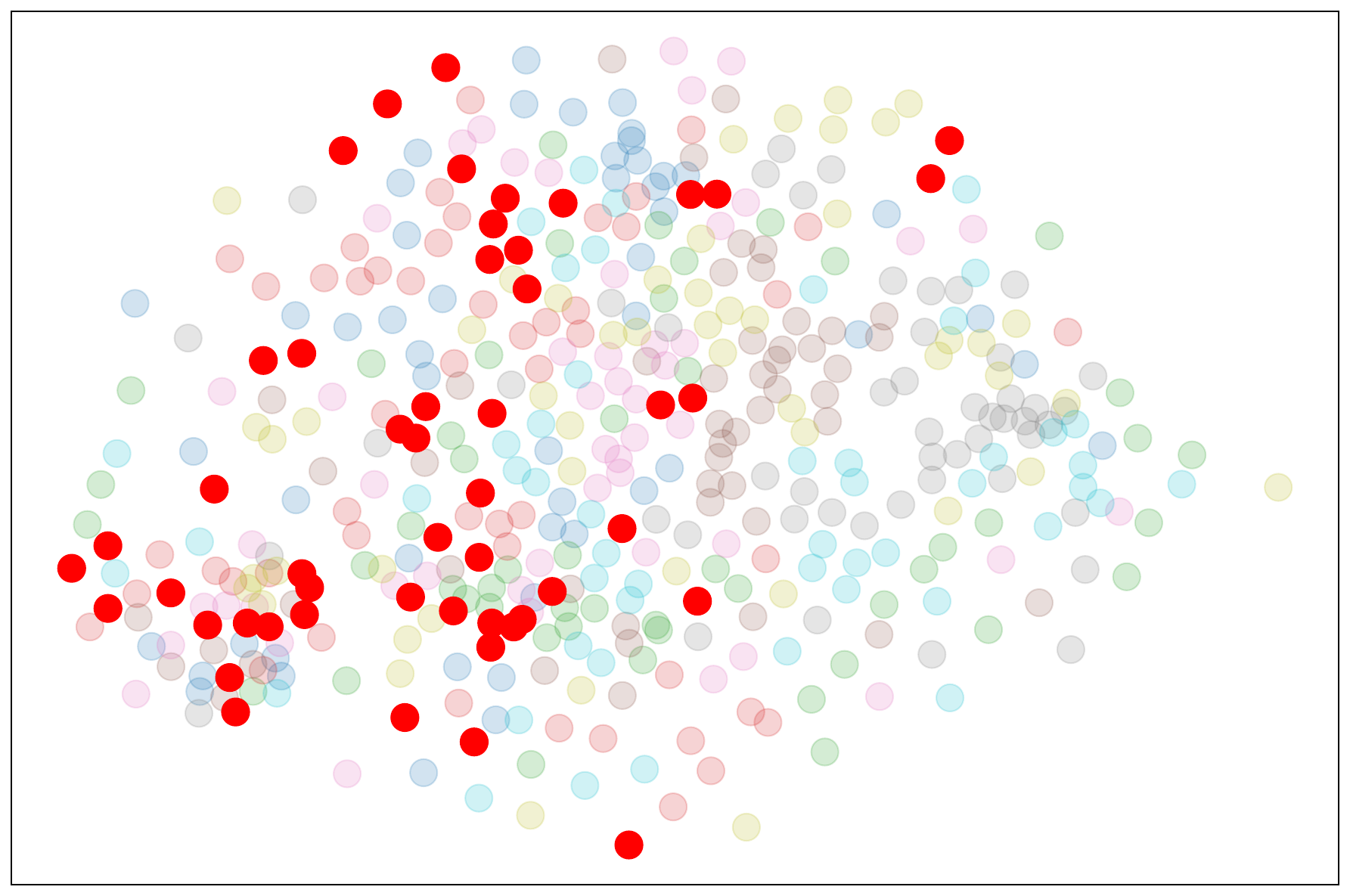}
        \caption{GA}
    \end{subfigure}
    \hspace{1em}
    \begin{subfigure}[b]{0.28\linewidth}
        \centering
        \includegraphics[width=\linewidth]{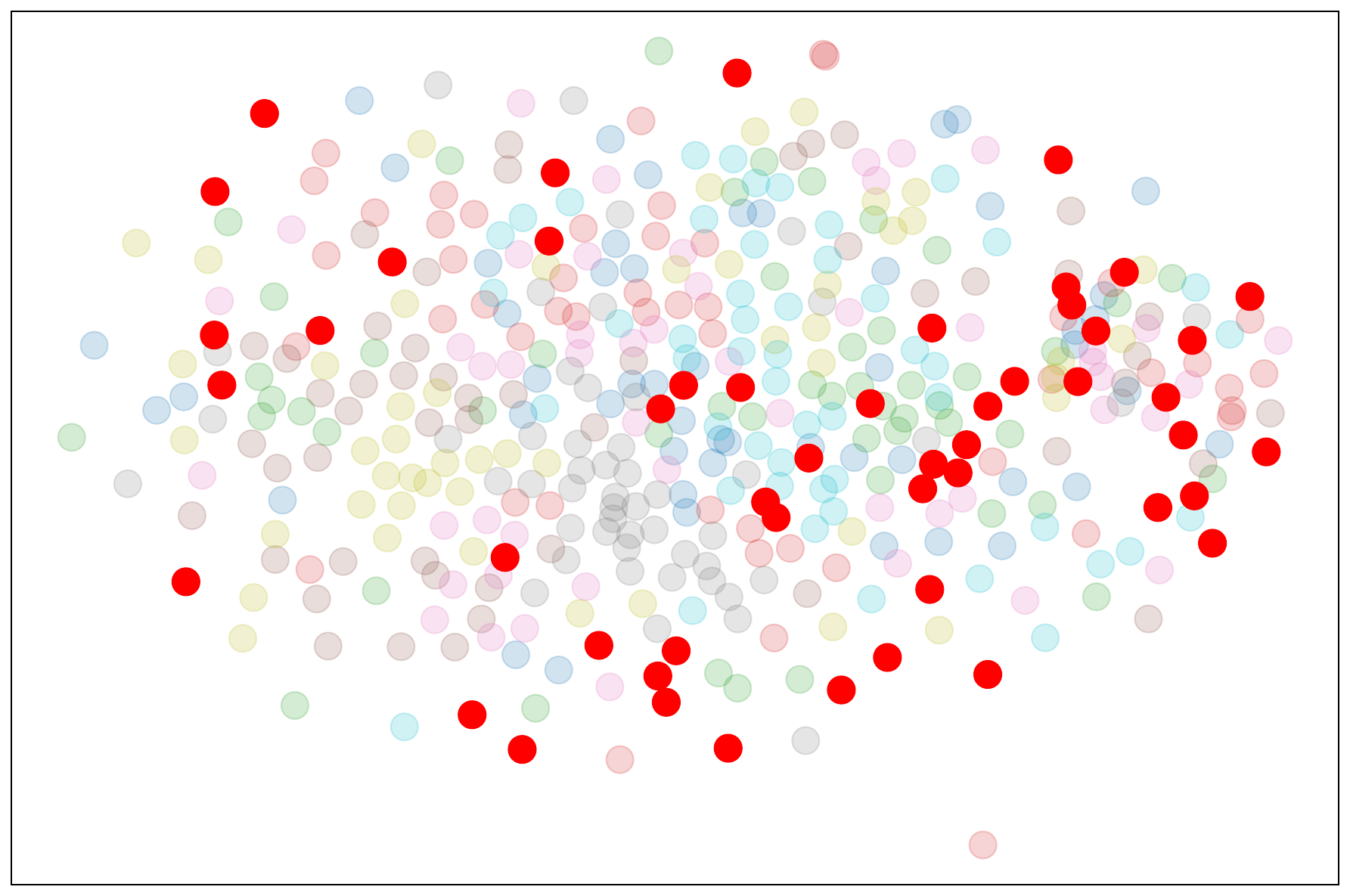}
        \caption{RL}
    \end{subfigure}
    \hspace{1em}
    \begin{subfigure}[b]{0.28\linewidth}
        \centering
        \includegraphics[width=\linewidth]{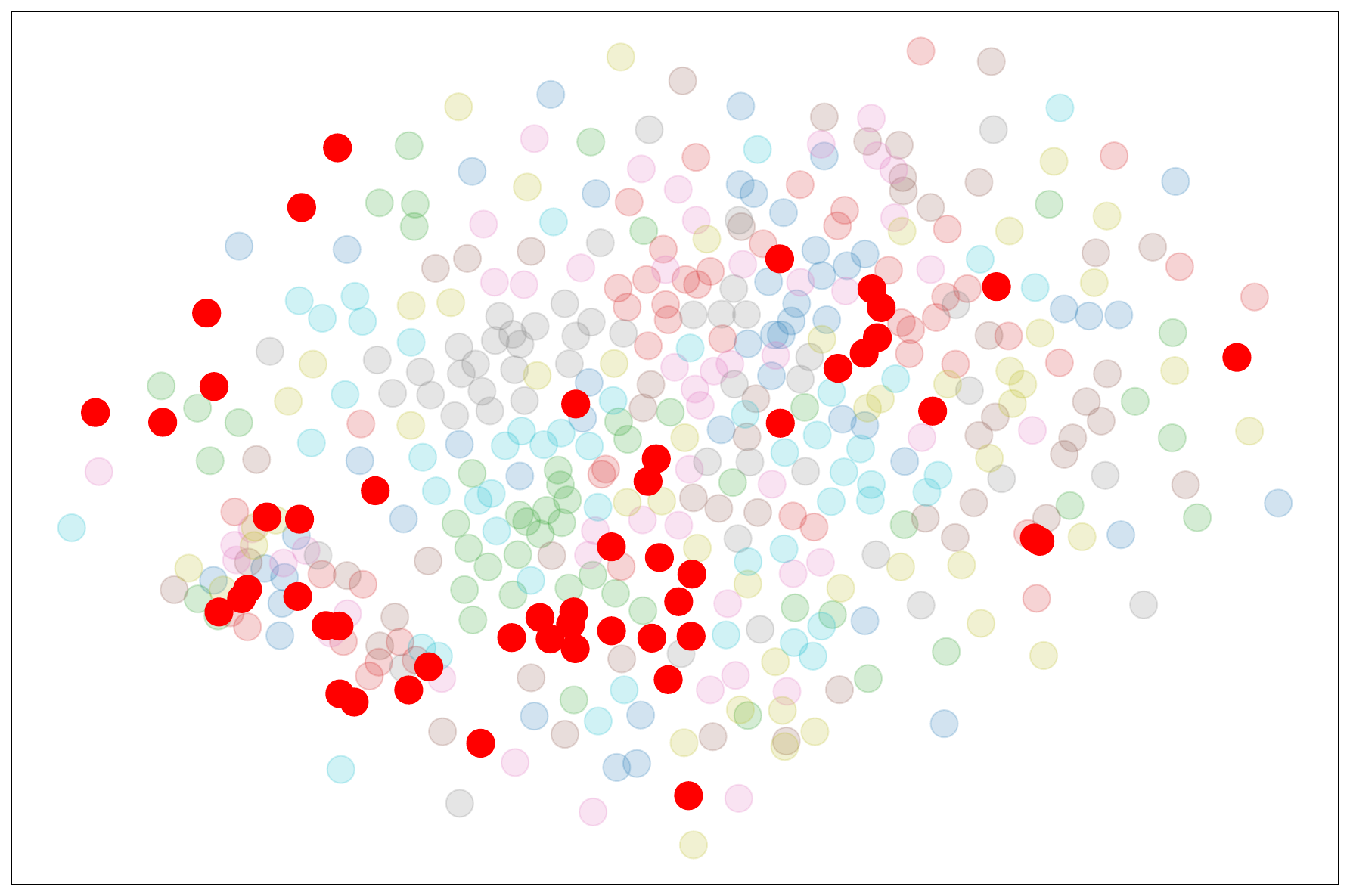}
        \caption{SalUn}
    \end{subfigure}
    \hfill
    \vspace{1em}
    \vspace{-8mm}
    \caption{We visualize the feature representations from the original (\thetao), retrained (\thetar), and unlearned models (\thetau) on a subset of ImageNet-1K using ResNet-50. Unlike \thetar, which serves as the gold standard, the forget classes (highlighted in red) in the existing unlearning baselines (\eg, (d), (e), and (f)) are severely dispersed throughout the entire class distribution.}
    \label{fig:visualization}
    \vspace{-5mm}
\end{figure*}

\noindent\textbf{t-SNE visualization.} \ \
We first conducted an analysis on several unlearning baselines using t-SNE in Fig.~\ref{fig:visualization}, which provides a visualization of feature embeddings from the encoder in a lower-dimensional space. While several unlearning algorithms (GA, RL, SalUn) achieve superior performance on logit-based evaluation metrics in small-scale settings,  Fig.~\ref{fig:visualization} reveals several important findings. First, (a) Original (\thetao), (b) Retrained (\thetar), and (c) Fine-Tuning (FT) models exhibit feature representations where each class forms distinct clusters. We expect that an effective unlearning algorithm ideally produces feature representations similar to those of (b) (\thetar). However, Figures (d)-(f) reveal that these approaches induce substantial collapse in feature representations, which deviate considerably from those of \thetar.


\begin{figure*}[t]
    \centering
    \begin{subfigure}[b]{0.46\textwidth}
        \centering
        \includegraphics[width=\textwidth]{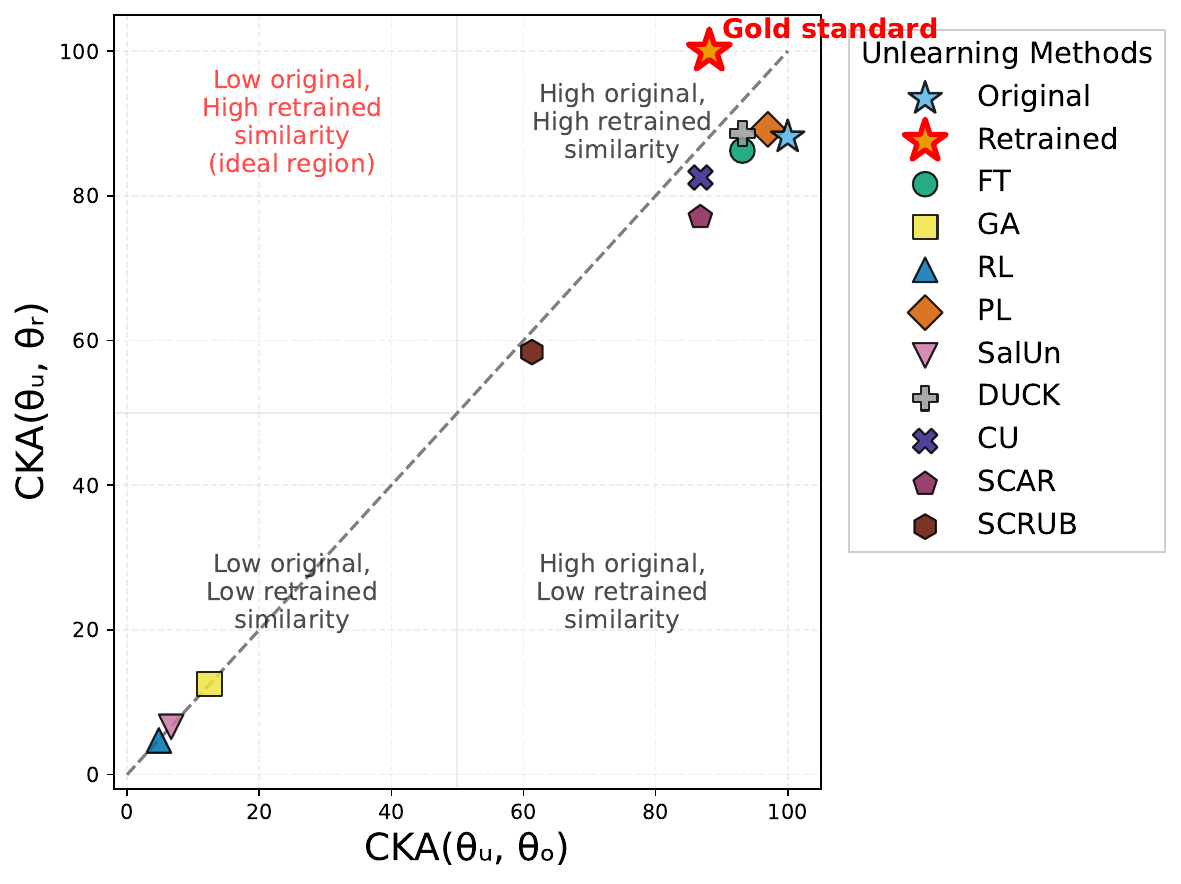}
        \caption{CKA similarity of unlearning baselines to the original or retrained model}
    \label{fig:originalvsretrained}
    \end{subfigure}
    \hspace{4mm}
    \begin{subfigure}[b]{0.46\textwidth}
        \centering
        \vspace{-4mm}
        \includegraphics[width=\textwidth]{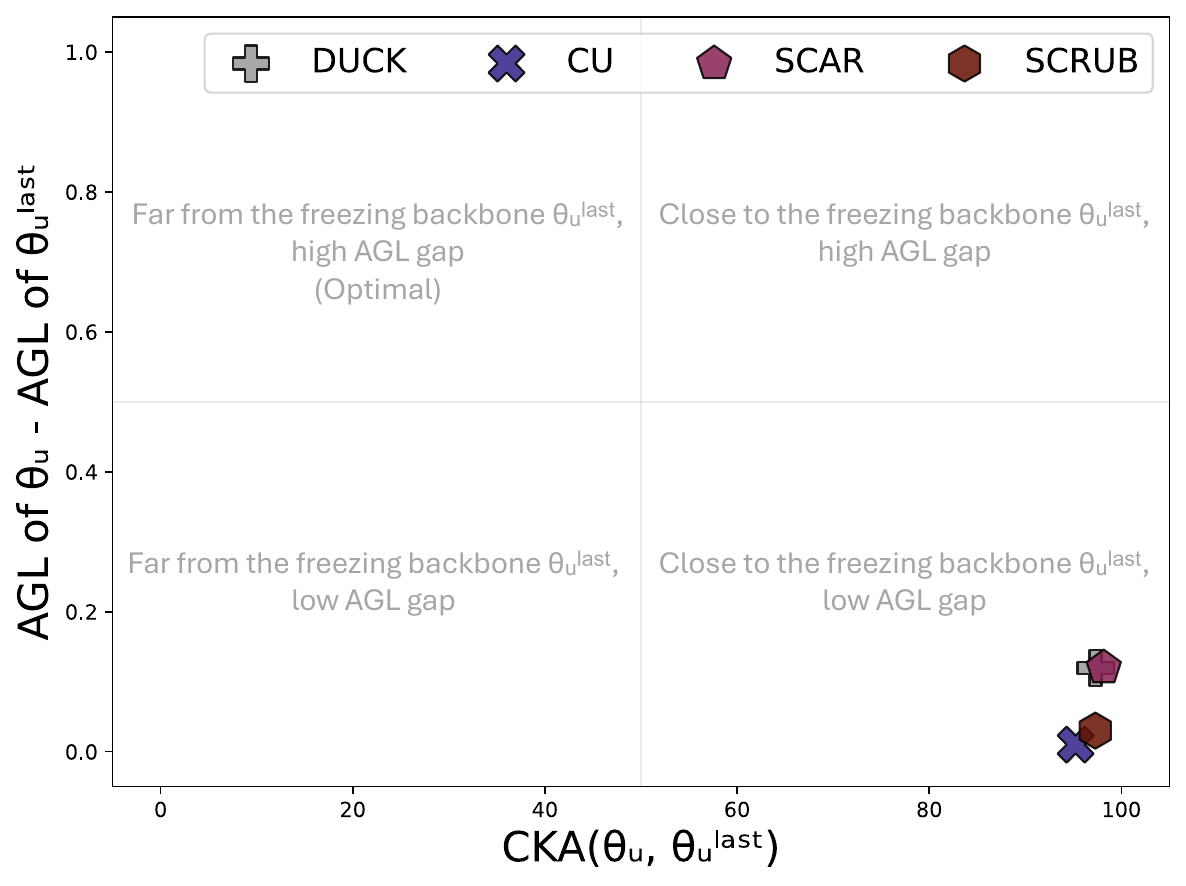}
        \caption{An analysis of the last layer \thetaulast}
    \label{fig:maintain}
    \end{subfigure}
    \caption{Two analyses using CKA similarity are presented: (a) CKA similarity analysis of various unlearning algorithms. The x-axis shows the similarity to \thetao and the y-axis represents the similarity to \thetar. In an ideal scenario, algorithms should be positioned near \thetar. However, most algorithms exhibit a high similarity to \thetao, suggesting that the transformation of representations during unlearning is suboptimal. (b) The X-axis depicts the feature similarity between the fully unlearned model (\thetau) and only the final layer unlearned model (\thetaulast). Y-axis represents the gap in AGL scores between \thetau and \thetaulast. The results indicate that existing algorithms primarily modify only the last layer while maintaining the original representation space.}
\end{figure*}

\noindent\textbf{CKA analysis.} \ \
In Fig.~\ref{fig:originalvsretrained}, we compare the Centered Kernel Alignment (CKA) similarity values between the unlearned (\thetau) and retrained (\thetar) models (on the y-axis) against those between the unlearned and original (\thetao) models (on the x-axis), using ImageNet-1K. Note that CKA quantifies the representational similarity between different models, providing a measure of how similarly two models interpret the given data. Intuitively, if two models are aligned in their feature representations, they can be considered to share a similar perspective on the data. In the context of successful unlearning, we expect the CKA similarity between the unlearned and retrained models to be higher than that between the unlearned and original models. However, our findings reveal a critical issue: all unlearning baselines show a higher similarity to the original model’s representations than to the retrained model’s representations. Also, certain baselines, such as GA, SalUn, and RL, exhibit very low similarity to both the original and retrained models.


\noindent\textbf{Why does the unlearned model maintain high similarity to the original model?} \ \
Several unlearning baselines show high logit-based metric scores (\ie, minimal deviation from the retrained model) in our experimental setup, while still maintaining a strong resemblance to the original model's representations. We hypothesize that this occurs because unlearning algorithms tend to focus on modifying the final layer to optimize logit-based metrics while leaving representational layers largely unchanged. In Fig.~\ref{fig:maintain}, we compare the CKA similarities between unlearned models, where either all layers are tuned (\thetau) or only the final layer is tuned (\thetaulast), alongside the gap in logit-based AGL (Eq.~\ref{eq:agl}) scores (y-axis). Ideally, the CKA similarity should be low (low x-values), which would correspond to a higher gap in AGL scores (high y-values). However, \thetau shows high CKA similarity to \thetaulast while presenting a relatively small gap in AGL, demonstrating that current unlearning algorithms predominantly focus on modifying the final layer. We posit that this behavior is a form of shortcut learning~\citep{arjovsky2019invariant}; algorithms converge to a sub-optimal solution by taking the path of least resistance to satisfy logit-based loss, without solving the harder problem of representational change. This highlights a crucial limitation in the current evaluation approach. To further investigate the optimization strategies that might overcome this shortcut learning phenomenon, we conduct additional experiments (\qq{Impact of optimization strategies on shortcut learning}) in Appendix B, examining the impact of acceleration methods like Nesterov momentum.

In summary, the results from our analyses highlight a key limitation in relying solely on logit-based metrics for unlearning evaluation. While some algorithms show high logit-based performance, their feature representations remain too similar to the original model. This underscores the need for representation-based metrics, which can more effectively capture the changes in feature representations and provide a more accurate measure of unlearning effectiveness, especially in large-scale, real-world applications.

\section{A Unified Benchmark Framework for Machine Unlearning}
\label{sec:evaluation_framework}
To establish a rigorous and comprehensive evaluation of unlearning algorithms, we propose a unified framework that integrates both logit-based and representation-based metrics. This framework enables a more holistic assessment of unlearning performance by capturing both \textbf{Feature Similarity} and \textbf{Transferability}—two fundamental aspects of an unlearned model's behavior. First, \textbf{Feature Similarity} quantifies the alignment between feature representations of the original and unlearned models on the same dataset. To measure this, we employ Centered Kernel Alignment (CKA), which provides a robust similarity measure between the learned representations of both models. Second, \textbf{Transferability} evaluates how well the unlearned model’s learned representations generalize to downstream tasks. We assess this by applying a $k$-Nearest Neighbors ($k$-NN) classifier to the encoder's output features and evaluating performance on several downstream datasets. This allows us to measure how well features generalize beyond the original training distribution. 

\subsection{Unified metric}
\label{subsec:unified_metric}
To establish a comprehensive evaluation of unlearning algorithms, we propose a unified measure that integrates both logit-based and representation-based evaluations. For representation-based evaluation, we use diverse downstream datasets, \(\mathcal{D}_{\text{down}}\),  which differ from the original train set distribution (\textit{e.g.}, ImageNet-1K). Inspired by transfer learning evaluation protocols~\citep{pan2010transfer}, this approach enables us to assess both the quality and generalizability of feature representations.
Specifically, we evaluate models on three downstream datasets: Office-Home~\citep{venkateswara2017deep}, CUB~\citep{wah2011caltech}, and DomainNet-126~\citep{peng2019moment}. To formally quantify representation changes, we define two key metrics: First, $\mathrm{G_{kNN}}$ measures the difference in classification accuracy  ($\text{A}^{k}$) of  a $k$-NN classifier between \thetau and \thetar using a downstream dataset:
\begin{equation}
\mathrm{G_{kNN}}(\theta_{u}, \mathcal{D}_{\text{down}}) = \bigl| \, \text{A}^{k}(\theta_{u}, \mathcal{D}_{\text{down}}) - \text{A}^{k}(\theta_{r}, \mathcal{D}_{\text{down}}) \bigr|.
\end{equation}
This metric evaluates the transfer learning ability of the given model’s encoder compared to that of the retrained model. 
In addition, we quantify the representation similarity between the given model and the retrained model for the downstream dataset, by using $\mathrm{CKA}_{\text{AGR}}$:
$\mathrm{CKA}_{\text{AGR}}(\theta_{u}, \mathcal{D}_{\text{down}}),$
which measures the representational similarity between the given and retrained models over the downstream dataset. Using these metrics, we define the Average GAP of Representation-based metrics (AGR) as:
\begin{equation}
AGR = \Bigl(1 - \mathrm{G_{kNN}}(\theta_u, \mathcal{D}_{\text{down}})\Bigr) \times \mathrm{CKA}_{\text{AGR}}(\theta_u, \mathcal{D}_{\text{down}}).
\label{eq:agr}
\end{equation}
The $(1 - \mathrm{G_{kNN}})$ term in Eq.~\ref{eq:agr} serves as a crucial normalization step to align the objectives of the two components. $\mathrm{G_{kNN}}$ measures the \textbf{gap} in $k$-NN accuracy, where a value closer to 0 is ideal. In contrast, $\mathrm{CKA}_{\text{AGR}}$ measures similarity, where a value closer to 1.0 is ideal. Applying the $(1 - \mathrm{G_{kNN}})$ operation inverts the gap metric into a score metric, making 1.0 (indicating zero gap) the ideal score. This aligns its objective with $\mathrm{CKA}_{\text{AGR}}$, ensuring that for the final AGR product, a higher value consistently represents better unlearning performance from a representation perspective.
This formulation balances the impact of both feature divergence and transferability differences, providing a more nuanced measure of unlearning effectiveness from a representation perspective. Finally, we introduce a unified evaluation measure, $H\text{-}LR$, which harmonizes both logit-based (AGL) and representation-based (AGR) metrics:
\begin{equation}
H\text{-}LR = \frac{2}{\frac{1}{AGL} + \frac{1}{AGR}}.
\end{equation}
This harmonic mean formulation ensures that both classification performance and representational differences are jointly considered. Intuitively, a high H-LR score indicates that an unlearning algorithm achieves both output-level forgetting and meaningful representation-level divergence from the original model.


\subsection{Re-evaluation of Unlearning Baselines under Large-Scale Scenarios}
\label{subsec:large_scale}
In this section, we present experimental results for Random Class-wise for in large-scale settings, evaluating both logit-based and representation-based performance.


\begin{wrapfigure}{r}{0.5\textwidth}
    \centering
    \includegraphics[width=0.5\textwidth]{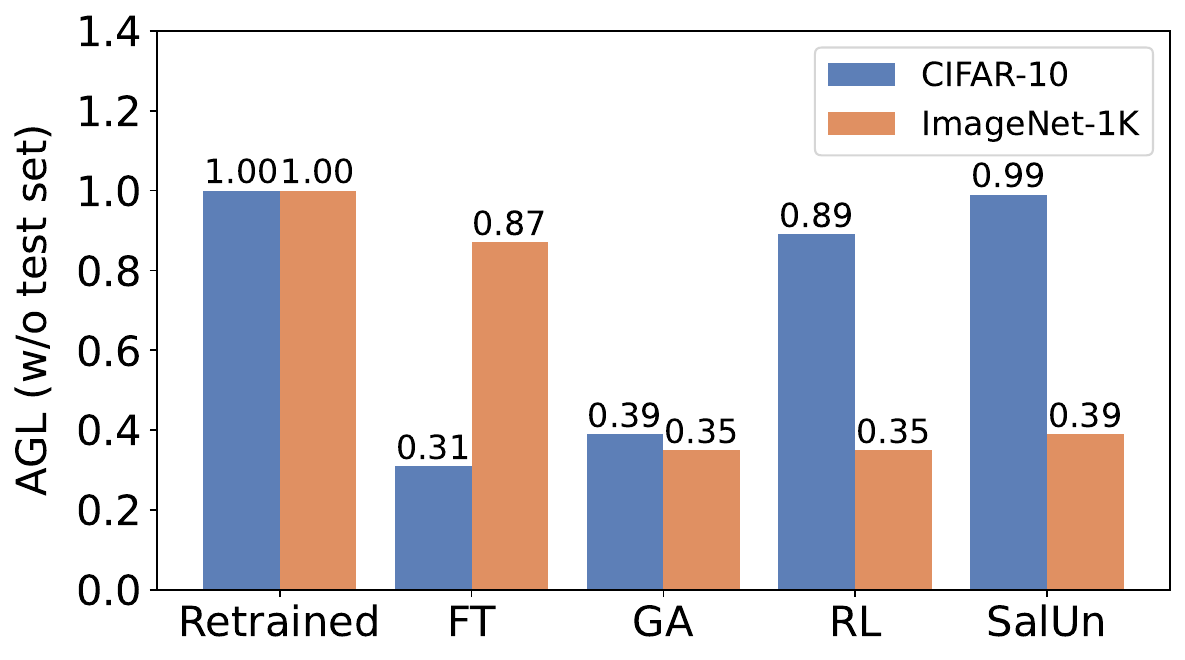}
    \caption{Variation in AGL values between small-scale (CIFAR-10) and large-scale (ImageNet-1K) settings.}
    \label{fig:scale_gap}
\end{wrapfigure}

\noindent\textbf{Small-scale \vs large-scale unlearning.}
Several unlearning algorithms have demonstrated strong performance in small-scale settings, as evaluated by logit-based metrics. As shown in Fig.~\ref{fig:scale_gap}, both RL and SalUn achieve superior AGL on CIFAR-10. However, the figure also clearly highlights a significant performance degradation on ImageNet-1K, underscoring the challenges associated with scaling unlearning algorithms to larger datasets. It emphasizes the limitations of both existing unlearning algorithms and the evaluation methods currently in use.

\begin{table}[t]
\centering
\resizebox{0.75\columnwidth}{!}{
\begin{tabular}{l ccccc c}
\toprule
\multicolumn{1}{c}{} & \multicolumn{5}{c}{\textbf{ImageNet-1K}} & \multicolumn{1}{c}{} \\
\cmidrule(lr){2-6}
\textbf{Method} & \textbf{FA} & \textbf{RA} & \textbf{TFA} & \textbf{TRA} & \textbf{Avg. Gap} & \textbf{AGL} $\uparrow$ \\
\midrule
\textbf{Original}  & 79.2 (\textcolor{blue}{79.2}) & 80.1 (\textcolor{blue}{4.1})  & 76.1 (\textcolor{blue}{76.1}) & 76.5 (\textcolor{blue}{0.9})  & \textcolor{blue}{40.1} & -       \\
\textbf{Retrained} & 0.0 (\textcolor{blue}{0.0})   & 76.0 (\textcolor{blue}{0.0})  & 0.0 (\textcolor{blue}{0.0})   & 75.6 (\textcolor{blue}{0.0})  & \textcolor{blue}{0.0}  & 1.00    \\
\textbf{FT}        & 9.9 (\textcolor{blue}{9.9})   & 79.5 (\textcolor{blue}{3.5})  & 10.5 (\textcolor{blue}{10.5}) & 76.0 (\textcolor{blue}{0.4})  & \textcolor{blue}{6.1}  & 0.78    \\
\midrule
\textbf{GA}        & 1.5 (\textcolor{blue}{1.5})   & 11.4 (\textcolor{blue}{64.6}) & 1.5 (\textcolor{blue}{1.5})   & 11.3 (\textcolor{blue}{64.3}) & \textcolor{blue}{33.0} & 0.12    \\
\textbf{RL}        & 6.3 (\textcolor{blue}{6.3})   & 14.3 (\textcolor{blue}{62.7}) & 5.5 (\textcolor{blue}{5.5})   & 13.0 (\textcolor{blue}{62.6}) & \textcolor{blue}{34.3} & 0.12    \\
\textbf{PL}        & 1.0 (\textcolor{blue}{1.0})   & 79.5 (\textcolor{blue}{3.5})  & 1.0 (\textcolor{blue}{1.0})   & 76.9 (\textcolor{blue}{0.9})  & \textcolor{blue}{1.6}  & 0.94    \\
\textbf{SalUn}     & 10.4 (\textcolor{blue}{10.4}) & 19.7 (\textcolor{blue}{56.3}) & 9.3 (\textcolor{blue}{9.3})   & 19.1 (\textcolor{blue}{56.5}) & \textcolor{blue}{32.8} & 0.15    \\
\midrule
\textbf{DUCK}      & 0.9 (\textcolor{blue}{0.9})   & 74.6 (\textcolor{blue}{1.4})  & 0.9 (\textcolor{blue}{0.9})   & 74.5 (\textcolor{blue}{1.1})  & \textcolor{blue}{\textbf{1.1}}  & \textbf{0.96}    \\
\textbf{CU}        & 2.1 (\textcolor{blue}{2.1})   & 73.9 (\textcolor{blue}{2.1})  & 2.2 (\textcolor{blue}{2.2})   & 73.3 (\textcolor{blue}{2.3})  & \textcolor{blue}{2.2}  & 0.92    \\
\textbf{SCAR}      & 4.2 (\textcolor{blue}{4.2})   & 80.5 (\textcolor{blue}{4.5})  & 3.9 (\textcolor{blue}{3.9})   & 77.4 (\textcolor{blue}{1.8})  & \textcolor{blue}{3.6}  & 0.87    \\
\textbf{SCRUB}     & 1.1 (\textcolor{blue}{1.1})   & 67.3 (\textcolor{blue}{8.7})  & 1.1 (\textcolor{blue}{1.1})   & 65.7 (\textcolor{blue}{9.9})  & \textcolor{blue}{5.2}  & 0.80    \\
\bottomrule
\end{tabular}
}
\caption{Logit-based evaluation on ImageNet-1K. The \textcolor{blue}{blue numbers} represent the gap from \thetar. The \textbf{bold numbers} represent optimal values.}
\label{tab:performance_comparison_part1}
\end{table}

\begin{table}[th]
\centering
\resizebox{\textwidth}{!}{
\large
\begin{tabular}{l ccc ccc c c c}
\toprule
\multicolumn{1}{c}{}
& \multicolumn{3}{c}{\textbf{$k$-NN}}
& \multicolumn{3}{c}{\textbf{CKA(\thetau, \thetar)}}
& \multicolumn{1}{c}{}
& \multicolumn{1}{c}{}
& \multicolumn{1}{c}{\textbf{CKA(\thetau, \thetar) $>$}}
\\
\cmidrule(lr){2-4}
\cmidrule(lr){5-7}
\textbf{Method}
& \textbf{Office-Home} & \textbf{CUB} & \textbf{DomainNet}
& \textbf{Office-Home} & \textbf{CUB} & \textbf{DomainNet}
& \textbf{AGR} $\uparrow$ & \textbf{H-LR} $\uparrow$ & \textbf{CKA(\thetau, \thetao)}
\\
\midrule
\textbf{Original}  & 80.3 (\textcolor{blue}{3.0})  & 43.4 (\textcolor{blue}{6.5})  & 84.0 (\textcolor{blue}{2.0})  & 89.8 (\textcolor{blue}{10.2}) & 82.1 (\textcolor{blue}{17.9}) & 81.8 (\textcolor{blue}{18.2}) & -    & -    & \xmark \\
\textbf{Retrained}& 77.3 (\textcolor{blue}{0.0})  & 36.9 (\textcolor{blue}{0.0})  & 82.0 (\textcolor{blue}{0.0})  & 100.0 (\textcolor{blue}{0.0}) & 100.0 (\textcolor{blue}{0.0}) & 100.0 (\textcolor{blue}{0.0}) & 1.00 & 1.00 & \cmark \\
\textbf{FT}        & 78.4 (\textcolor{blue}{1.1})  & \textbf{39.0 (\textcolor{blue}{2.1})}  & \textbf{81.6 (\textcolor{blue}{0.4})}  & 87.6 (\textcolor{blue}{12.4}) & 79.0 (\textcolor{blue}{21.0}) & 79.9 (\textcolor{blue}{20.1}) & 0.81 & 0.79 & \xmark \\
\midrule
\textbf{GA}        & 29.7 (\textcolor{blue}{47.6}) & 10.4 (\textcolor{blue}{26.5}) & 42.2 (\textcolor{blue}{39.8}) & 8.3 (\textcolor{blue}{91.7})  & 9.8 (\textcolor{blue}{90.2})  & 10.2 (\textcolor{blue}{89.8}) & 0.06 & 0.08 & \xmark \\
\textbf{RL}        & 44.0 (\textcolor{blue}{33.3}) & 8.0 (\textcolor{blue}{28.9})  & 35.2 (\textcolor{blue}{46.8}) & 6.3 (\textcolor{blue}{93.7})  & 5.8 (\textcolor{blue}{94.2})  & 4.7 (\textcolor{blue}{95.3})  & 0.04 & 0.06 & \xmark \\
\textbf{PL}        & 80.3 (\textcolor{blue}{3.0})  & 43.3 (\textcolor{blue}{6.4})  & 84.0 (\textcolor{blue}{2.0})  & \textbf{91.6 (\textcolor{blue}{8.4})}  & \textbf{84.7 (\textcolor{blue}{15.3})} & 84.5 (\textcolor{blue}{15.5}) & 0.84 & 0.89 & \xmark \\
\textbf{SalUn}     & 38.5 (\textcolor{blue}{38.8}) & 8.0 (\textcolor{blue}{28.9})  & 42.5 (\textcolor{blue}{39.5}) & 9.7 (\textcolor{blue}{90.3})  & 8.0 (\textcolor{blue}{92.0})  & 8.5 (\textcolor{blue}{91.5})  & 0.06 & 0.09 & \xmark \\
\midrule
\textbf{DUCK}      & 79.8 (\textcolor{blue}{2.5})  & \textbf{39.0 (\textcolor{blue}{2.1})}  & 82.5 (\textcolor{blue}{0.7})  & 90.7 (\textcolor{blue}{9.3})  & 83.2 (\textcolor{blue}{16.8}) & \textbf{84.9 (\textcolor{blue}{15.1})} & \textbf{0.85} & \textbf{0.90} & \xmark \\
\textbf{CU}        & 75.8 (\textcolor{blue}{1.5})  & 33.9 (\textcolor{blue}{3.0})  & 80.1 (\textcolor{blue}{1.9})  & 85.6 (\textcolor{blue}{14.4}) & 75.9 (\textcolor{blue}{24.1}) & 76.1 (\textcolor{blue}{23.9}) & 0.78 & 0.84 & \xmark \\
\textbf{SCAR}      & \textbf{78.0 (\textcolor{blue}{0.7})}  & 42.8 (\textcolor{blue}{5.9})  & 83.1 (\textcolor{blue}{1.1})  & 74.2 (\textcolor{blue}{25.8}) & 65.4 (\textcolor{blue}{34.6}) & 58.8 (\textcolor{blue}{41.2}) & 0.64 & 0.74 & \xmark \\
\textbf{SCRUB}     & 74.7 (\textcolor{blue}{2.6})  & 42.2 (\textcolor{blue}{5.3})  & 80.9 (\textcolor{blue}{1.1})  & 68.9 (\textcolor{blue}{51.1}) & 63.8 (\textcolor{blue}{36.2}) & 52.8 (\textcolor{blue}{47.2}) & 0.60 & 0.69 & \xmark \\
\bottomrule
\end{tabular}
}
\caption{Representation-based evaluation across different downstream datasets. The \textcolor{blue}{blue numbers} represent the gap from \thetar. The \textbf{bold numbers} represent optimal values.}
\label{tab:performance_comparison_part2}
\vspace{-4mm}
\end{table}

\noindent\textbf{Experimental results in large-scale unlearning.} \ \
Table~\ref{tab:performance_comparison_part1} presents the logit-based evaluation results for all baselines and evaluation methods in our large-scale unlearning scenarios. Through logit-based evaluation and AGL scores, we observe that PL (which does not use \datar) achieves performance comparable to, or even better than, state-of-the-art algorithms that do utilize \datar (\ie, DUCK, CU, SCAR, and SCRUB). This result marks a notable departure from the trends observed in small-scale unlearning experiments~\citep{cotogni2023duck,Mahadevan2021CertifiableMU,bonato2024retain,kurmanji2023towards}. Table~\ref{tab:performance_comparison_part2} presents the representation-based evaluation. In representation-based evaluation (measured by AGR), PL consistently outperforms other algorithms, achieving superior H-LR among all baselines. These results challenge prior expectations, showing that many state-of-the-art algorithms degrade in large-scale scenarios.

\subsection{Experiments on Top Class-wise Forgetting}
\label{subsec:top_class}

\begin{wrapfigure}{r}{0.45\textwidth}
    \centering
    \includegraphics[width=0.45\textwidth]{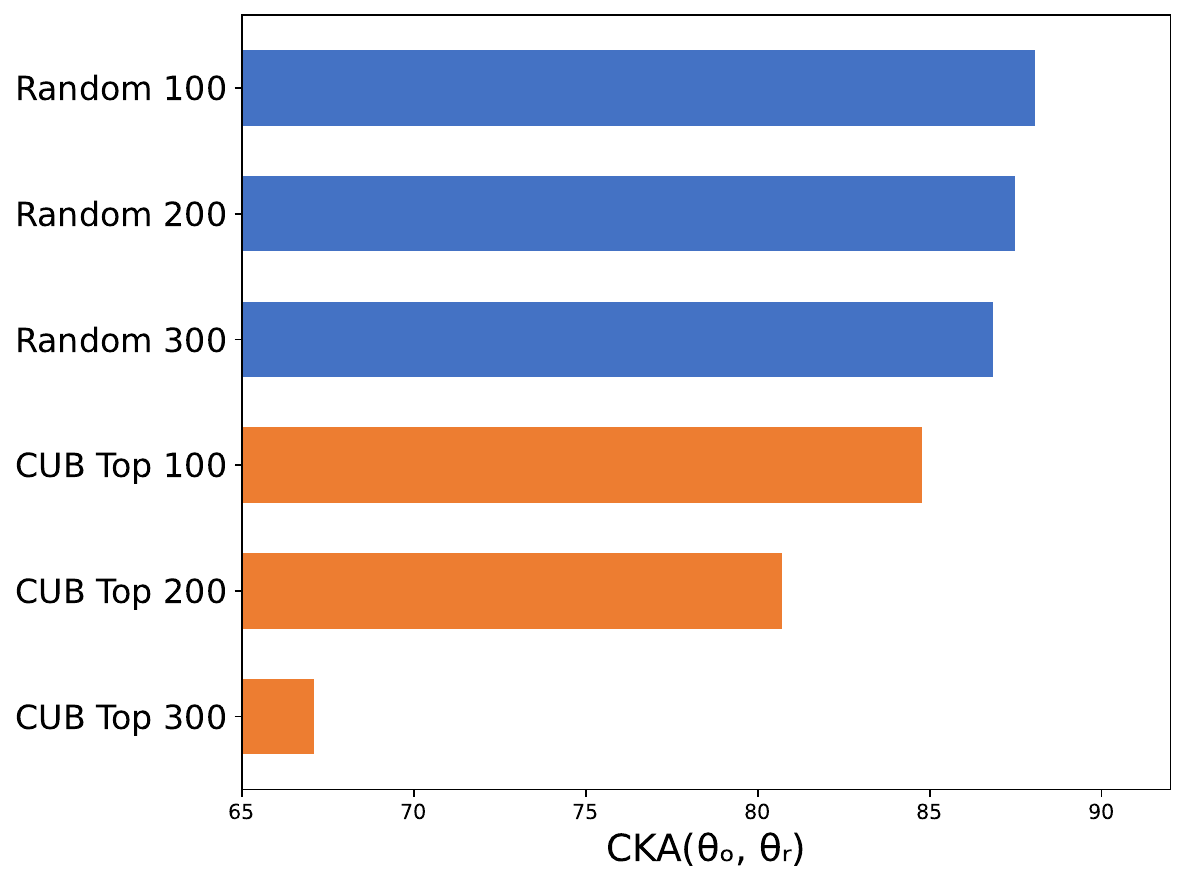}
    \vspace{-4mm}
    \caption{CKA similarities between \thetao and \thetar across various unlearning scenarios. The similarities decrease under Top Class-wise Forgetting compared to Random Class-wise Forgetting and with increasing numbers of forgetting classes.}
    \label{fig:top_necessary}
    \vspace{-4mm}
\end{wrapfigure}

\noindent\textbf{Why Top Class-wise Forgetting is necessary?} \ \
In the Random Class-wise Forgetting scenario, the CKA similarity between \thetao and \thetar reaches a high value of 88\%, as shown in Fig.~\ref{fig:maintain}. We hypothesize that this high similarity arises from the presence of multiple alternative classes within ImageNet-1K, which may preserve similar representational knowledge. This results in a significant overlap between both models, presenting challenges in verifying the effectiveness of unlearning from a representational perspective. To address this, we propose a \textit{Top Class-wise Forgetting} scenario, where forgetting classes are selected based on their semantic similarity to the downstream dataset classes. This ensures that no alternative classes with similar representations remain in \datar, leading to a substantial difference between \thetar and \thetao in learned representation. In other words, this makes unlearning more challenging and allows for a more rigorous evaluation. As shown in Fig.~\ref{fig:top_necessary}, this scenario minimizes CKA similarity between \thetao and \thetar, thereby facilitating a more rigorous evaluation. Consequently, removing more similar classes results in a more stringent evaluation, enabling a more reliable assessment.

\noindent\textbf{Similarity-based Top-class selection.} \ \
In the Top-N Class-wise Forgetting scenario, we select N forgetting classes that are most similar to the classes in the downstream datasets. To measure similarity, we extract feature maps for each class from the encoder (\eg, average pooling layer of ResNet-50) and compute cosine similarity between the class-wise averaged feature maps of two datasets. We rank the classes and select the top N most similar ones for the forgetting classes. For instance, in the case of Office-Home Top-100 Class-wise Forgetting, where the dataset consists of 65 classes, we identify the most similar ImageNet-1K classes to Office-Home as the forgetting classes. We additionally select the next 35 most similar classes, creating a 100 class forget set.

\begin{figure}[t]
    \centering
    \begin{subfigure}[b]{0.24\textwidth}
        \centering
        \includegraphics[width=\linewidth]{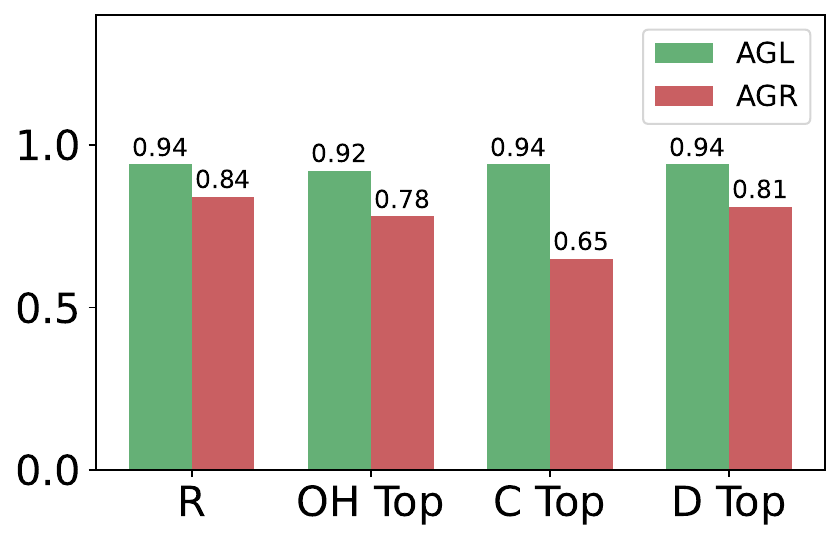}
        \vspace{-4mm}
        \caption{PL}
        \label{fig:aglagr_gap_pl}
    \end{subfigure}
    \begin{subfigure}[b]{0.24\textwidth}
        \centering
        \includegraphics[width=\linewidth]{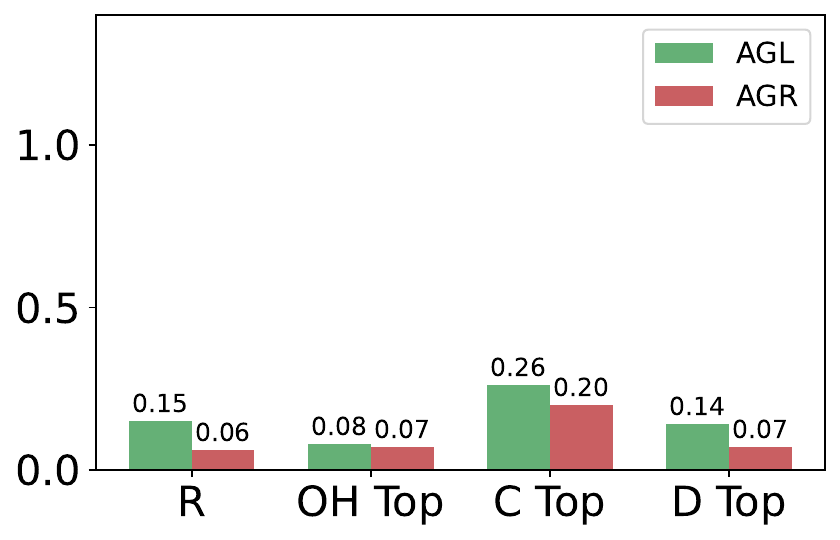}
        \vspace{-4mm}
        \caption{SalUn}
        \label{fig:aglagr_gap_salun}
    \end{subfigure}
    \begin{subfigure}[b]{0.24\textwidth}
        \centering
        \includegraphics[width=\linewidth]{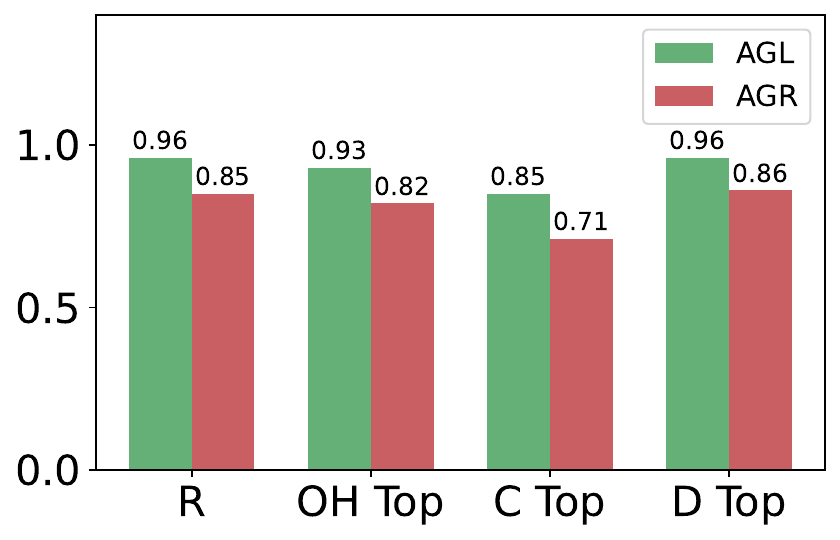}
        \vspace{-4mm}
        \caption{DUCK}
        \label{fig:aglagr_gap_duck}
    \end{subfigure}
    \begin{subfigure}[b]{0.24\textwidth}
        \centering
        \includegraphics[width=\linewidth]{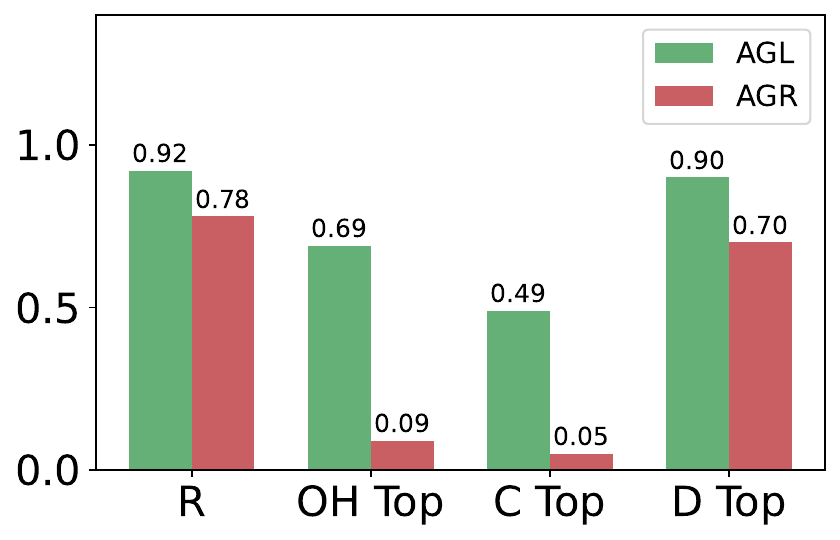}
        \vspace{-4mm}
        \caption{CU}
        \label{fig:aglagr_gap_cu}
    \end{subfigure}
    \vspace{-2mm}
    \caption{AGL and AGR comparison on unlearning baselines, using Random (R), Office-Home (OH), CUB (C), and DomainNet-126 (D) to select 100 similar classes for Top Class-wise Forgetting.}
    \label{fig:aglagr_gap}
\end{figure}

\noindent\textbf{Random \vs Top Class-wise Forgetting.} \ \ 
We conduct experiments comparing Random Class-wise Forgetting and Top Class-wise Forgetting across downstream datasets. The results are presented in Fig.~\ref{fig:aglagr_gap}. An ideal unlearning algorithm should exhibit consistent relative gaps between AGL and AGR, regardless of the scenario. This indicates that the unlearning algorithm achieves relatively stable unlearning performance across different scenarios. As shown in Fig.~\ref{fig:aglagr_gap}, PL and DUCK have consistent gaps across all settings. However, CU shows a drop in AGR scores specifically under the Top Class-wise Forgetting scenario, despite maintaining reasonably good AGL scores. This result highlights its weaknesses under Top Class-wise Forgetting and underscores the importance of including the Top Class-wise Forgetting scenario for rigorous evaluation. Additionally, SalUn shows considerably degraded performance in terms of both AGL and AGR in all scenarios.

\begin{figure}[t]
    \centering
    \begin{subfigure}[b]{0.48\textwidth}
        \centering
        \includegraphics[width=\linewidth]{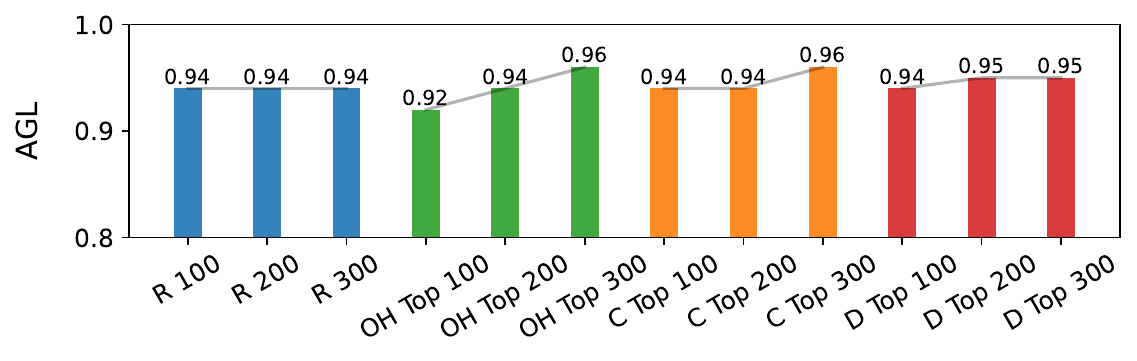}
        \vspace{-4mm}
        \caption{AGL scores of PL}
        \vspace{-1mm}
        \label{fig:increasing_agl}
    \end{subfigure}
    \begin{subfigure}[b]{0.48\textwidth}
        \centering
    \includegraphics[width=\linewidth]{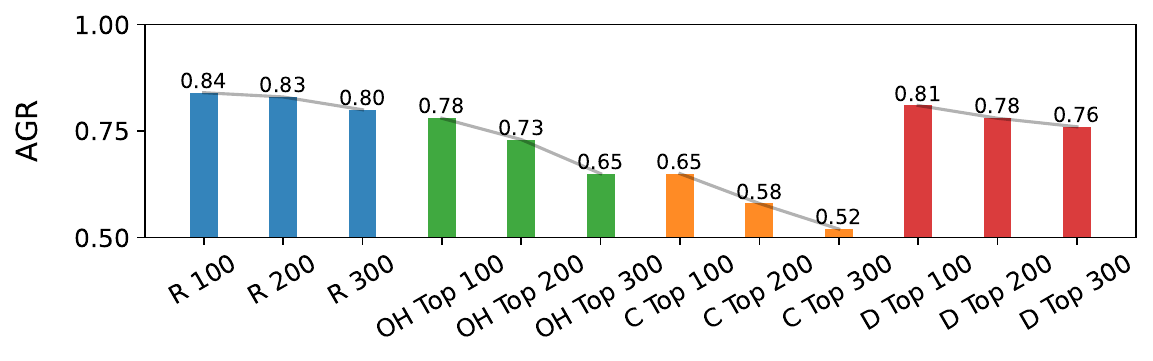}
        \vspace{-4mm}
        \caption{AGR scores of PL}
        \vspace{-1mm}
        \label{fig:increasing_agr}
    \end{subfigure}
    \begin{subfigure}[b]{0.48\textwidth}
        \centering
        \includegraphics[width=\linewidth]{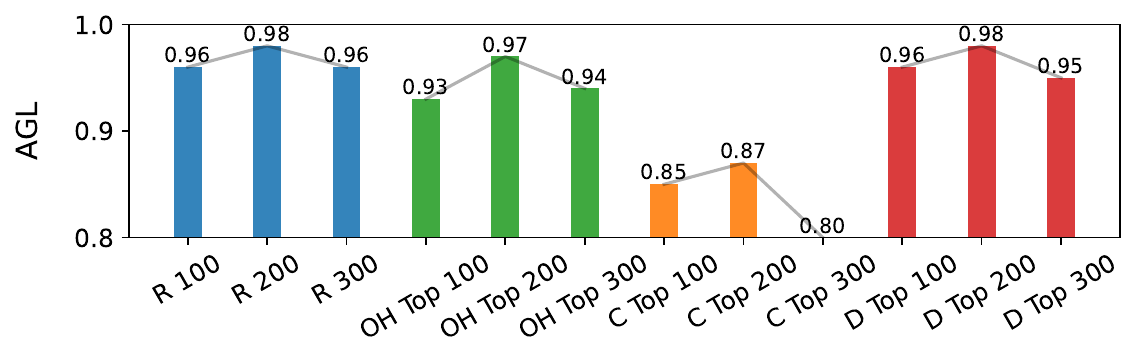}
        \vspace{-4mm}
        \caption{AGL scores of DUCK}
    \end{subfigure}
    \begin{subfigure}[b]{0.48\textwidth}
        \centering
        \includegraphics[width=\linewidth]{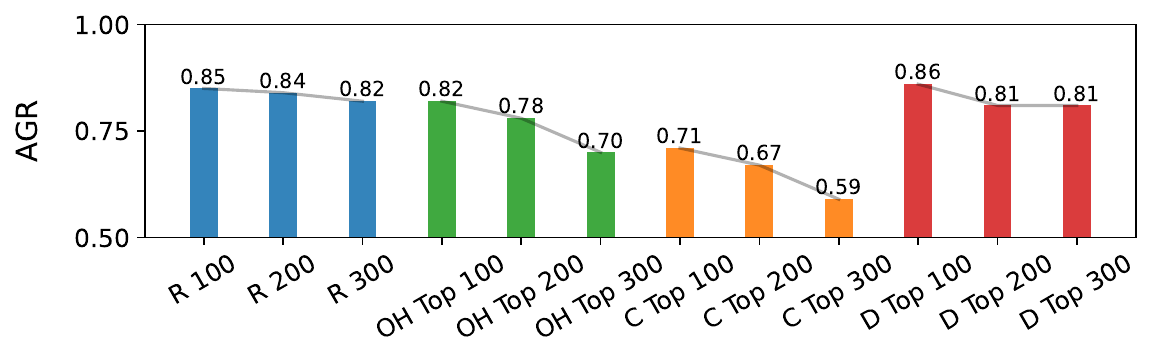}
        \vspace{-4mm}
        \caption{AGR scores of DUCK}
    \end{subfigure}
    \vspace{-1mm}
    \caption{AGL and AGR scores on PL and DUCK under Random and Top Class-wise Forgetting scenarios. (a) AGL scores are consistently high in all settings. (b) AGR scores decrease, as the number of forgetting classes increases. Similar trends are observed for the DUCK dataset ((c), (d)). This is evident in Top Class-wise Forgetting, where \thetar has a low similarity to \thetao.}
    \vspace{-2mm}
    \label{fig:increasing}
\end{figure}

\noindent\textbf{Need for considering both AGL and AGR, respectively.} \ \
Fig.~\ref{fig:aglagr_gap} reveals that unlearning algorithms exhibit unexpected behaviors, which become evident when considering both AGR and AGL scores. Despite PL and DUCK often showing little disparity between these scores, there is an inherent need to assess these measures independently to fully understand the performance of the unlearning algorithms. To demonstrate this, we extended the evaluation of PL by conducting experiments on 100 to 300 Top Class-wise Forgetting scenarios. As the number of forgetting classes increases, particularly in Top Class-wise Forgetting, the nearest classes diverge significantly from the original ones. Fig.~\ref{fig:increasing} illustrates that AGL scores remain largely unaffected by the number of forgetting classes. However, AGR scores notably decrease as the number of forgotten classes increases, indicating that PL and DUCK still retain substantial similarity to \thetao’s representation. These findings not only highlight the discrepancy between the unlearning performed by these algorithms and the expected results but also emphasize the need for our proposed multi-faceted evaluation approach.

\begin{figure}[t]
    \centering
    \includegraphics[width=1.0\linewidth]{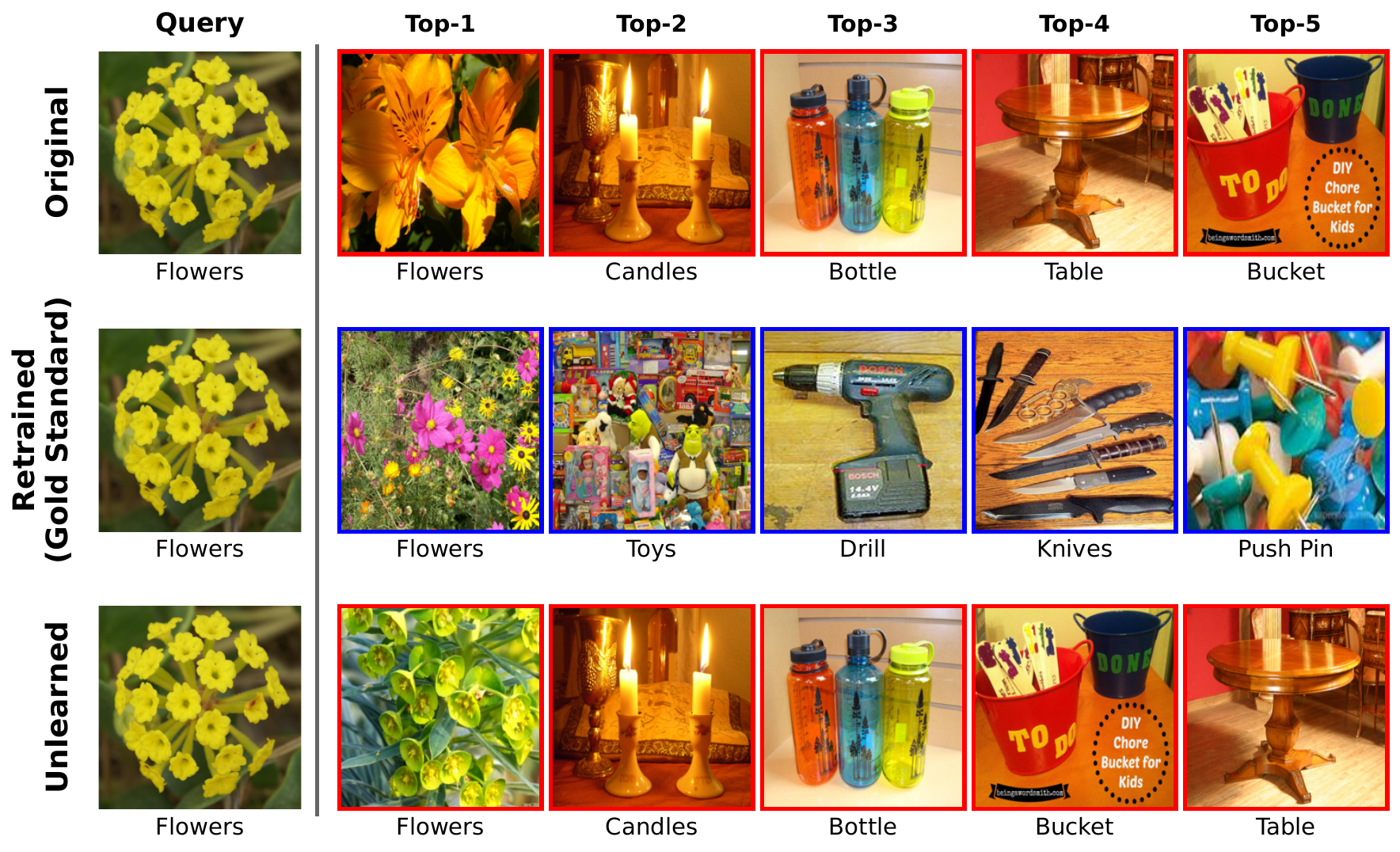}
    \caption{Qualitative comparison of $k$-NN retrievals for the query image across the original, retrained, and unlearned (PL) models in the Office-Home Top-100 setting. The nearest retrieved classes of the unlearned model resemble those of the original model more closely than those of the retrained model.}
    \label{fig:knn_qualitative}
\end{figure}

\noindent\textbf{Qualitative analysis via $k$-NN visualization.} \ \

To provide an intuitive understanding of the representational changes, we further conducted a qualitative analysis by visualizing the $k$ nearest neighbors. We specifically examined the Pseudo Labeling (PL) method under the Office-Home Top-100 Class-wise Forgetting scenario. Using the same query image from the Office-Home dataset that semantically corresponds to the forget set, Fig.~\ref{fig:knn_qualitative} illustrates the top-5 nearest neighbors retrieved by the original ($\theta_o$), retrained ($\theta_r$), and unlearned ($\theta_u$) models. Consistent with our quantitative findings, the unlearned model retrieves neighbors that are highly similar to those of the original model. In contrast, the retrained model retrieves a distinct set of neighbors, reflecting a significant shift in the feature space. These results visually corroborate that current unlearning algorithms fail to fundamentally alter the learned representations to match the gold standard (\ie, the retrained model), instead preserving the original model's representations.

\subsection{Additional Discussions}

\begin{figure}[t]
    \centering
    \begin{subfigure}[b]{0.48\textwidth}
        \centering
        \includegraphics[width=\linewidth]{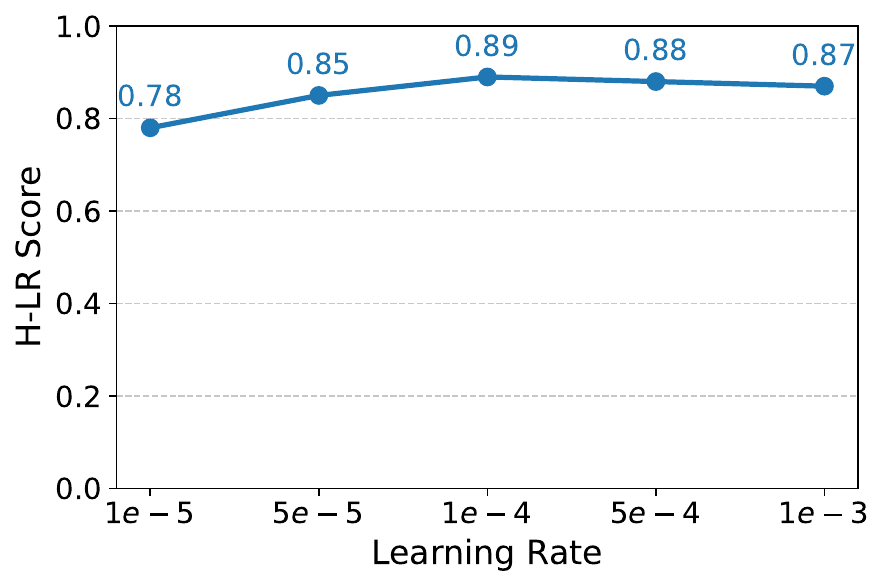}
        \caption{Learning Rate Sensitivity of PL}
    \end{subfigure}
    \hfill
    \begin{subfigure}[b]{0.48\textwidth}
        \centering
        \includegraphics[width=\linewidth]{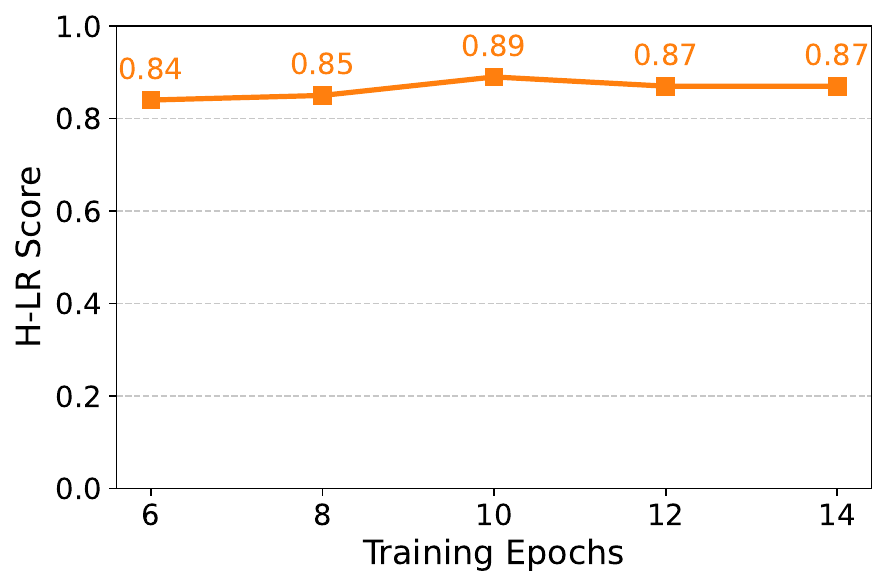}
        \caption{Training Epochs Sensitivity of PL}
    \end{subfigure}
    \vspace{4mm}
    \begin{subfigure}[b]{0.48\textwidth}
        \centering
        \includegraphics[width=\linewidth]{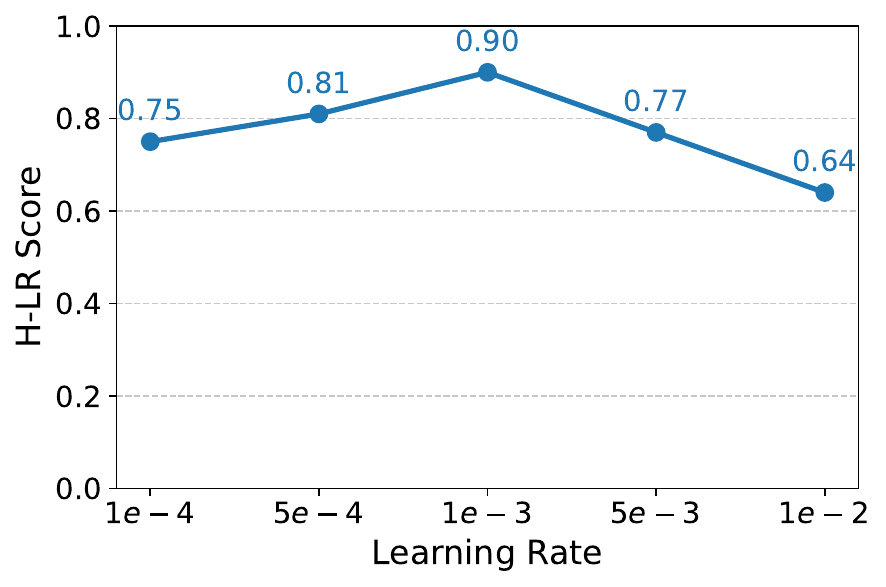}
        \caption{Learning Rate Sensitivity of DUCK}
    \end{subfigure}
    \hfill
    \begin{subfigure}[b]{0.48\textwidth}
        \centering
        \includegraphics[width=\linewidth]{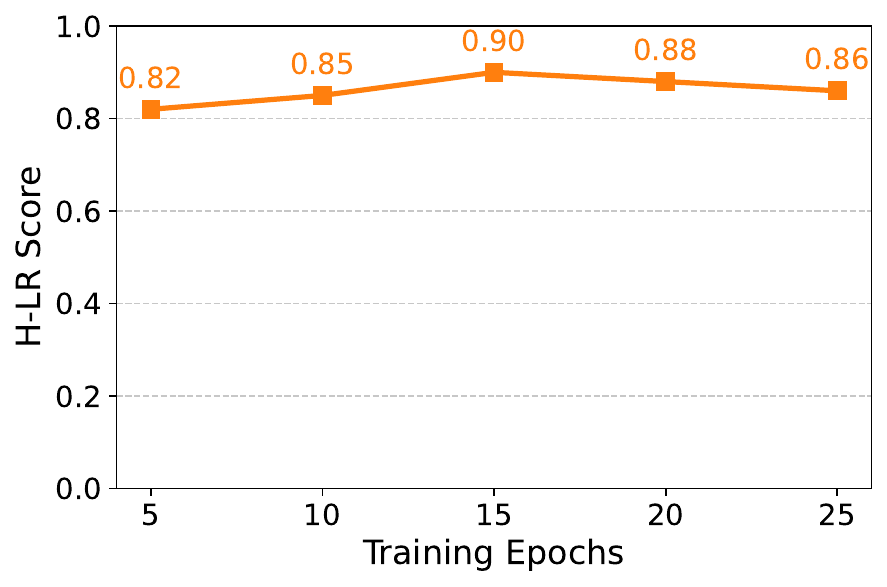}
        \caption{Training Epochs Sensitivity of DUCK}
    \end{subfigure}
    \caption{Hyperparameter sensitivity analysis for representative baselines. We visualize the impact of learning rate and training epochs on the H-LR score. The results demonstrate that performance is stable around the optimal values we selected, providing a clear justification for our experimental configurations.}
    \label{fig:sensitivity_analysis}
\end{figure}

\noindent\textbf{Analysis of hyperparameter sensitivity.} \ \
We visualize the impact of key hyperparameters (learning rate and number of epochs) on the H-LR score for representative baseline algorithms, PL and DUCK. As shown in Figure~\ref{fig:sensitivity_analysis}, the plots demonstrate the stability of the algorithms near their chosen parameters. This analysis provides a clear justification for the optimal configurations used in our main evaluation.

\noindent\textbf{Analysis of algorithmic failure cases.} \ \
Our analysis in Section 3.7 revealed that many unlearning algorithms achieve high logit-based scores while failing to alter their internal representations, which remain highly similar to the original model ($\theta_o$). In this section, we provide a deeper analysis of the root causes for these failures, hypothesizing two primary modes.
\indent First, we posit that the common \qq{classifier-only} modification, empirically demonstrated in Figure 4b, is a form of \textit{shortcut learning}. The optimization landscapes of current unlearning objectives appear to offer a \qq{path of least resistance}. Modifying the final classifier's weights is a much simpler optimization task than performing the complex, non-linear reorganization required to fundamentally alter the encoder's feature representations. Because logit-based metrics (\eg, accuracy) are the primary objective, the optimization converges to this sub-optimal solution, satisfying the metrics without achieving true representational unlearning.
\indent Second, we analyze the catastrophic failures observed in some baselines, such as the representational collapse of GA (Fig. 3d) and the degradation of SCRUB in our Top Class-wise Forgetting scenarios. We hypothesize this is due to \textit{unbounded loss functions}. Methods like GA explicitly maximize the loss on the forget set. Similarly, SCRUB's objective includes maximizing the distance from the forget set's teacher. This type of unbounded objective can lead to optimization instability, or a \qq{gradient explosion}, especially when applied to large-scale unlearning (as in our Top Class-wise scenario). This instability does not surgically erase information but instead catastrophically collapses the entire feature space.

\noindent\textbf{Analysis of computational cost.} \ \
To address practical concerns, we analyzed the computational overhead of our unified benchmark in Table~\ref{tab:computational_cost}. The primary cost of traditional evaluation (AGL) stems from inference on the large-scale ImageNet-1K dataset. Our proposed AGR metric, in contrast, only requires inference on significantly smaller downstream datasets (\eg, CUB, which is about 1\% the size of ImageNet-1K). As shown in the table, the additional time required for CKA and $k$-NN calculations on these small datasets is minimal. Therefore, our framework provides a much more comprehensive and rigorous evaluation with only a marginal increase in computational overhead, ensuring its practicality for researchers in resource-constrained environments.

\begin{table}[ht]
\centering
\resizebox{0.7\columnwidth}{!}{%
\begin{tabular}{lcccc} 
\toprule
\multicolumn{1}{c}{} & \multicolumn{3}{c}{\textbf{Representation-based}} & \multicolumn{1}{c}{\textbf{Logit-based}} \\
\cmidrule(lr){2-4} \cmidrule(lr){5-5}
\textbf{Metric} & \textbf{CKA} & \textbf{$k$-NN} & \textbf{Total (AGR)} & \textbf{AGL} \\
\midrule
\textbf{Cost (TFLOPs)} & 50.58 & 23.83 & \textbf{74.41} & 5477.46 \\
\textbf{Memory (GB)}   & 1.52 & 1.58 & \textbf{3.1} & 5.47     \\
\textbf{Time (second)}      & 20.82 & 17.07 & \textbf{37.89} & 1002.18 \\
\bottomrule
\end{tabular}
}
\caption{Computational cost comparison of our evaluation framework and the traditional evaluation framework. Costs are measured under the CUB Top-100 Class-wise Forgetting scenario. The results highlight that our proposed representation-based framework under CUB Top-100 Class-wise Forgetting scenario incurs marginal computational overhead compared to the standard logit-based evaluation.}
\label{tab:computational_cost}
\end{table}

\noindent\textbf{Analysis of Membership Inference Attack (MIA) and time efficiency.} \ \
We acknowledge that output-based evaluations encompass not only logit-based metrics but also attack-based families, such as Membership Inference Attacks (MIA). We report the results of MIA efficacy in Table~\ref{tab:mia}. These results indicate that most unlearning methods (with the exception of RL) achieve high MIA scores, suggesting successful unlearning from this specific perspective. However, we argue that this demonstrates a limitation of using MIA as a primary metric in our class-wise evaluation setting. MIA is designed to evaluate the \qq{member} or \qq{non-member} status of individual samples based on output confidence, making it highly appropriate for sample-wise unlearning. In our class-wise setting, the objective is to verify the removal of an entire class's feature distribution. A confidence-based metric designed for single samples may not be the most suitable tool for this task, further reinforcing the need for the representation-based evaluations proposed in our benchmark.
\\
Beyond theoretical effectiveness, we measure the running time of the unlearning method, recorded in minutes, to evaluate its time efficiency~\citep{fan2024salun}. As expected, all approximate unlearning methods are significantly faster than the gold standard (4666.7 min). However, the results reveal a clear distinction: methods that operate only on the forget set, such as GA, RL, and PL (24-35 min), are orders of magnitude more efficient than methods that also leverage the retain set, such as DUCK, CU, and SCRUB (330-913 min). This is because the latter must process a much larger volume of retain data to prevent catastrophic forgetting. This highlights a critical trade-off between unlearning speed and performance, which must be considered for real-world deployment, especially in large-scale scenarios.

\begin{table}[t]
\centering
\resizebox{0.35\textwidth}{!}{
\begin{tabular}{lcc}
\toprule
\textbf{Method}
& \textbf{MIA}
& \textbf{RTE}
\\
\midrule
\textbf{Retrain} 
& 100.0 (\textcolor{blue}{0.0})
& 4666.7
\\
\textbf{FT} 
& 87.0 (\textcolor{blue}{13.0})
& 1200.7
\\
\midrule
\textbf{GA} 
& 98.0 (\textcolor{blue}{2.0})
& 26.7
\\
\textbf{RL} 
& 8.0 (\textcolor{blue}{92.0})
& \textbf{24.7}
\\
\textbf{PL}
& \textbf{99.0 (\textcolor{blue}{1.0})}
& 30.1
\\
\textbf{SalUn} 
& 88.0 (\textcolor{blue}{12.0})
& 34.3
\\
\midrule
\textbf{DUCK} 
& \textbf{99.0 (\textcolor{blue}{1.0})}
& 913.6
\\
\textbf{CU} 
& 97.0 (\textcolor{blue}{3.0})
& 568.4
\\
\textbf{SCAR} 
& 95.0 (\textcolor{blue}{5.0})
& 566.7
\\
\textbf{SCRUB} 
& \textbf{99.0 (\textcolor{blue}{1.0})}
& 330.0
\\
\bottomrule
\end{tabular}
}
\caption{The comparison of MIA and RTE (Run Time Efficiency) across various methods on the Random-100 Class-wise Forgetting scenario. The \textcolor{blue}{blue numbers} indicate the gap relative to the retrained model. The \textbf{bold numbers} represent optimal values.}
\label{tab:mia}
\end{table}

\noindent\textbf{Extension to other domains.} \ \
While our framework's primary contribution is establishing a rigorous evaluation protocol for image classification, its principles connect to broader research frontiers in privacy, fairness, and emerging domains like federated and graph unlearning \citep{liu2024survey, zhang2025dynamic}. Our analysis reveals a critical gap between logit-based and representation-based evaluation, a challenge that is not unique to the image classification task. We believe extending our framework's principles to Graph Neural Networks (GNNs), where privacy-preserving inference is also a key concern \citep{xu2024oblivgnn}, is a valuable future research direction. For example, our \qq{Top Class-wise Forgetting} scenario could be adapted into a \qq{GNN Top-subgraph Forgetting} scenario. While such a substantial extension is beyond the scope of our current manuscript, it highlights a promising path for future investigation.

\begin{figure}[t]
    \centering
    \includegraphics[width=0.7\textwidth]{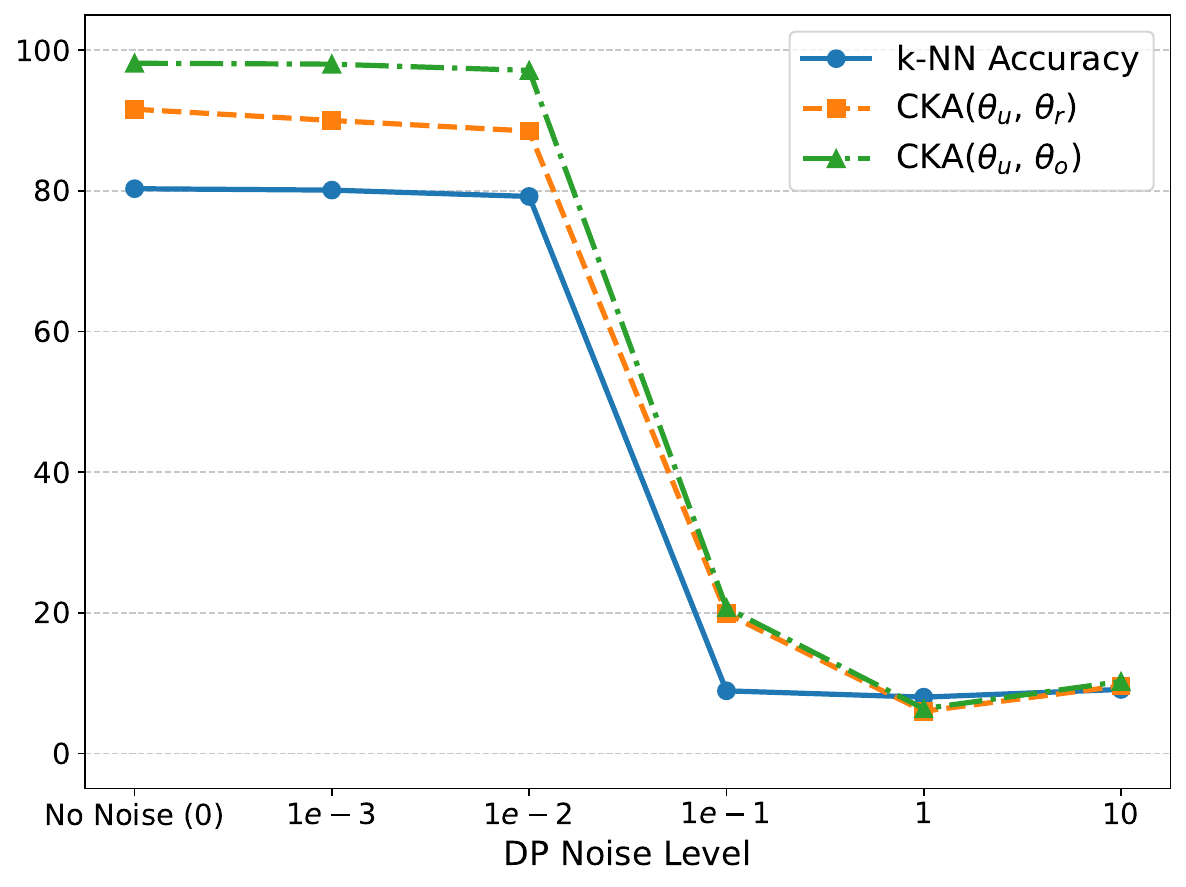}
    \caption{The trade-off between DP noise level and representational quality. We apply varying levels of noise (from 'No Noise' to $\sigma=10$) to the PL unlearning method during the unlearning process. Both $k$-NN Accuracy and CKA values show a sharp cliff, collapsing when the noise level reaches $1e-2$. This demonstrates that a stricter privacy guarantee can severely degrade the model's representational quality.}
    \label{fig:dp_tradeoff}
\end{figure}

\noindent\textbf{Analysis of the trade-off with differential privacy.} \ \
A critical consideration for real-world deployment is the interplay between unlearning effectiveness and formal privacy guarantees like Differential Privacy (DP)~\citep{jiang2024efficient}. While our benchmark measures the former, DP provides a provable guarantee about the process itself. A key question is: how does enforcing a DP guarantee impact the model's representational quality? To investigate this, we conducted an empirical study on the PL unlearning method under the Random-100 Class-wise Forgetting scenario. We simulated the application of DP by injecting varying levels of Gaussian noise during the unlearning process. We then evaluated these models using our representation-based metrics: $k$-NN Accuracy, CKA similarity to the retrained model ($CKA(\theta_u, \theta_r)$), and CKA similarity to the original model ($CKA(\theta_u, \theta_o)$) on the Office-Home dataset. As shown in Figure~\ref{fig:dp_tradeoff}, the results reveal a trade-off. For low noise levels ($\sigma \le 1e-2$), the model maintains high representational quality. The $k$-NN Accuracy and both CKA similarity scores remain nearly identical to the non-private ($\sigma = 0$) baseline. However, as the noise level increases to $\sigma = 1e-1$ to provide a stronger privacy guarantee, all three metrics collapse simultaneously, indicating a catastrophic loss of the model's representational quality. This validates that a trade-off exists between provable privacy and model utility, showing that applying DP noise does not simply erase the forget data (lowering $CKA(\theta_u, \theta_o)$) but can destroy the entire feature space (lowering $k$-NN and $CKA(\theta_u, \theta_r)$) if not calibrated carefully.

\section{Concluding Remarks}
\subsection{Main Contributions}
We propose a unified benchmark to evaluate machine unlearning under realistic, large-scale conditions. Our framework incorporates representation-based metrics and a novel Top Class-wise Forgetting scenario to reveal the limitations of current methods beyond logit-based metrics. We show that many algorithms fail to alter internal representations, despite showing high logit-level scores. Our benchmark offers a scalable and thorough evaluation protocol, laying the groundwork for more rigorous and reliable research.

\subsection{Limitations}
\noindent\textbf{Task limitation to classification.} \ \
Our benchmark is primarily designed for classification tasks with accessible encoders and is not directly applicable to generative or black-box models. CKA and $k$-NN metrics, while informative, may not capture all semantic shifts. Future extensions should address dynamic unlearning and broader task types. 

\noindent\textbf{Limitation to class-wise unlearning.} \ \
Furthermore, our framework is designed for class-wise unlearning. Evaluating sample-wise unlearning presents a distinct challenge, as the removal of a single data point is unlikely to cause a statistically significant shift in the model's overall representation manifold. This makes distribution-level metrics like CKA insufficiently sensitive. Notably, logit-based evaluations face the same limitation, as task-level accuracy is also ill-suited for verifying single-sample removal. Designing a robust framework for sample-wise unlearning thus remains an important open research challenge.

\noindent\textbf{Reliance on the retrained model as gold standard.} \ \
Our framework's reliance on the retrained model as the gold standard is a common practice in the unlearning literature. However, retraining the model can be computationally burdensome. We believe that the CKA similarity between the unlearned model and the original model, $CKA(\theta_u, \theta_o)$, could serve as a proxy to estimate the degree of unlearning in the absence of a fully retrained gold standard. However, we acknowledge that establishing a robust evaluation protocol that does not rely on an oracle model remains a significant open challenge.

\subsection{Broader Impact}
Our benchmark enables a more accountable and rigorous evaluation of unlearning algorithms, which is critical for privacy-sensitive applications such as healthcare or personalized AI. However, there is a risk that future methods might overfit to benchmark metrics without achieving true forgetting. We advocate combining our benchmark with formal privacy guarantees and adversarial robustness checks to ensure safe deployment.

\appendix

\section{Implementation Details}
\noindent\textbf{Training and retraining details.} \ \
\label{training_and_retraining_details}

For the experiments, both the original and retrained models are required. In the case of ResNet-50, we trained it from scratch. For ConvNeXt-T and Swin-T, we initialized them with ImageNet-21K pre-trained weights, following~\cite{jeon2024information}. Detailed configuration information is provided in Table~\ref{tab:training_details}.

\begin{table}[ht]
    \centering
    \resizebox{0.6\columnwidth}{!}{%
    \begin{tabular}{lccc}
        \toprule
        \multicolumn{1}{c}{}
        & \multicolumn{3}{c}{\textbf{ImageNet-1K}} \\
        \cmidrule(lr){2-4}
        \textbf{Settings} & \multicolumn{1}{c}{\textbf{ResNet-50}} & \multicolumn{1}{c}{\textbf{ConvNeXt-T}} & \multicolumn{1}{c}{\textbf{Swin-T}} \\
        \midrule
        \textbf{Epochs} & \multicolumn{1}{c}{182} & \multicolumn{1}{c}{30} & \multicolumn{1}{c}{30} \\
        \textbf{Batch Size} & \multicolumn{3}{c}{256} \\
        \textbf{Learning Rate} & \multicolumn{3}{c}{0.1} \\
        \textbf{Optimizer} & \multicolumn{1}{c}{SGD} & \multicolumn{2}{c}{AdamW}  \\
        \textbf{Momentum} & \multicolumn{3}{c}{0.9} \\
        \textbf{Scheduler} & \multicolumn{1}{c}{Step} & \multicolumn{2}{c}{CosineAnnealing} \\
        \bottomrule
    \end{tabular}%
    }
    \caption{Training details for original and retrain models.}
\label{tab:training_details}
\end{table}

\noindent\textbf{Unlearning details.} \ \
\label{unlearning_details}
We refer to the hyperparameters from the original studies. However, because most prior works proposed and tested these methods on smaller datasets (e.g., CIFAR-10, CIFAR-100), critical implementation details are not specified when removing multiple classes from large-scale datasets. To cope with this, we perform a grid search for each method to identify the optimal learning rate and the number of epochs. Specifically, we retrain the model (excluding the forget set) using the same learning rate and number of epochs as in the original training for the Retrain method. We limit FT to 40 epochs to mitigate excessive time costs. For PL, GA, RL, and SalUn, we find that strong unlearning can be achieved within 10 epochs on large-scale data. Finally, considering the balance of logit-based and representation-based metrics, we fix the maximum epochs for DUCK, CU, SCAR, and SCRUB at 15, 80, 20, and 90, respectively. These upper limits are first determined under the scenario of random 100 class unlearning. For Random-N class experiments (\(N \in \{100,200,300\}\)), we choose the actual number of epochs while respecting these maximum values and are guided by the H-LR score. For Top-N class unlearning, we reuse the hyperparameter configurations established in the corresponding Random-N setting. All experiments were conducted using a single NVIDIA RTX 4090 GPU.

\section{Additional Experimental Results}
\label{appendix:additional_experiments}
Our comprehensive results in all unlearning baselines across all unlearning scenarios can be found in Tables~\ref{tab:random}-\ref{tab:all_h-lr}.

\noindent\textbf{Different backbone architectures.} \ \
Table~\ref{tab:backbone_architecture} provides an extended comparison of the baselines using alternative backbones, specifically ConvNeXt-T \citep{liu2022convnext} and Swin-T \citep{liu2021swin}. While our primary analysis was conducted using ResNet-50 (as it is the standard backbone in the unlearning literature), we included these modern architectures to confirm that our benchmark's findings are robust and not tied to a specific architectural design. As the results in Table~\ref{tab:backbone_architecture} demonstrate, the similar trends observed with ResNet-50 re-emerge. Notably, the limitations of certain methods (\eg, GA, RL, and SalUn showing poor representation-level unlearning) and the relative strengths of methods like PL and CU hold across these different backbones. This consistency confirms that our framework is capable of reliably assessing unlearning effectiveness across diverse architectures, demonstrating that the identified issues (\eg, the gap between logit-based and representation-based forgetting) are fundamental to the unlearning algorithms themselves and not an artifact of a single backbone. We compare only model-agnostic methods.

\begin{table}[ht]
\centering
\resizebox{\textwidth}{!}{
\begin{tabular}{lccccccccccc}
\toprule
\multicolumn{2}{c}{} 
& \multicolumn{5}{c}{\textbf{ImageNet-1K}} 
& \multicolumn{1}{c}{} 
& \multicolumn{2}{c}{\textbf{Office-Home, CUB, DomainNet}} \\
\cmidrule(lr){3-7} \cmidrule(lr){9-10}
\textbf{Backbone}
& \textbf{Method}
& \textbf{FA} 
& \textbf{RA} 
& \textbf{TFA} 
& \textbf{TRA} 
& \textbf{Avg. Gap} 
& \textbf{AGL} $\uparrow$
& \textbf{Avg. $k$-NN Gap} 
& \textbf{Avg. CKA}
& \textbf{AGR} $\uparrow$
& \textbf{H-LR} $\uparrow$
\\
\midrule
\multirow{8}{*}{\textbf{ConvNeXt-T}} 
& \textbf{Retrained}
& 0.0 (\textcolor{blue}{0.0}) 
& 92.7 (\textcolor{blue}{0.0}) 
& 0.0 (\textcolor{blue}{0.0}) 
& 81.0 (\textcolor{blue}{0.0}) 
& \textcolor{blue}{0.0} 
& 1.00
& \textcolor{blue}{0.0} 
& \textcolor{blue}{100.0}
& 1.00
& 1.00
\\
& \textbf{FT}
& 79.4 (\textcolor{blue}{79.4}) 
& \textbf{92.5 (\textcolor{blue}{0.2})} 
& 65.7 (\textcolor{blue}{65.7}) 
& \textbf{80.7 (\textcolor{blue}{0.3})} 
& \textcolor{blue}{36.4} 
& 0.07
& \textcolor{blue}{4.2} 
& \textbf{\textcolor{blue}{86.0}}
& \textbf{0.88}
& 0.13
\\
& \textbf{GA}
& 0.6 (\textcolor{blue}{0.6}) 
& 0.8 (\textcolor{blue}{91.9}) 
& 0.6 (\textcolor{blue}{0.6}) 
& 0.8 (\textcolor{blue}{71.2}) 
& \textcolor{blue}{41.1} 
& 0.02
& \textcolor{blue}{31.1} 
& \textcolor{blue}{34.7} 
& 0.19
& 0.03
\\
& \textbf{RL}
& 0.5 (\textcolor{blue}{0.5}) 
& 0.7 (\textcolor{blue}{92.0}) 
& 0.4 (\textcolor{blue}{0.4}) 
& 0.6 (\textcolor{blue}{80.4}) 
& \textcolor{blue}{43.3} 
& 0.02
& \textcolor{blue}{37.7} 
& \textcolor{blue}{28.8}
& 0.12
& 0.03
\\
& \textbf{PL}
& 1.5 (\textcolor{blue}{1.5}) 
& 89.9 (\textcolor{blue}{2.8}) 
& 1.0 (\textcolor{blue}{1.0}) 
& 79.8 (\textcolor{blue}{0.9}) 
& \textbf{\textcolor{blue}{1.6}} 
& \textbf{0.94}
& \textcolor{blue}{3.4} 
& \textcolor{blue}{83.9}
& 0.85
& \textbf{0.89}
\\
& \textbf{SalUn}
& 0.5 (\textcolor{blue}{0.5}) 
& 0.7 (\textcolor{blue}{92.0}) 
& 0.5 (\textcolor{blue}{0.5}) 
& 0.6 (\textcolor{blue}{80.4}) 
& \textcolor{blue}{43.4} 
& 0.02
& \textcolor{blue}{38.1} 
& \textcolor{blue}{30.1}
& 0.12
& 0.03
\\
& \textbf{CU}
& 2.8 (\textcolor{blue}{2.8}) 
& 83.4 (\textcolor{blue}{6.6}) 
& 3.0 (\textcolor{blue}{3.0}) 
& 78.0 (\textcolor{blue}{3.0}) 
& \textcolor{blue}{3.9} 
& 0.83
& \textbf{\textcolor{blue}{3.0}} 
& \textcolor{blue}{81.2}
& 0.81
& 0.82
\\
& \textbf{SCRUB}
& \textbf{0.0 (\textcolor{blue}{0.0})} 
& 0.1 (\textcolor{blue}{92.6}) 
& \textbf{0.0 (\textcolor{blue}{0.0})} 
& 0.1 (\textcolor{blue}{80.9}) 
& \textcolor{blue}{43.4} 
& 0.01
& \textcolor{blue}{47.3} 
& \textcolor{blue}{17.4}
& 0.05
& 0.02
\\
\midrule
\multirow{8}{*}{\textbf{Swin-T}} 
& \textbf{Retrained} 
& 0.0 (\textcolor{blue}{0.0}) 
& 90.0 (\textcolor{blue}{0.0}) 
& 0.0 (\textcolor{blue}{0.0}) 
& 81.2 (\textcolor{blue}{0.0}) 
& \textcolor{blue}{0.0} 
& 1.00
& \textcolor{blue}{0.0} 
& \textcolor{blue}{100.0}
& 1.00
& 1.00
\\
& \textbf{FT}
& 63.4 (\textcolor{blue}{63.4}) 
& \textbf{88.6 (\textcolor{blue}{1.4})} 
& 55.8 (\textcolor{blue}{55.8}) 
& \textbf{81.3 (\textcolor{blue}{0.1})} 
& \textcolor{blue}{30.2} 
& 0.16
& \textbf{\textcolor{blue}{2.6}} 
& \textbf{\textcolor{blue}{89.6}}
& \textbf{0.89}
& 0.27
\\
& \textbf{GA}
& \textbf{0.0 (\textcolor{blue}{0.0})} 
& 0.1 (\textcolor{blue}{89.9}) 
& \textbf{0.0 (\textcolor{blue}{0.0})} 
& 0.0 (\textcolor{blue}{81.2}) 
& \textcolor{blue}{42.8} 
& 0.02
& \textcolor{blue}{32.1} 
& \textcolor{blue}{35.4}
& 0.20
& 0.03
\\
& \textbf{RL}
& 6.4 (\textcolor{blue}{6.4}) 
& 8.0 (\textcolor{blue}{80.6}) 
& 6.3 (\textcolor{blue}{6.3}) 
& 7.4 (\textcolor{blue}{73.8}) 
& \textcolor{blue}{41.8} 
& 0.05
& \textcolor{blue}{24.3} 
& \textcolor{blue}{54.5}
& 0.38
& 0.09
\\
& \textbf{PL}
& 4.3 (\textcolor{blue}{4.3}) 
& 86.2 (\textcolor{blue}{3.8}) 
& 3.1 (\textcolor{blue}{3.1}) 
& 80.2 (\textcolor{blue}{1.0}) 
& \textbf{\textcolor{blue}{3.1}} 
& \textbf{0.88}
& \textcolor{blue}{4.1} 
& \textcolor{blue}{88.4}
& 0.86
& \textbf{0.87}
\\
& \textbf{SalUn}
& 6.4 (\textcolor{blue}{6.4}) 
& 8.0 (\textcolor{blue}{80.6}) 
& 6.5 (\textcolor{blue}{6.5}) 
& 7.7 (\textcolor{blue}{73.5}) 
& \textcolor{blue}{41.8} 
& 0.05
& \textcolor{blue}{24.1} 
& \textcolor{blue}{54.5}
& 0.38
& 0.09
\\
& \textbf{CU}
& 2.1 (\textcolor{blue}{2.1}) 
& 80.0 (\textcolor{blue}{10.0}) 
& 2.2 (\textcolor{blue}{2.2}) 
& 77.0 (\textcolor{blue}{4.3}) 
& \textcolor{blue}{4.7} 
& 0.82
& \textcolor{blue}{2.9} 
& \textcolor{blue}{84.3}
& 0.84
& 0.83
\\
& \textbf{SCRUB}
& 8.2 (\textcolor{blue}{8.2}) 
& 28.0 (\textcolor{blue}{52.0}) 
& 8.8 (\textcolor{blue}{8.8}) 
& 31.8 (\textcolor{blue}{49.2}) 
& \textcolor{blue}{29.6} 
& 0.20
& \textcolor{blue}{14.8} 
& \textcolor{blue}{56.6}
& 0.49
& 0.29
\\
\bottomrule
\end{tabular}
}
\caption{Unlearning evaluation with 100 Random Class-wise Forgetting on ConvNeXt and Swin Transformers. The \textbf{bold numbers} represent optimal values.}
\label{tab:backbone_architecture}
\end{table}

\noindent\textbf{Standard deviations.} \ \
The standard deviations are in Table~\ref{tab:std_table} to quantify the variability and uncertainty of the measured results.

\begin{table}[ht]
\centering
\resizebox{\textwidth}{!}{
\begin{tabular}{lcccccccccc}
\toprule
\multicolumn{1}{c}{}
& \multicolumn{4}{c}{\textbf{ImageNet-1K}}
& \multicolumn{1}{c}{\textbf{Office-Home}}
& \multicolumn{1}{c}{\textbf{CUB}}
& \multicolumn{1}{c}{\textbf{DomainNet}}
& \multicolumn{1}{c}{\textbf{Office-Home}}
& \multicolumn{1}{c}{\textbf{CUB}}
& \multicolumn{1}{c}{\textbf{DomainNet}}
\\
\cmidrule(lr){2-5}
\cmidrule(lr){6-8}
\cmidrule(lr){9-11}
\textbf{Method}
& \textbf{FA}
& \textbf{RA}
& \textbf{TFA}
& \textbf{TRA}
& \multicolumn{3}{c}{$k$-NN}
& \multicolumn{3}{c}{CKA(\thetau, \thetar)}
\\
\midrule
\textbf{Original} 
& 79.2$\scriptstyle\pm0.10$
& 80.1$\scriptstyle\pm0.12$
& 76.1$\scriptstyle\pm0.08$
& 76.5$\scriptstyle\pm0.11$
& 80.3$\scriptstyle\pm0.15$
& 43.4$\scriptstyle\pm0.13$
& 84.0$\scriptstyle\pm0.14$
& 89.8$\scriptstyle\pm0.10$
& 82.1$\scriptstyle\pm0.09$
& 81.8$\scriptstyle\pm0.11$
\\
\textbf{Retrained}
& 0.0$\scriptstyle\pm0.20$
& 76.0$\scriptstyle\pm0.18$
& 0.0$\scriptstyle\pm0.22$
& 75.6$\scriptstyle\pm0.19$
& 77.3$\scriptstyle\pm0.25$
& 36.9$\scriptstyle\pm0.27$
& 82.0$\scriptstyle\pm0.24$
& 100.0$\scriptstyle\pm0.20$
& 100.0$\scriptstyle\pm0.23$
& 100.0$\scriptstyle\pm0.21$
\\
\textbf{FT} 
& 9.9$\scriptstyle\pm0.25$
& 79.5$\scriptstyle\pm0.30$
& 10.5$\scriptstyle\pm0.22$
& 76.0$\scriptstyle\pm0.28$
& 78.4$\scriptstyle\pm0.40$
& 39.0$\scriptstyle\pm0.35$
& 81.6$\scriptstyle\pm0.45$
& 87.6$\scriptstyle\pm0.30$
& 79.0$\scriptstyle\pm0.28$
& 79.9$\scriptstyle\pm0.32$
\\
\midrule
\textbf{GA} 
& 1.5$\scriptstyle\pm1.20$
& 11.4$\scriptstyle\pm1.50$
& 1.5$\scriptstyle\pm1.10$
& 11.3$\scriptstyle\pm1.40$
& 29.7$\scriptstyle\pm2.00$
& 10.4$\scriptstyle\pm1.80$
& 42.2$\scriptstyle\pm2.20$
& 8.3$\scriptstyle\pm1.10$
& 9.8$\scriptstyle\pm1.30$
& 10.2$\scriptstyle\pm1.25$
\\
\textbf{RL} 
& 6.3$\scriptstyle\pm1.80$
& 14.3$\scriptstyle\pm2.20$
& 5.5$\scriptstyle\pm1.60$
& 13.0$\scriptstyle\pm2.00$
& 44.0$\scriptstyle\pm2.50$
& 8.0$\scriptstyle\pm1.70$
& 35.2$\scriptstyle\pm2.10$
& 6.3$\scriptstyle\pm1.20$
& 5.8$\scriptstyle\pm1.15$
& 4.7$\scriptstyle\pm1.05$
\\
\textbf{PL}
& 1.0$\scriptstyle\pm0.30$
& 79.5$\scriptstyle\pm0.35$
& 1.0$\scriptstyle\pm0.28$
& 76.9$\scriptstyle\pm0.33$
& 80.3$\scriptstyle\pm0.45$
& 43.3$\scriptstyle\pm0.42$
& 84.0$\scriptstyle\pm0.40$
& 91.6$\scriptstyle\pm0.27$
& 84.7$\scriptstyle\pm0.25$
& 84.5$\scriptstyle\pm0.29$
\\
\textbf{SalUn} 
& 10.4$\scriptstyle\pm1.00$
& 19.7$\scriptstyle\pm1.30$
& 9.3$\scriptstyle\pm0.90$
& 19.1$\scriptstyle\pm1.20$
& 38.5$\scriptstyle\pm1.50$
& 8.0$\scriptstyle\pm1.10$
& 42.5$\scriptstyle\pm1.60$
& 9.7$\scriptstyle\pm1.40$
& 8.0$\scriptstyle\pm1.25$
& 8.5$\scriptstyle\pm1.35$
\\
\textbf{DUCK} 
& 0.9$\scriptstyle\pm0.45$
& 74.6$\scriptstyle\pm0.50$
& 0.9$\scriptstyle\pm0.42$
& 74.5$\scriptstyle\pm0.48$
& 79.8$\scriptstyle\pm0.55$
& 39.0$\scriptstyle\pm0.48$
& 82.5$\scriptstyle\pm0.52$
& 90.7$\scriptstyle\pm0.40$
& 83.2$\scriptstyle\pm0.38$
& 84.9$\scriptstyle\pm0.45$
\\
\textbf{CU} 
& 2.1$\scriptstyle\pm0.60$
& 73.9$\scriptstyle\pm0.65$
& 2.2$\scriptstyle\pm0.58$
& 73.3$\scriptstyle\pm0.62$
& 75.8$\scriptstyle\pm0.70$
& 33.9$\scriptstyle\pm0.55$
& 80.1$\scriptstyle\pm0.75$
& 85.6$\scriptstyle\pm0.50$
& 75.9$\scriptstyle\pm0.45$
& 76.1$\scriptstyle\pm0.48$
\\
\textbf{SCAR} 
& 4.2$\scriptstyle\pm1.10$
& 80.5$\scriptstyle\pm1.45$
& 3.9$\scriptstyle\pm1.05$
& 77.4$\scriptstyle\pm1.35$
& 78.0$\scriptstyle\pm1.60$
& 42.8$\scriptstyle\pm1.30$
& 83.1$\scriptstyle\pm1.70$
& 74.2$\scriptstyle\pm1.50$
& 65.4$\scriptstyle\pm1.45$
& 58.8$\scriptstyle\pm1.55$
\\
\textbf{SCRUB}
& 1.1$\scriptstyle\pm0.85$
& 67.3$\scriptstyle\pm1.15$
& 1.1$\scriptstyle\pm0.80$
& 65.7$\scriptstyle\pm1.10$
& 74.7$\scriptstyle\pm1.40$
& 42.2$\scriptstyle\pm0.95$
& 80.9$\scriptstyle\pm1.45$
& 68.9$\scriptstyle\pm1.30$
& 63.8$\scriptstyle\pm1.25$
& 52.8$\scriptstyle\pm1.35$
\\
\bottomrule
\end{tabular}
}
\caption{The standard deviation of each value in Tables~\ref{tab:performance_comparison_part1}, \ref{tab:performance_comparison_part2} in the main paper. The standard deviations correspond to 95\% confidence intervals, with each algorithm run 5 times.}
\label{tab:std_table}
\end{table}

\begin{figure*}[t]
    \vspace{-2em}
    \hfill
    \begin{subfigure}[b]{0.24\linewidth}
        \centering
        \includegraphics[width=\linewidth]{Fig3a_tsne_original.pdf}
        \caption{Original (\thetao)}
    \end{subfigure}
    \hspace{2em}
    \begin{subfigure}[b]{0.23\linewidth}
        \centering
        \includegraphics[width=\linewidth]{Fig3b_tsne_retrained.pdf}
        \caption{Retrained (\thetar)}
    \end{subfigure}
    \hspace{2em}
    \begin{subfigure}[b]{0.23\linewidth}
        \centering
        \includegraphics[width=\linewidth]{Fig3c_tsne_FT.pdf}
        \caption{FT}
    \end{subfigure}
    \hfill
    \vspace{1em}

    \begin{tabular}{cccc}
        \begin{subfigure}[b]{0.23\linewidth}
            \centering
            \includegraphics[width=\linewidth]{Fig3d_tsne_GA.pdf}
            \caption{GA}
        \end{subfigure} &
        \begin{subfigure}[b]{0.23\linewidth}
            \centering
            \includegraphics[width=\linewidth]{Fig3e_tsne_RL.pdf}
            \caption{RL}
        \end{subfigure} &
        \begin{subfigure}[b]{0.23\linewidth}
            \centering
            \includegraphics[width=\linewidth]{Fig3f_tsne_SalUn.pdf}
            \caption{SalUn}
        \end{subfigure} &
        \begin{subfigure}[b]{0.23\linewidth}
            \centering
            \includegraphics[width=\linewidth]{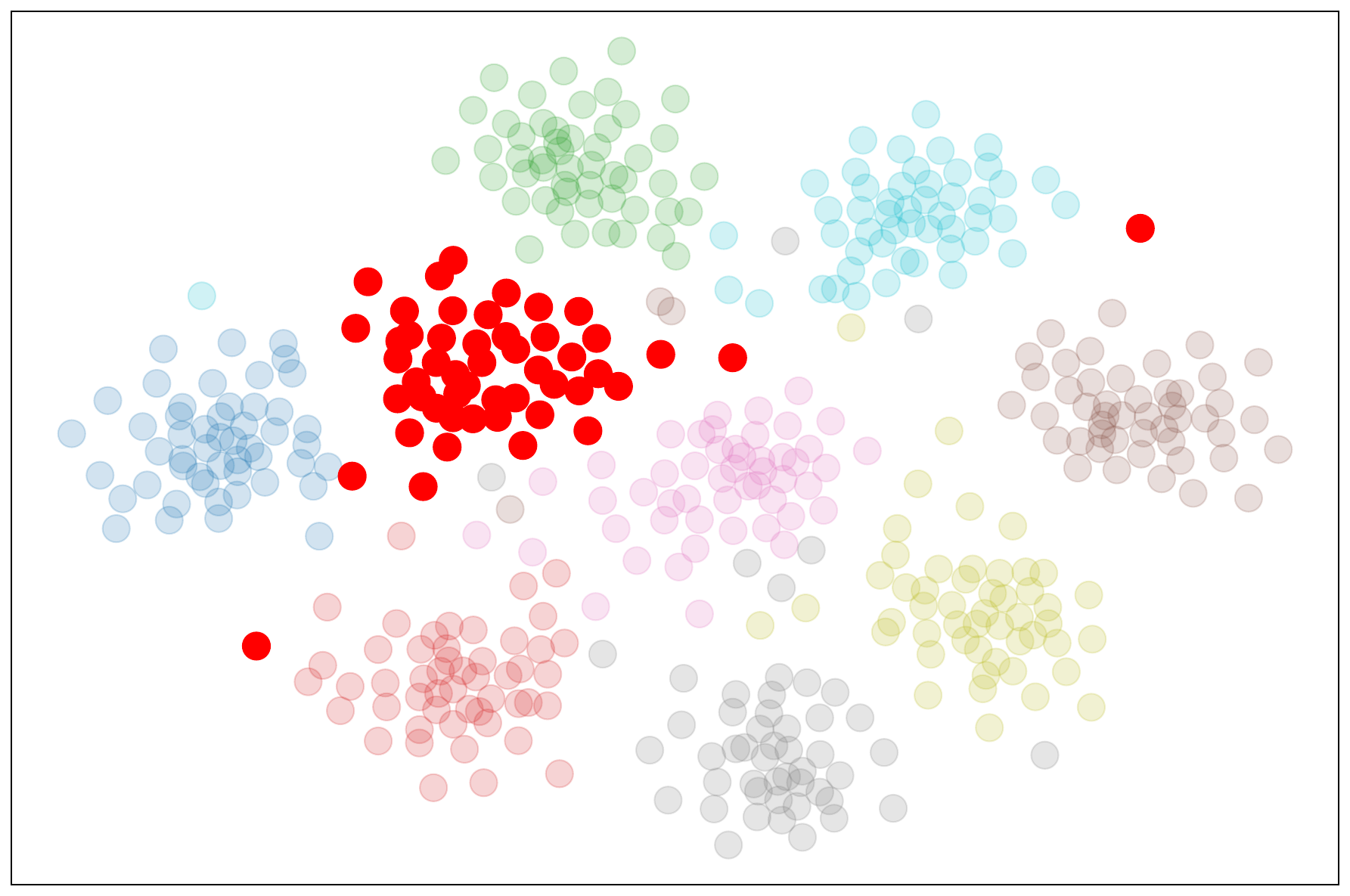}
            \caption{PL}
        \end{subfigure}
        \\
    \end{tabular}

    \begin{tabular}{cccc}
        \begin{subfigure}[b]{0.23\linewidth}
            \centering
            \includegraphics[width=\linewidth]{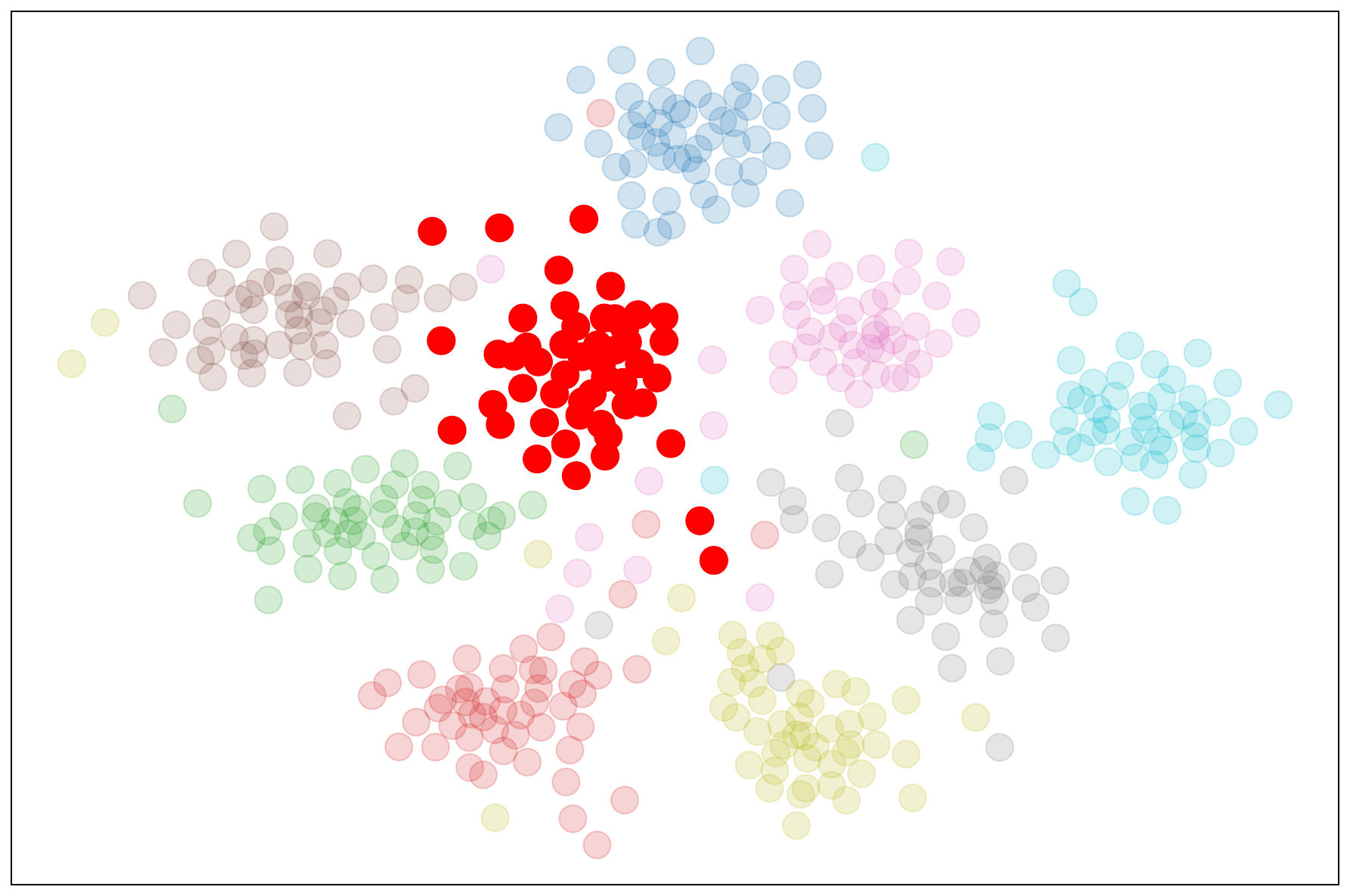}
            \caption{DUCK}
        \end{subfigure} &
        \begin{subfigure}[b]{0.23\linewidth}
            \centering
            \includegraphics[width=\linewidth]{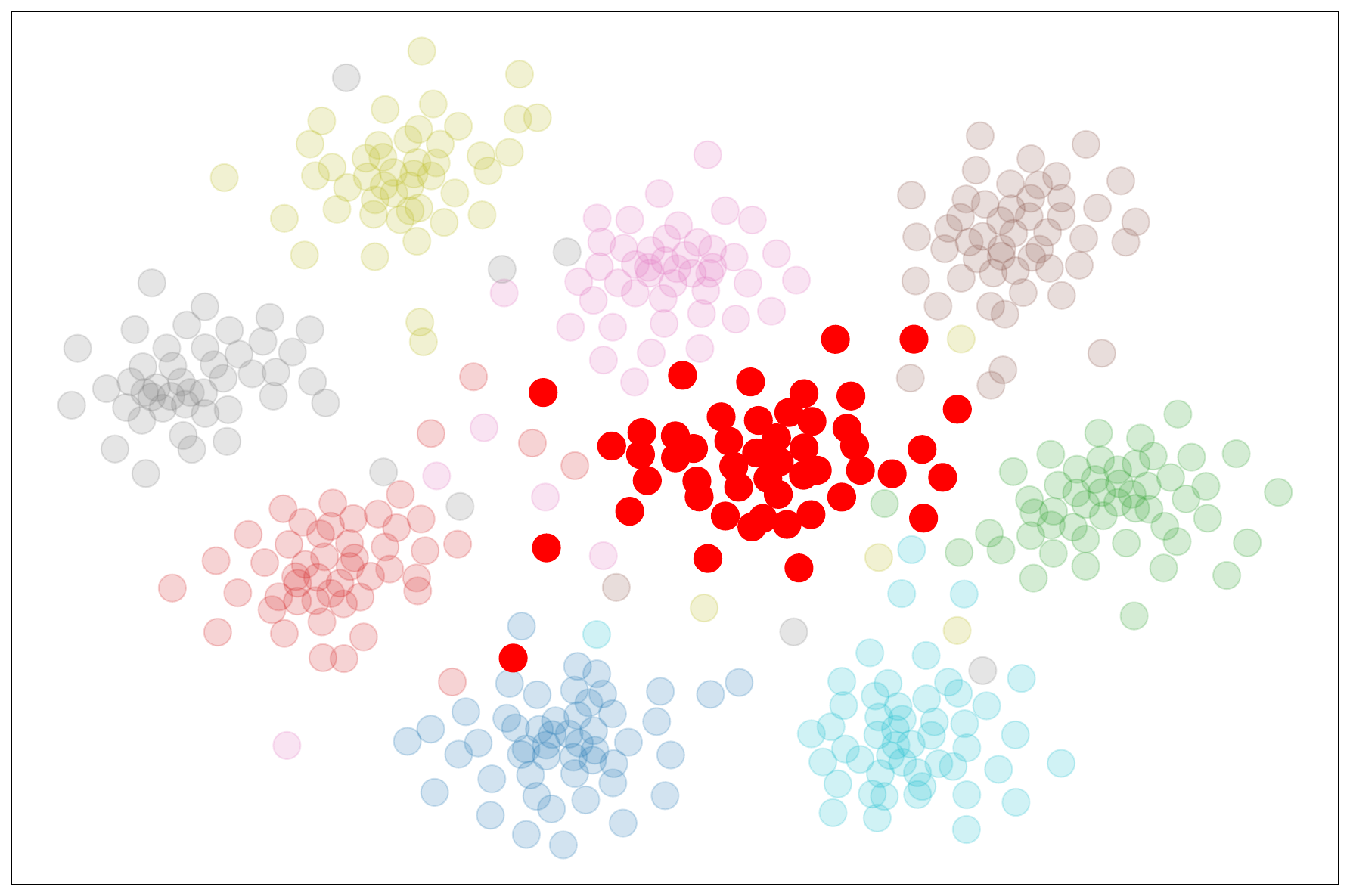}
            \caption{CU}
        \end{subfigure} &
        \begin{subfigure}[b]{0.23\linewidth}
            \centering
            \includegraphics[width=\linewidth]{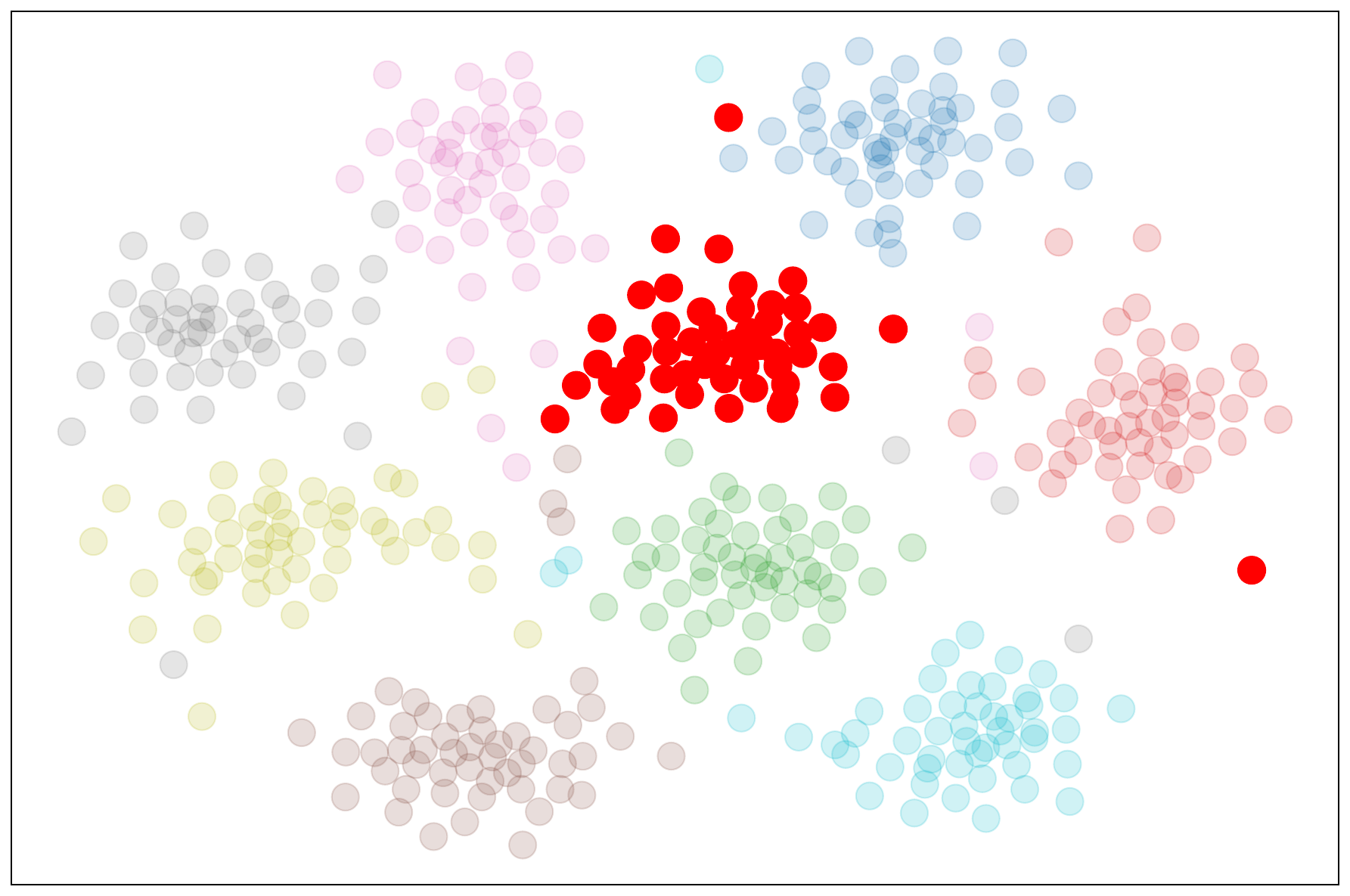}
            \caption{SCAR}
        \end{subfigure} &
        \begin{subfigure}[b]{0.23\linewidth}
            \centering
            \includegraphics[width=\linewidth]{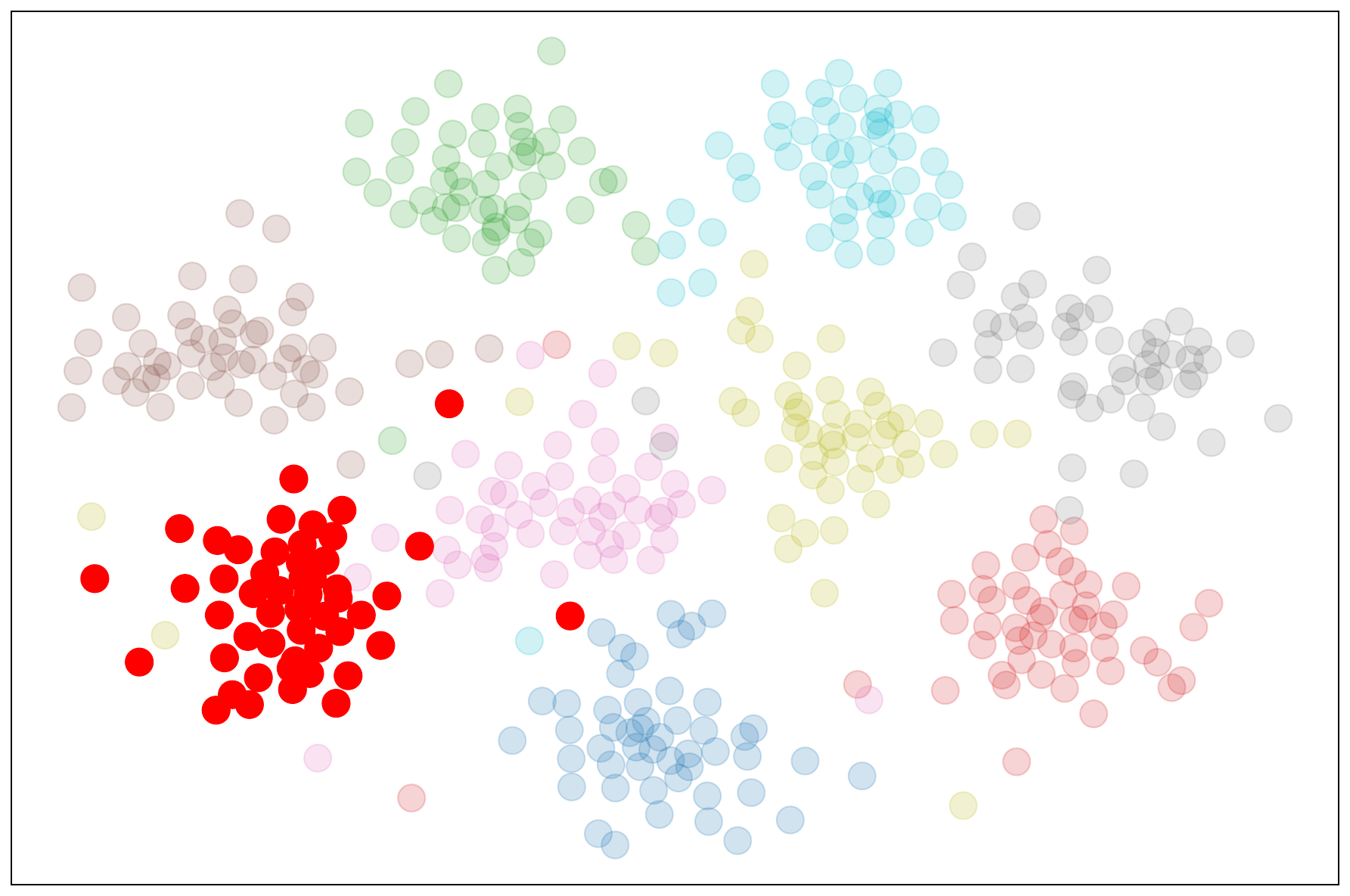}
            \caption{SCRUB}
        \end{subfigure}
        \\
    \end{tabular}
    \caption{Visualizations of the feature representations from the original (\thetao), retrained (\thetar), and unlearned models (\thetau) on a subset of ImageNet-1K, using ResNet-50.}
    \label{fig:full_visualization}
\end{figure*}

\noindent\textbf{t-SNE visualizations.} \ \
Fig.~\ref{fig:full_visualization} provides a comprehensive t-SNE visualizations for all baseline methods, expanding upon the subset shown in Figure \ref{fig:visualization} in the main text. These plots visualize the feature embeddings from the encoder for a consistent subset of retain and forget classes (red).

\noindent\textbf{Impact of scaling the number of forgetting classes.} \ \
The full results in Tables~\ref{tab:random}-~\ref{tab:all_h-lr} allow for a detailed analysis of algorithm scalability as the number of forgetting classes (N) increases from 100 to 300. This analysis reveals a critical divergence in stability. As detailed by the H-LR scores in Table~\ref{tab:all_h-lr}, methods like PL, DUCK, CU demonstrate remarkable robustness. For instance, in the Random scenario, PL's H-LR score only drops slightly from 0.89 (N=100) to 0.86 (N=300). In stark contrast, several methods that rely on more complex optimization objectives suffer a performance collapse. \textbf{SCRUB}'s H-LR score plummets from 0.69 (N=100) to a mere \textbf{0.33} (N=300). This scalability failure reinforces our hypothesis from Section 4.4 about \qq{unbounded loss functions} and optimization instability. It provides empirical evidence that these methods are not practically scalable for large-scale unlearning tasks, despite their promising results in small-scale or small-N settings.

\begin{figure}[t]
    \centering
    \includegraphics[width=0.5\linewidth]{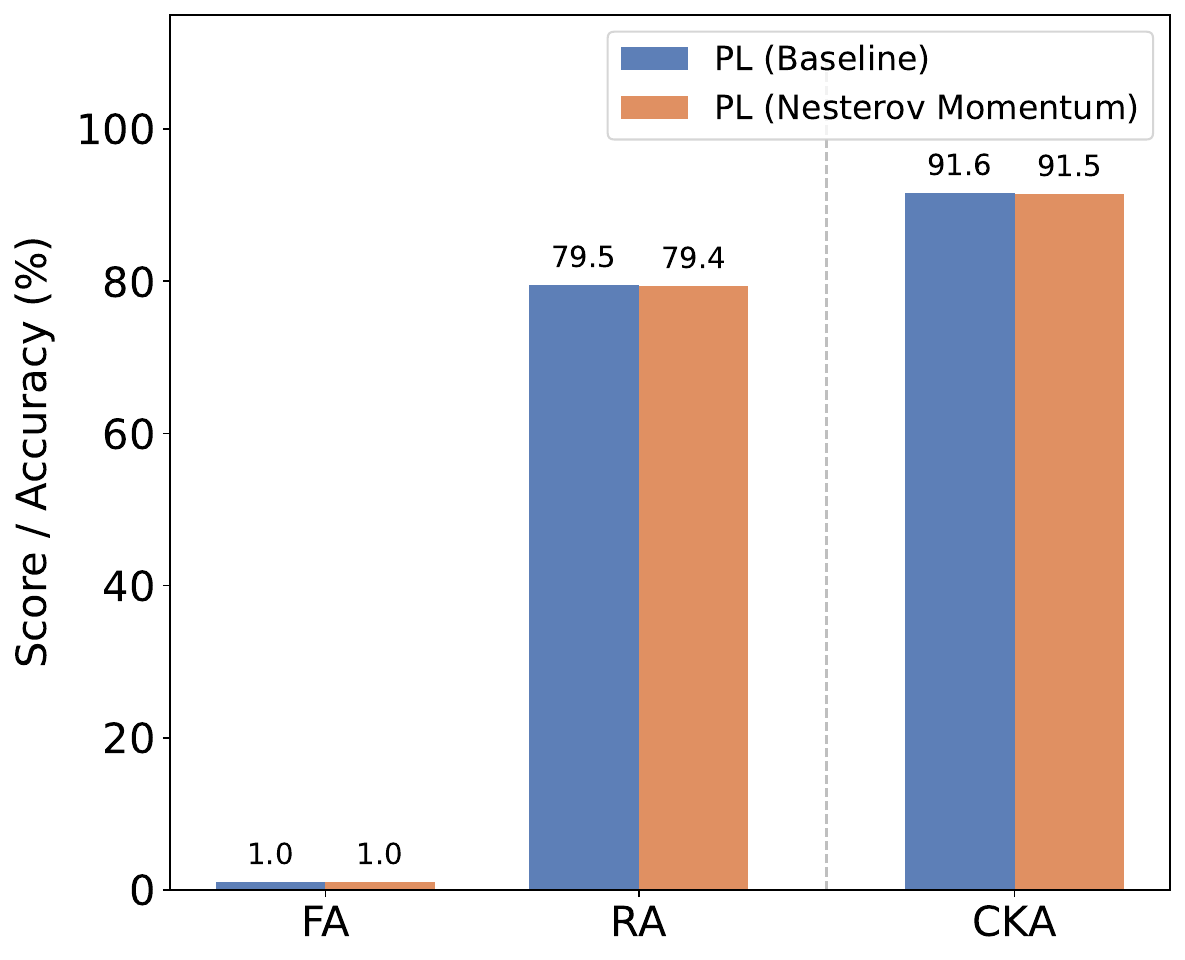}
    \caption{Comparison of PL baseline and PL with Nesterov Momentum.}
    \label{fig:momentum_effect}
\end{figure}

\noindent\textbf{Impact of optimization strategies on shortcut learning.} \ \
In Section 3.7 (\qq{Why does the unlearned model maintain high similarity to the original model?}), we discussed the phenomenon of \qq{shortcut learning}, where the unlearning process merely modifies the classifier. Addressing the hypothesis that optimization dynamics might be the cause, we investigated whether acceleration methods, specifically Nesterov momentum, could enable the model to bypass these sub-optimal solutions~\citep{jiang2024feduhb}. We compared the Pseudo Labeling (PL) baseline against a version optimized with Nesterov momentum under the Random-100 Class-wise Forgetting scenario. As illustrated in Fig.~\ref{fig:momentum_effect}, the inclusion of momentum resulted in negligible differences across both logit-based (FA, RA) and representation-based ($CKA(\theta_u, \theta_r)$) metrics. These results indicate that conventional momentum strategies alone cannot mitigate the shortcut learning embedded in current unlearning objectives. This suggests that addressing shortcut learning may require broader changes—such as revisiting the objective formulation or modifying architectural components. We leave the exploration of more advanced strategies to future work.

\begin{table}[th]
\centering
\resizebox{\columnwidth}{!}{
\begin{tabular}{llccccc}
\toprule
\multicolumn{7}{c}{\textbf{Random}} \\
\cmidrule(lr){3-7}
\textbf{Classes} & \textbf{Method} & \textbf{FA} & \textbf{RA} & \textbf{TFA} & \textbf{TRA} & \textbf{Avg. Gap} \\
\midrule
\multirow{10}{*}{\textbf{100}}
& \textbf{Retrained} & 0.0 (\textcolor{blue}{0.0}) & 76.0 (\textcolor{blue}{0.0}) & 0.0 (\textcolor{blue}{0.0}) & 75.6 (\textcolor{blue}{0.0}) & \textcolor{blue}{0.0} \\
& \textbf{FT} & 9.9 (\textcolor{blue}{9.9}) & 79.5 (\textcolor{blue}{3.5}) & 10.5 (\textcolor{blue}{10.5}) & 76.0 (\textcolor{blue}{0.4}) & \textcolor{blue}{6.1} \\
& \textbf{GA} & 1.5 (\textcolor{blue}{1.5}) & 11.4 (\textcolor{blue}{64.6}) & 1.5 (\textcolor{blue}{1.5}) & 11.3 (\textcolor{blue}{64.3}) & \textcolor{blue}{33.0} \\
& \textbf{RL} & 6.3 (\textcolor{blue}{6.3}) & 14.3 (\textcolor{blue}{61.7}) & 5.5 (\textcolor{blue}{5.5}) & 13.0 (\textcolor{blue}{62.6}) & \textcolor{blue}{34.3} \\
& \textbf{PL} & 1.0 (\textcolor{blue}{1.0}) & 79.5 (\textcolor{blue}{3.5}) & 1.0 (\textcolor{blue}{1.0}) & 76.9 (\textcolor{blue}{0.9}) & \textcolor{blue}{1.6} \\
& \textbf{SalUn} & 10.4 (\textcolor{blue}{10.4}) & 22.1 (\textcolor{blue}{53.9}) & 9.3 (\textcolor{blue}{9.3}) & 19.1 (\textcolor{blue}{56.5}) & \textcolor{blue}{32.5} \\
& \textbf{DUCK} & 0.9 (\textcolor{blue}{0.9}) & 74.6 (\textcolor{blue}{1.4}) & 0.9 (\textcolor{blue}{0.9}) & 74.5 (\textcolor{blue}{1.1}) & \textbf{\textcolor{blue}{1.1}} \\
& \textbf{CU} & 2.1 (\textcolor{blue}{2.1}) & 73.9 (\textcolor{blue}{2.1}) & 2.2 (\textcolor{blue}{2.2}) & 73.3 (\textcolor{blue}{2.3}) & \textcolor{blue}{2.2} \\
& \textbf{SCAR} & 4.2 (\textcolor{blue}{4.2}) & 80.5 (\textcolor{blue}{4.5}) & 3.9 (\textcolor{blue}{3.9}) & 77.4 (\textcolor{blue}{1.8}) & \textcolor{blue}{3.6} \\
& \textbf{SCRUB} & 1.1 (\textcolor{blue}{1.1}) & 67.3 (\textcolor{blue}{8.7}) & 1.1 (\textcolor{blue}{1.1}) & 65.7 (\textcolor{blue}{9.9}) & \textcolor{blue}{5.2} \\
\midrule
\multirow{10}{*}{\textbf{200}}
& \textbf{Retrained} & 0.0 (\textcolor{blue}{0.0}) & 77.7 (\textcolor{blue}{0.0}) & 0.0 (\textcolor{blue}{0.0}) & 77.0 (\textcolor{blue}{0.0}) & \textcolor{blue}{0.0} \\
& \textbf{FT} & 11.7 (\textcolor{blue}{11.7}) & 80.7 (\textcolor{blue}{3.0}) & 13.0 (\textcolor{blue}{13.0}) & 77.2 (\textcolor{blue}{0.2}) & \textcolor{blue}{7.0} \\
& \textbf{GA} & 0.1 (\textcolor{blue}{0.1}) & 0.3 (\textcolor{blue}{77.4}) & 0.1 (\textcolor{blue}{0.1}) & 0.2 (\textcolor{blue}{76.8}) & \textcolor{blue}{38.6} \\
& \textbf{RL} & 4.4 (\textcolor{blue}{4.4}) & 8.0 (\textcolor{blue}{69.7}) & 3.8 (\textcolor{blue}{3.8}) & 6.4 (\textcolor{blue}{70.6}) & \textcolor{blue}{37.1} \\
& \textbf{PL} & 1.2 (\textcolor{blue}{1.2}) & 80.2 (\textcolor{blue}{2.5}) & 1.0 (\textcolor{blue}{1.0}) & 78.1 (\textcolor{blue}{1.1}) & \textcolor{blue}{1.5} \\
& \textbf{SalUn} & 9.7 (\textcolor{blue}{9.7}) & 14.4 (\textcolor{blue}{63.3}) & 8.5 (\textcolor{blue}{8.5}) & 12.6 (\textcolor{blue}{64.4}) & \textcolor{blue}{36.5} \\
& \textbf{DUCK} & 0.8 (\textcolor{blue}{0.8}) & 78.1 (\textcolor{blue}{0.4}) & 0.8 (\textcolor{blue}{0.8}) & 77.3 (\textcolor{blue}{0.3}) & \textbf{\textcolor{blue}{0.6}} \\
& \textbf{CU} & 1.0 (\textcolor{blue}{1.0}) & 79.7 (\textcolor{blue}{2.0}) & 1.2 (\textcolor{blue}{1.2}) & 77.4 (\textcolor{blue}{0.4}) & \textcolor{blue}{1.2} \\
& \textbf{SCAR} & 15.0 (\textcolor{blue}{15.0}) & 80.8 (\textcolor{blue}{3.1}) & 15.6 (\textcolor{blue}{15.6}) & 78.7 (\textcolor{blue}{1.7}) & \textcolor{blue}{8.9} \\
& \textbf{SCRUB} & 18.3 (\textcolor{blue}{18.3}) & 72.5 (\textcolor{blue}{5.2}) & 16.9 (\textcolor{blue}{16.9}) & 69.7 (\textcolor{blue}{7.3}) & \textcolor{blue}{11.9} \\
\midrule
\multirow{10}{*}{\textbf{300}}
& \textbf{Retrained} & 0.0 (\textcolor{blue}{0.0}) & 78.9 (\textcolor{blue}{0.0}) & 0.0 (\textcolor{blue}{0.0}) & 77.6 (\textcolor{blue}{0.0}) & \textcolor{blue}{0.0} \\
& \textbf{FT} & 14.0 (\textcolor{blue}{14.0}) & 82.0 (\textcolor{blue}{3.1}) & 15.4 (\textcolor{blue}{15.4}) & 78.9 (\textcolor{blue}{1.3}) & \textcolor{blue}{8.5} \\
& \textbf{GA} & 0.3 (\textcolor{blue}{0.3}) & 0.3 (\textcolor{blue}{78.6}) & 0.2 (\textcolor{blue}{0.2}) & 0.3 (\textcolor{blue}{77.3}) & \textcolor{blue}{39.1} \\
& \textbf{RL} & 6.4 (\textcolor{blue}{6.4}) & 9.0 (\textcolor{blue}{69.9}) & 5.5 (\textcolor{blue}{5.5}) & 8.1 (\textcolor{blue}{69.5}) & \textcolor{blue}{37.8} \\
& \textbf{PL} & 1.0 (\textcolor{blue}{1.0}) & 81.1 (\textcolor{blue}{2.2}) & 0.9 (\textcolor{blue}{0.9}) & 79.3 (\textcolor{blue}{1.7}) & \textcolor{blue}{1.5} \\
& \textbf{SalUn} & 10.5 (\textcolor{blue}{10.5}) & 14.3 (\textcolor{blue}{64.6}) & 8.6 (\textcolor{blue}{8.6}) & 12.4 (\textcolor{blue}{65.2}) & \textcolor{blue}{37.2} \\
& \textbf{DUCK} & 0.9 (\textcolor{blue}{0.9}) & 79.5 (\textcolor{blue}{0.6}) & 1.0 (\textcolor{blue}{1.0}) & 75.6 (\textcolor{blue}{2.0}) & \textbf{\textcolor{blue}{1.1}} \\
& \textbf{CU} & 0.7 (\textcolor{blue}{0.7}) & 81.1 (\textcolor{blue}{2.2}) & 0.8 (\textcolor{blue}{0.8}) & 78.6 (\textcolor{blue}{1.0}) & \textcolor{blue}{1.2} \\
& \textbf{SCAR} & 13.7 (\textcolor{blue}{13.7}) & 81.7 (\textcolor{blue}{2.8}) & 14.6 (\textcolor{blue}{14.6}) & 80.0 (\textcolor{blue}{2.4}) & \textcolor{blue}{8.4} \\
& \textbf{SCRUB} & 55.6 (\textcolor{blue}{55.6}) & 80.9 (\textcolor{blue}{2.0}) & 51.7 (\textcolor{blue}{51.7}) & 77.5 (\textcolor{blue}{0.1}) & \textcolor{blue}{27.4} \\
\bottomrule
\end{tabular}
}
\caption{Random class-wise Forgetting results. The \textcolor{blue}{blue numbers} indicate the gap. The \textbf{bold numbers} represent optimal values.}
\label{tab:random}
\end{table}

\begin{table}[ht]
\centering
\resizebox{\columnwidth}{!}{
\begin{tabular}{llccccc}
\toprule
\multicolumn{7}{c}{\textbf{Office-Home Top}} \\
\cmidrule(lr){3-7}
\textbf{Classes} & \textbf{Method} & \textbf{FA} & \textbf{RA} & \textbf{TFA} & \textbf{TRA} & \textbf{Avg. Gap} \\
\midrule
\multirow{10}{*}{\textbf{100}}
& \textbf{Retrained} & 0.0 (\textcolor{blue}{0.0}) & 76.2 (\textcolor{blue}{0.0}) & 0.0 (\textcolor{blue}{0.0}) & 76.1 (\textcolor{blue}{0.0}) & \textcolor{blue}{0.0} \\
& \textbf{FT} & 7.1 (\textcolor{blue}{7.1}) & 80.1 (\textcolor{blue}{4.1}) & 6.6 (\textcolor{blue}{6.6}) & 77.2 (\textcolor{blue}{1.1}) & \textcolor{blue}{4.7} \\
& \textbf{GA} & 0.5 (\textcolor{blue}{0.5}) & 0.2 (\textcolor{blue}{76.0}) & 0.4 (\textcolor{blue}{0.4}) & 0.2 (\textcolor{blue}{75.9}) & \textcolor{blue}{38.2} \\
& \textbf{RL} & 0.3 (\textcolor{blue}{0.3}) & 4.5 (\textcolor{blue}{71.7}) & 0.1 (\textcolor{blue}{0.1}) & 3.6 (\textcolor{blue}{72.5}) & \textcolor{blue}{36.6} \\
& \textbf{PL} & 1.1 (\textcolor{blue}{1.1}) & 80.0 (\textcolor{blue}{3.8}) & 1.0 (\textcolor{blue}{1.0}) & 77.9 (\textcolor{blue}{1.8}) & \textcolor{blue}{2.0} \\
& \textbf{SalUn} & 1.6 (\textcolor{blue}{1.6}) & 6.4 (\textcolor{blue}{69.8}) & 1.4 (\textcolor{blue}{1.4}) & 4.9 (\textcolor{blue}{71.2}) & \textcolor{blue}{36.0} \\
& \textbf{DUCK} & 1.5 (\textcolor{blue}{1.5}) & 73.9 (\textcolor{blue}{2.3}) & 1.4 (\textcolor{blue}{1.4}) & 74.4 (\textcolor{blue}{1.7}) & \textbf{\textcolor{blue}{1.7}} \\
& \textbf{CU} & 1.3 (\textcolor{blue}{1.3}) & 59.8 (\textcolor{blue}{16.4}) & 1.5 (\textcolor{blue}{1.5}) & 61.1 (\textcolor{blue}{15.0}) & \textcolor{blue}{8.6} \\
& \textbf{SCAR} & 5.0 (\textcolor{blue}{5.0}) & 81.3 (\textcolor{blue}{5.1}) & 5.0 (\textcolor{blue}{5.0}) & 78.7 (\textcolor{blue}{2.6}) & \textcolor{blue}{4.4} \\
& \textbf{SCRUB} & 0.0 (\textcolor{blue}{0.0}) & 0.1 (\textcolor{blue}{76.1}) & 0.0 (\textcolor{blue}{0.0}) & 0.1 (\textcolor{blue}{76.0}) & \textcolor{blue}{38.0} \\
\midrule
\multirow{10}{*}{\textbf{200}}
& \textbf{Retrained} & 0.0 (\textcolor{blue}{0.0}) & 78.6 (\textcolor{blue}{0.0}) & 0.0 (\textcolor{blue}{0.0}) & 78.2 (\textcolor{blue}{0.0}) & \textcolor{blue}{0.0} \\
& \textbf{FT} & 21.1 (\textcolor{blue}{21.1}) & 84.4 (\textcolor{blue}{5.8}) & 20.1 (\textcolor{blue}{20.1}) & 67.1 (\textcolor{blue}{11.1}) & \textcolor{blue}{14.5} \\
& \textbf{GA} & 0.2 (\textcolor{blue}{0.2}) & 0.6 (\textcolor{blue}{78.0}) & 0.2 (\textcolor{blue}{0.2}) & 0.5 (\textcolor{blue}{77.7}) & \textcolor{blue}{39.0} \\
& \textbf{RL} & 15.7 (\textcolor{blue}{15.7}) & 33.9 (\textcolor{blue}{44.7}) & 14.6 (\textcolor{blue}{14.6}) & 31.9 (\textcolor{blue}{46.3}) & \textcolor{blue}{30.3} \\
& \textbf{PL} & 1.1 (\textcolor{blue}{1.1}) & 81.1 (\textcolor{blue}{2.5}) & 1.0 (\textcolor{blue}{1.0}) & 79.5 (\textcolor{blue}{1.3}) & \textcolor{blue}{1.5} \\
& \textbf{SalUn} & 7.9 (\textcolor{blue}{7.9}) & 18.3 (\textcolor{blue}{60.3}) & 7.3 (\textcolor{blue}{7.3}) & 16.8 (\textcolor{blue}{61.4}) & \textcolor{blue}{34.2} \\
& \textbf{DUCK} & 1.0 (\textcolor{blue}{1.0}) & 78.4 (\textcolor{blue}{0.2}) & 1.2 (\textcolor{blue}{1.2}) & 78.4 (\textcolor{blue}{0.2}) & \textbf{\textcolor{blue}{0.7}} \\
& \textbf{CU} & 0.8 (\textcolor{blue}{0.8}) & 79.5 (\textcolor{blue}{0.9}) & 1.1 (\textcolor{blue}{1.1}) & 78.0 (\textcolor{blue}{0.2}) & \textcolor{blue}{0.8} \\
& \textbf{SCAR} & 13.7 (\textcolor{blue}{13.7}) & 79.1 (\textcolor{blue}{0.5}) & 13.7 (\textcolor{blue}{13.7}) & 78.4 (\textcolor{blue}{0.2}) & \textcolor{blue}{7.0} \\
& \textbf{SCRUB} & 0.0 (\textcolor{blue}{0.0}) & 0.1 (\textcolor{blue}{78.5}) & 0.0 (\textcolor{blue}{0.0}) & 0.1 (\textcolor{blue}{78.1}) & \textcolor{blue}{39.2} \\
\midrule
\multirow{10}{*}{\textbf{300}}
& \textbf{Retrained} & 0.0 (\textcolor{blue}{0.0}) & 80.7 (\textcolor{blue}{0.0}) & 0.0 (\textcolor{blue}{0.0}) & 79.9 (\textcolor{blue}{0.0}) & \textcolor{blue}{0.0} \\
& \textbf{FT} & 28.4 (\textcolor{blue}{28.4}) & 85.3 (\textcolor{blue}{4.6}) & 28.6 (\textcolor{blue}{28.6}) & 82.3 (\textcolor{blue}{2.4}) & \textcolor{blue}{16.0} \\
& \textbf{GA} & 0.2 (\textcolor{blue}{0.2}) & 0.2 (\textcolor{blue}{80.5}) & 0.2 (\textcolor{blue}{0.2}) & 0.3 (\textcolor{blue}{79.6}) & \textcolor{blue}{40.1} \\
& \textbf{RL} & 11.5 (\textcolor{blue}{11.5}) & 22.9 (\textcolor{blue}{57.8}) & 10.2 (\textcolor{blue}{10.2}) & 21.2 (\textcolor{blue}{58.7}) & \textcolor{blue}{34.6} \\
& \textbf{PL} & 1.0 (\textcolor{blue}{1.0}) & 81.9 (\textcolor{blue}{1.2}) & 1.0 (\textcolor{blue}{1.0}) & 81.0 (\textcolor{blue}{1.0}) & \textbf{\textcolor{blue}{1.1}} \\
& \textbf{SalUn} & 8.3 (\textcolor{blue}{8.3}) & 14.7 (\textcolor{blue}{66.0}) & 8.2 (\textcolor{blue}{8.2}) & 13.4 (\textcolor{blue}{66.5}) & \textcolor{blue}{37.3} \\
& \textbf{DUCK} & 2.4 (\textcolor{blue}{2.4}) & 80.2 (\textcolor{blue}{0.5}) & 2.7 (\textcolor{blue}{2.7}) & 80.0 (\textcolor{blue}{0.1}) & \textcolor{blue}{1.4} \\
& \textbf{CU} & 1.6 (\textcolor{blue}{1.6}) & 81.5 (\textcolor{blue}{0.8}) & 2.0 (\textcolor{blue}{2.0}) & 79.9 (\textcolor{blue}{0.0}) & \textbf{\textcolor{blue}{1.1}} \\
& \textbf{SCAR} & 20.8 (\textcolor{blue}{20.8}) & 82.7 (\textcolor{blue}{2.0}) & 21.0 (\textcolor{blue}{21.0}) & 81.7 (\textcolor{blue}{1.8}) & \textcolor{blue}{11.4} \\
& \textbf{SCRUB} & 0.1 (\textcolor{blue}{0.1}) & 0.1 (\textcolor{blue}{80.6}) & 0.1 (\textcolor{blue}{0.1}) & 0.1 (\textcolor{blue}{79.8}) & \textcolor{blue}{40.2} \\
\bottomrule
\end{tabular}
}
\caption{Office-Home Top Class-wise Forgetting results. The \textcolor{blue}{blue numbers} indicate the gap. The \textbf{bold numbers} represent optimal values.}
\label{tab:officehome}
\end{table}

\begin{table}[ht]
\centering
\resizebox{\columnwidth}{!}{
\begin{tabular}{llccccc}
\toprule
\multicolumn{7}{c}{\textbf{CUB Top}} \\
\cmidrule(lr){3-7}
\textbf{Classes} & \textbf{Method} & \textbf{FA} & \textbf{RA} & \textbf{TFA} & \textbf{TRA} & \textbf{Avg. Gap} \\
\midrule
\multirow{10}{*}{\textbf{100}}
& \textbf{Retrained} & 0.0 (\textcolor{blue}{0.0}) & 74.9 (\textcolor{blue}{0.0}) & 0.0 (\textcolor{blue}{0.0}) & 74.3 (\textcolor{blue}{0.0}) & \textcolor{blue}{0.0} \\
& \textbf{FT} & 24.4 (\textcolor{blue}{24.4}) & 78.5 (\textcolor{blue}{3.6}) & 28.4 (\textcolor{blue}{28.4}) & 74.9 (\textcolor{blue}{0.6}) & \textcolor{blue}{14.3} \\
& \textbf{GA} & 0.1 (\textcolor{blue}{0.1}) & 3.6 (\textcolor{blue}{71.3}) & 0.0 (\textcolor{blue}{0.0}) & 3.8 (\textcolor{blue}{70.5}) & \textcolor{blue}{35.5} \\
& \textbf{RL} & 9.6 (\textcolor{blue}{9.6}) & 40.2 (\textcolor{blue}{34.7}) & 6.4 (\textcolor{blue}{6.4}) & 34.7 (\textcolor{blue}{39.6}) & \textcolor{blue}{22.4} \\
& \textbf{PL} & 0.7 (\textcolor{blue}{0.7}) & 78.6 (\textcolor{blue}{3.7}) & 0.6 (\textcolor{blue}{0.6}) & 75.5 (\textcolor{blue}{1.2}) & \textbf{\textcolor{blue}{1.6}} \\
& \textbf{SalUn} & 5.2 (\textcolor{blue}{5.2}) & 30.3 (\textcolor{blue}{44.6}) & 3.2 (\textcolor{blue}{3.2}) & 25.8 (\textcolor{blue}{48.5}) & \textcolor{blue}{25.4} \\
& \textbf{DUCK} & 6.8 (\textcolor{blue}{6.8}) & 74.7 (\textcolor{blue}{0.2}) & 8.5 (\textcolor{blue}{8.5}) & 73.6 (\textcolor{blue}{0.7}) & \textcolor{blue}{4.1} \\
& \textbf{CU} & 0.5 (\textcolor{blue}{0.5}) & 43.4 (\textcolor{blue}{31.5}) & 0.4 (\textcolor{blue}{0.4}) & 46.2 (\textcolor{blue}{28.1}) & \textcolor{blue}{15.1} \\
& \textbf{SCAR} & 29.0 (\textcolor{blue}{29.0}) & 76.1 (\textcolor{blue}{1.2}) & 33.4 (\textcolor{blue}{33.4}) & 74.6 (\textcolor{blue}{0.3}) & \textcolor{blue}{16.0} \\
& \textbf{SCRUB} & 0.0 (\textcolor{blue}{0.0}) & 3.5 (\textcolor{blue}{71.4}) & 0.0 (\textcolor{blue}{0.0}) & 4.2 (\textcolor{blue}{70.1}) & \textcolor{blue}{35.4} \\
\midrule
\multirow{10}{*}{\textbf{200}}
& \textbf{Retrained} & 0.0 (\textcolor{blue}{0.0}) & 75.4 (\textcolor{blue}{0.0}) & 0.0 (\textcolor{blue}{0.0}) & 74.2 (\textcolor{blue}{0.0}) & \textcolor{blue}{0.0} \\
& \textbf{FT} & 44.5 (\textcolor{blue}{44.5}) & 82.0 (\textcolor{blue}{6.6}) & 46.9 (\textcolor{blue}{46.9}) & 77.0 (\textcolor{blue}{2.8}) & \textcolor{blue}{25.2} \\
& \textbf{GA} & 0.1 (\textcolor{blue}{0.1}) & 3.6 (\textcolor{blue}{71.8}) & 0.1 (\textcolor{blue}{0.1}) & 3.2 (\textcolor{blue}{71.0}) & \textcolor{blue}{35.8} \\
& \textbf{RL} & 17.1 (\textcolor{blue}{17.1}) & 42.1 (\textcolor{blue}{33.3}) & 13.2 (\textcolor{blue}{13.2}) & 36.2 (\textcolor{blue}{38.0}) & \textcolor{blue}{25.4} \\
& \textbf{PL} & 0.9 (\textcolor{blue}{0.9}) & 78.2 (\textcolor{blue}{2.8}) & 0.9 (\textcolor{blue}{0.9}) & 75.4 (\textcolor{blue}{1.2}) & \textbf{\textcolor{blue}{1.5}} \\
& \textbf{SalUn} & 3.3 (\textcolor{blue}{3.3}) & 14.7 (\textcolor{blue}{60.7}) & 2.5 (\textcolor{blue}{2.5}) & 12.2 (\textcolor{blue}{62.0}) & \textcolor{blue}{32.1} \\
& \textbf{DUCK} & 5.7 (\textcolor{blue}{5.7}) & 74.8 (\textcolor{blue}{0.6}) & 7.0 (\textcolor{blue}{7.0}) & 73.7 (\textcolor{blue}{0.5}) & \textcolor{blue}{3.5} \\
& \textbf{CU} & 3.4 (\textcolor{blue}{3.4}) & 75.5 (\textcolor{blue}{0.1}) & 4.6 (\textcolor{blue}{4.6}) & 73.5 (\textcolor{blue}{0.7}) & \textcolor{blue}{2.2} \\
& \textbf{SCAR} & 35.5 (\textcolor{blue}{35.5}) & 75.9 (\textcolor{blue}{0.5}) & 37.1 (\textcolor{blue}{37.1}) & 74.4 (\textcolor{blue}{0.2}) & \textcolor{blue}{18.3} \\
& \textbf{SCRUB} & 0.0 (\textcolor{blue}{0.0}) & 3.9 (\textcolor{blue}{71.5}) & 0.0 (\textcolor{blue}{0.0}) & 5.0 (\textcolor{blue}{69.2}) & \textcolor{blue}{35.2} \\
\midrule
\multirow{10}{*}{\textbf{300}}
& \textbf{Retrained} & 0.0 (\textcolor{blue}{0.0}) & 76.9 (\textcolor{blue}{0.0}) & 0.0 (\textcolor{blue}{0.0}) & 75.1 (\textcolor{blue}{0.0}) & \textcolor{blue}{0.0} \\
& \textbf{FT} & 45.5 (\textcolor{blue}{45.5}) & 82.7 (\textcolor{blue}{5.8}) & 46.7 (\textcolor{blue}{46.7}) & 77.7 (\textcolor{blue}{2.6}) & \textcolor{blue}{25.2} \\
& \textbf{GA} & 0.2 (\textcolor{blue}{0.2}) & 2.0 (\textcolor{blue}{74.9}) & 0.2 (\textcolor{blue}{0.2}) & 1.8 (\textcolor{blue}{73.3}) & \textcolor{blue}{37.2} \\
& \textbf{RL} & 11.9 (\textcolor{blue}{11.9}) & 24.1 (\textcolor{blue}{52.7}) & 9.2 (\textcolor{blue}{9.2}) & 20.7 (\textcolor{blue}{54.4}) & \textcolor{blue}{32.0} \\
& \textbf{PL} & 1.0 (\textcolor{blue}{1.0}) & 78.6 (\textcolor{blue}{1.7}) & 0.9 (\textcolor{blue}{0.9}) & 76.0 (\textcolor{blue}{0.9}) & \textbf{\textcolor{blue}{1.1}} \\
& \textbf{SalUn} & 7.3 (\textcolor{blue}{7.3}) & 17.0 (\textcolor{blue}{59.9}) & 6.0 (\textcolor{blue}{6.0}) & 15.3 (\textcolor{blue}{59.8}) & \textcolor{blue}{33.3} \\
& \textbf{DUCK} & 9.3 (\textcolor{blue}{9.3}) & 76.5 (\textcolor{blue}{0.4}) & 11.5 (\textcolor{blue}{11.5}) & 74.6 (\textcolor{blue}{0.5}) & \textcolor{blue}{5.4} \\
& \textbf{CU} & 6.7 (\textcolor{blue}{6.7}) & 77.9 (\textcolor{blue}{1.0}) & 8.8 (\textcolor{blue}{8.8}) & 74.9 (\textcolor{blue}{0.2}) & \textcolor{blue}{4.2} \\
& \textbf{SCAR} & 40.5 (\textcolor{blue}{40.5}) & 79.6 (\textcolor{blue}{2.7}) & 41.1 (\textcolor{blue}{41.1}) & 76.9 (\textcolor{blue}{1.8}) & \textcolor{blue}{21.4} \\
& \textbf{SCRUB} & 0.2 (\textcolor{blue}{0.2}) & 3.9 (\textcolor{blue}{73.0}) & 0.2 (\textcolor{blue}{0.2}) & 4.5 (\textcolor{blue}{70.6}) & \textcolor{blue}{36.0} \\
\bottomrule
\end{tabular}
}
\caption{CUB Top Class-wise Forgetting results. The \textcolor{blue}{blue numbers} indicate the gap. The \textbf{bold numbers} represent optimal values.}
\label{tab:cub}
\end{table}

\begin{table}[ht]
\centering
\resizebox{\columnwidth}{!}{
\begin{tabular}{llccccc}
\toprule
\multicolumn{7}{c}{\textbf{DomainNet Top}} \\
\cmidrule(lr){3-7}
\textbf{Classes} & \textbf{Method} & \textbf{FA} & \textbf{RA} & \textbf{TFA} & \textbf{TRA} & \textbf{Avg. Gap} \\
\midrule
\multirow{10}{*}{\textbf{100}}
& \textbf{Retrained} & 0.0 (\textcolor{blue}{0.0}) & 76.4 (\textcolor{blue}{0.0}) & 0.0 (\textcolor{blue}{0.0}) & 75.9 (\textcolor{blue}{0.0}) & \textcolor{blue}{0.0} \\
& \textbf{FT} & 10.8 (\textcolor{blue}{10.8}) & 79.6 (\textcolor{blue}{3.2}) & 11.7 (\textcolor{blue}{11.7}) & 76.3 (\textcolor{blue}{0.4}) & \textcolor{blue}{6.5} \\
& \textbf{GA} & 0.0 (\textcolor{blue}{0.0}) & 0.4 (\textcolor{blue}{76.0}) & 0.0 (\textcolor{blue}{0.0}) & 0.5 (\textcolor{blue}{75.4}) & \textcolor{blue}{37.7} \\
& \textbf{RL} & 7.3 (\textcolor{blue}{7.3}) & 18.2 (\textcolor{blue}{58.2}) & 6.2 (\textcolor{blue}{6.2}) & 16.6 (\textcolor{blue}{59.3}) & \textcolor{blue}{32.7} \\
& \textbf{PL} & 1.1 (\textcolor{blue}{1.1}) & 79.5 (\textcolor{blue}{3.1}) & 1.0 (\textcolor{blue}{1.0}) & 77.0 (\textcolor{blue}{1.1}) & \textcolor{blue}{1.6} \\
& \textbf{SalUn} & 7.4 (\textcolor{blue}{7.4}) & 17.3 (\textcolor{blue}{59.1}) & 6.2 (\textcolor{blue}{6.2}) & 16.5 (\textcolor{blue}{59.4}) & \textcolor{blue}{33.0} \\
& \textbf{DUCK} & 1.7 (\textcolor{blue}{1.7}) & 76.8 (\textcolor{blue}{0.4}) & 2.1 (\textcolor{blue}{2.1}) & 76.0 (\textcolor{blue}{0.1}) & \textbf{\textcolor{blue}{1.1}} \\
& \textbf{CU} & 2.0 (\textcolor{blue}{2.0}) & 73.3 (\textcolor{blue}{3.1}) & 2.9 (\textcolor{blue}{2.9}) & 73.0 (\textcolor{blue}{2.9}) & \textcolor{blue}{2.7} \\
& \textbf{SCAR} & 13.0 (\textcolor{blue}{13.0}) & 77.1 (\textcolor{blue}{0.7}) & 13.4 (\textcolor{blue}{13.4}) & 76.0 (\textcolor{blue}{0.1}) & \textcolor{blue}{6.8} \\
& \textbf{SCRUB} & 0.0 (\textcolor{blue}{0.0}) & 2.0 (\textcolor{blue}{74.4}) & 0.0 (\textcolor{blue}{0.0}) & 2.8 (\textcolor{blue}{73.1}) & \textcolor{blue}{36.9} \\
\midrule
\multirow{10}{*}{\textbf{200}}
& \textbf{Retrained} & 0.0 (\textcolor{blue}{0.0}) & 78.5 (\textcolor{blue}{0.0}) & 0.0 (\textcolor{blue}{0.0}) & 77.7 (\textcolor{blue}{0.0}) & \textcolor{blue}{0.0} \\
& \textbf{FT} & 17.3 (\textcolor{blue}{17.3}) & 84.1 (\textcolor{blue}{5.6}) & 17.9 (\textcolor{blue}{17.9}) & 80.4 (\textcolor{blue}{2.7}) & \textcolor{blue}{10.9} \\
& \textbf{GA} & 0.3 (\textcolor{blue}{0.3}) & 0.5 (\textcolor{blue}{78.0}) & 0.2 (\textcolor{blue}{0.2}) & 0.7 (\textcolor{blue}{77.0}) & \textcolor{blue}{38.9} \\
& \textbf{RL} & 19.1 (\textcolor{blue}{19.1}) & 29.8 (\textcolor{blue}{48.7}) & 17.9 (\textcolor{blue}{17.9}) & 27.8 (\textcolor{blue}{49.9}) & \textcolor{blue}{33.9} \\
& \textbf{PL} & 1.2 (\textcolor{blue}{1.2}) & 80.6 (\textcolor{blue}{2.1}) & 1.0 (\textcolor{blue}{1.0}) & 78.9 (\textcolor{blue}{1.2}) & \textcolor{blue}{1.4} \\
& \textbf{SalUn} & 9.5 (\textcolor{blue}{9.5}) & 14.0 (\textcolor{blue}{64.5}) & 9.1 (\textcolor{blue}{9.1}) & 12.8 (\textcolor{blue}{64.9}) & \textcolor{blue}{37.0} \\
& \textbf{DUCK} & 0.7 (\textcolor{blue}{0.7}) & 78.4 (\textcolor{blue}{0.1}) & 1.0 (\textcolor{blue}{1.0}) & 77.8 (\textcolor{blue}{0.1}) & \textbf{\textcolor{blue}{0.5}} \\
& \textbf{CU} & 1.4 (\textcolor{blue}{1.4}) & 75.7 (\textcolor{blue}{2.8}) & 1.9 (\textcolor{blue}{1.9}) & 75.4 (\textcolor{blue}{2.3}) & \textcolor{blue}{2.1} \\
& \textbf{SCAR} & 11.3 (\textcolor{blue}{11.3}) & 78.8 (\textcolor{blue}{0.3}) & 12.5 (\textcolor{blue}{12.5}) & 78.1 (\textcolor{blue}{0.4}) & \textcolor{blue}{6.1} \\
& \textbf{SCRUB} & 0.2 (\textcolor{blue}{0.2}) & 1.2 (\textcolor{blue}{77.3}) & 0.3 (\textcolor{blue}{0.3}) & 1.8 (\textcolor{blue}{75.9}) & \textcolor{blue}{38.4} \\
\midrule
\multirow{10}{*}{\textbf{300}}
& \textbf{Retrained} & 0.0 (\textcolor{blue}{0.0}) & 80.3 (\textcolor{blue}{0.0}) & 0.0 (\textcolor{blue}{0.0}) & 79.7 (\textcolor{blue}{0.0}) & \textcolor{blue}{0.0} \\
& \textbf{FT} & 21.7 (\textcolor{blue}{21.7}) & 85.4 (\textcolor{blue}{5.1}) & 22.3 (\textcolor{blue}{22.3}) & 82.3 (\textcolor{blue}{2.6}) & \textcolor{blue}{12.9} \\
& \textbf{GA} & 0.2 (\textcolor{blue}{0.2}) & 0.3 (\textcolor{blue}{80.0}) & 0.2 (\textcolor{blue}{0.2}) & 0.4 (\textcolor{blue}{79.3}) & \textcolor{blue}{39.9} \\
& \textbf{RL} & 14.4 (\textcolor{blue}{14.4}) & 19.4 (\textcolor{blue}{60.9}) & 13.3 (\textcolor{blue}{13.3}) & 18.0 (\textcolor{blue}{61.7}) & \textcolor{blue}{37.6} \\
& \textbf{PL} & 1.0 (\textcolor{blue}{1.0}) & 81.9 (\textcolor{blue}{1.6}) & 0.8 (\textcolor{blue}{0.8}) & 80.9 (\textcolor{blue}{1.2}) & \textcolor{blue}{1.2} \\
& \textbf{SalUn} & 10.5 (\textcolor{blue}{10.5}) & 14.5 (\textcolor{blue}{65.8}) & 9.5 (\textcolor{blue}{9.5}) & 13.4 (\textcolor{blue}{66.3}) & \textcolor{blue}{38.0} \\
& \textbf{DUCK} & 1.7 (\textcolor{blue}{1.7}) & 80.4 (\textcolor{blue}{0.1}) & 1.9 (\textcolor{blue}{1.9}) & 80.3 (\textcolor{blue}{0.6}) & \textbf{\textcolor{blue}{1.1}} \\
& \textbf{CU} & 1.0 (\textcolor{blue}{1.0}) & 82.0 (\textcolor{blue}{1.7}) & 1.2 (\textcolor{blue}{1.2}) & 80.7 (\textcolor{blue}{1.0}) & \textcolor{blue}{1.2} \\
& \textbf{SCAR} & 15.1 (\textcolor{blue}{15.1}) & 82.7 (\textcolor{blue}{2.4}) & 15.6 (\textcolor{blue}{15.6}) & 81.8 (\textcolor{blue}{2.1}) & \textcolor{blue}{8.8} \\
& \textbf{SCRUB} & 1.1 (\textcolor{blue}{1.1}) & 12.9 (\textcolor{blue}{67.4}) & 1.1 (\textcolor{blue}{1.1}) & 15.6 (\textcolor{blue}{64.1}) & \textcolor{blue}{33.4} \\
\bottomrule
\end{tabular}
}
\caption{DomainNet Top Class-wise Forgetting results. The \textcolor{blue}{blue numbers} indicate the gap. The \textbf{bold numbers} represent optimal values.}
\label{tab:domainnet}
\end{table}

\begin{table}[ht]
\centering
\resizebox{\textwidth}{!}{
\begin{tabular}{lccccccc}
\toprule
\multicolumn{2}{c}{}
& \multicolumn{3}{c}{\textbf{Random}} 
& \multicolumn{1}{c}{\textbf{Office-Home Top}} 
& \multicolumn{1}{c}{\textbf{CUB Top}} 
& \multicolumn{1}{c}{\textbf{DomainNet Top}} 
\\
\cmidrule(lr){3-5}
\cmidrule(lr){6-6}
\cmidrule(lr){7-7}
\cmidrule(lr){8-8}
\textbf{Classes}
& \textbf{Method}
& \textbf{Office-Home $k$-NN} & \textbf{CUB $k$-NN} & \textbf{DomainNet $k$-NN} 
& \textbf{Office-Home $k$-NN} 
& \textbf{CUB $k$-NN}
& \textbf{DomainNet $k$-NN}
\\
\midrule
\multirow{11}{*}{\textbf{100}} 

& \textbf{Original}
& 80.3 (\textcolor{blue}{3.0})
& 43.4 (\textcolor{blue}{6.5})
& 84.0 (\textcolor{blue}{2.0}) 
& 80.3 (\textcolor{blue}{14.2})
& 43.4 (\textcolor{blue}{19.5})
& 84.0 (\textcolor{blue}{3.4}) \\

& \textbf{Retrained}
& 77.3 (\textcolor{blue}{0.0})
& 36.9 (\textcolor{blue}{0.0})
& 82.0 (\textcolor{blue}{0.0}) 
& 66.1 (\textcolor{blue}{0.0})
& 23.9 (\textcolor{blue}{0.0})
& 80.6 (\textcolor{blue}{0.0}) \\

& \textbf{FT}
& 78.4 (\textcolor{blue}{1.1})
& \textbf{39.0 (\textcolor{blue}{2.1})}
& \textbf{81.6 (\textcolor{blue}{0.4})} 
& \textbf{73.1 (\textcolor{blue}{7.0})}
& 34.5 (\textcolor{blue}{10.6})
& \textbf{81.0 (\textcolor{blue}{0.4})} \\

& \textbf{GA}
& 29.7 (\textcolor{blue}{47.6})
& 10.4 (\textcolor{blue}{26.5})
& 42.2 (\textcolor{blue}{39.8}) 
& 12.7 (\textcolor{blue}{53.4})
& 3.0 (\textcolor{blue}{20.9})
& 20.5 (\textcolor{blue}{60.1}) \\

& \textbf{RL}
& 44.0 (\textcolor{blue}{33.3})
& 8.0 (\textcolor{blue}{28.9})
& 35.2 (\textcolor{blue}{46.8}) 
& 47.8 (\textcolor{blue}{34.2})
& 4.6 (\textcolor{blue}{19.3})
& 40.5 (\textcolor{blue}{40.1}) \\

& \textbf{PL}
& 80.3 (\textcolor{blue}{3.0})
& 43.3 (\textcolor{blue}{6.4})
& 84.0 (\textcolor{blue}{2.0}) 
& 76.2 (\textcolor{blue}{10.1})
& 37.3 (\textcolor{blue}{13.4})
& 83.3 (\textcolor{blue}{2.7}) \\

& \textbf{SalUn}
& 38.5 (\textcolor{blue}{38.8})
& 8.0 (\textcolor{blue}{28.9})
& 42.5 (\textcolor{blue}{39.5}) 
& 25.5 (\textcolor{blue}{40.6})
& 3.7 (\textcolor{blue}{20.2})
& 40.5 (\textcolor{blue}{40.1}) \\

& \textbf{DUCK}
& 79.8 (\textcolor{blue}{2.5})
& \textbf{39.0 (\textcolor{blue}{2.1})}
& 82.7 (\textcolor{blue}{0.7}) 
& 73.6 (\textcolor{blue}{7.5})
& 33.6 (\textcolor{blue}{9.7})
& 82.3 (\textcolor{blue}{1.7}) \\

& \textbf{CU}
& 75.8 (\textcolor{blue}{1.5})
& 33.9 (\textcolor{blue}{3.0})
& 80.1 (\textcolor{blue}{1.9}) 
& 53.3 (\textcolor{blue}{12.8})
& \textbf{13.7 (\textcolor{blue}{9.2})}
& 77.6 (\textcolor{blue}{3.0}) \\

& \textbf{SCAR}
& \textbf{78.0 (\textcolor{blue}{0.7})}
& 42.8 (\textcolor{blue}{5.9})
& 83.1 (\textcolor{blue}{1.1}) 
& 78.1 (\textcolor{blue}{12.0})
& 38.5 (\textcolor{blue}{14.6})
& 82.9 (\textcolor{blue}{2.3}) \\

& \textbf{SCRUB}
& 74.7 (\textcolor{blue}{2.6})
& 42.2 (\textcolor{blue}{5.3})
& 80.9 (\textcolor{blue}{1.1}) 
& 8.9 (\textcolor{blue}{57.2})
& 5.3 (\textcolor{blue}{18.6})
& 40.8 (\textcolor{blue}{39.8}) \\

\midrule
\multirow{10}{*}{\textbf{200}} 

& \textbf{Retrained}
& 77.9 (\textcolor{blue}{0.0})
& 34.9 (\textcolor{blue}{0.0})
& 81.2 (\textcolor{blue}{0.0}) 
& 63.3 (\textcolor{blue}{0.0})
& 19.1 (\textcolor{blue}{0.0})
& 78.2 (\textcolor{blue}{0.0}) \\

& \textbf{FT}
& 77.0 (\textcolor{blue}{0.9})
& 37.2 (\textcolor{blue}{2.3})
& \textbf{81.7 (\textcolor{blue}{0.5})} 
& 77.9 (\textcolor{blue}{14.6})
& \textbf{16.3 (\textcolor{blue}{2.8})}
& 84.0 (\textcolor{blue}{5.8}) \\

& \textbf{GA}
& 13.0 (\textcolor{blue}{64.9})
& 2.4 (\textcolor{blue}{32.5})
& 20.0 (\textcolor{blue}{61.2}) 
& 14.7 (\textcolor{blue}{48.6})
& 3.3 (\textcolor{blue}{15.8})
& 19.6 (\textcolor{blue}{58.6}) \\

& \textbf{RL}
& 40.5 (\textcolor{blue}{37.4})
& 8.7 (\textcolor{blue}{26.2})
& 52.9 (\textcolor{blue}{28.3}) 
& 31.1 (\textcolor{blue}{32.2})
& 5.4 (\textcolor{blue}{13.7})
& 46.9 (\textcolor{blue}{31.3}) \\

& \textbf{PL}
& 79.4 (\textcolor{blue}{1.5})
& 41.4 (\textcolor{blue}{6.5})
& 83.7 (\textcolor{blue}{2.5}) 
& 76.4 (\textcolor{blue}{13.1})
& 36.0 (\textcolor{blue}{16.9})
& 83.4 (\textcolor{blue}{5.2}) \\

& \textbf{SalUn}
& 28.8 (\textcolor{blue}{49.1})
& 5.7 (\textcolor{blue}{29.2})
& 36.5 (\textcolor{blue}{44.7}) 
& 23.6 (\textcolor{blue}{39.7})
& 4.7 (\textcolor{blue}{14.4})
& 37.8 (\textcolor{blue}{40.4}) \\

& \textbf{DUCK}
& \textbf{78.0 (\textcolor{blue}{0.1})}
& 39.6 (\textcolor{blue}{4.7})
& 70.6 (\textcolor{blue}{10.6}) 
& \textbf{71.9 (\textcolor{blue}{8.6})}
& 29.5 (\textcolor{blue}{10.4})
& 81.8 (\textcolor{blue}{3.6}) \\

& \textbf{CU}
& 80.2 (\textcolor{blue}{2.3})
& 43.6 (\textcolor{blue}{8.7})
& 83.2 (\textcolor{blue}{2.1}) 
& 72.7 (\textcolor{blue}{9.4})
& 28.2 (\textcolor{blue}{9.1})
& \textbf{77.1 (\textcolor{blue}{1.1})} \\

& \textbf{SCAR}
& 80.7 (\textcolor{blue}{2.8})
& 41.6 (\textcolor{blue}{6.7})
& 84.0 (\textcolor{blue}{2.8}) 
& 75.3 (\textcolor{blue}{12.0})
& 38.5 (\textcolor{blue}{19.4})
& 83.2 (\textcolor{blue}{5.0}) \\

& \textbf{SCRUB}
& 78.1 (\textcolor{blue}{0.2})
& \textbf{34.0 (\textcolor{blue}{0.9})}
& \textbf{80.7 (\textcolor{blue}{0.5})} 
& 13.8 (\textcolor{blue}{49.5})
& 5.6 (\textcolor{blue}{13.5})
& 31.7 (\textcolor{blue}{46.5}) \\

\midrule
\multirow{10}{*}{\textbf{300}} 

& \textbf{Retrained}
& 75.5 (\textcolor{blue}{0.0})
& 30.3 (\textcolor{blue}{0.0})
& 79.8 (\textcolor{blue}{0.0}) 
& 55.1 (\textcolor{blue}{0.0})
& 14.0 (\textcolor{blue}{0.0})
& 77.1 (\textcolor{blue}{0.0}) \\

& \textbf{FT}
& 77.8 (\textcolor{blue}{2.3})
& \textbf{38.0 (\textcolor{blue}{7.7})}
& \textbf{81.8 (\textcolor{blue}{2.0})} 
& 78.0 (\textcolor{blue}{22.9})
& 35.8 (\textcolor{blue}{21.8})
& 83.8 (\textcolor{blue}{6.7}) \\

& \textbf{GA}
& 15.9 (\textcolor{blue}{59.6})
& 2.6 (\textcolor{blue}{27.7})
& 17.8 (\textcolor{blue}{62.0}) 
& 12.7 (\textcolor{blue}{42.4})
& 2.7 (\textcolor{blue}{11.3})
& 17.4 (\textcolor{blue}{59.7}) \\

& \textbf{RL}
& 31.1 (\textcolor{blue}{44.4})
& 6.6 (\textcolor{blue}{23.7})
& 37.8 (\textcolor{blue}{42.0}) 
& 27.0 (\textcolor{blue}{28.1})
& 6.3 (\textcolor{blue}{7.7})
& 41.3 (\textcolor{blue}{35.8}) \\

& \textbf{PL}
& 80.1 (\textcolor{blue}{4.6})
& 38.7 (\textcolor{blue}{8.4})
& 83.8 (\textcolor{blue}{4.0}) 
& 75.6 (\textcolor{blue}{20.5})
& 35.8 (\textcolor{blue}{21.8})
& 83.2 (\textcolor{blue}{6.1}) \\

& \textbf{SalUn}
& 30.6 (\textcolor{blue}{44.9})
& 7.0 (\textcolor{blue}{23.3})
& 37.8 (\textcolor{blue}{42.0}) 
& 22.0 (\textcolor{blue}{33.1})
& \textbf{6.8 (\textcolor{blue}{7.2})}
& 38.2 (\textcolor{blue}{38.9}) \\

& \textbf{DUCK}
& 79.4 (\textcolor{blue}{3.9})
& 40.2 (\textcolor{blue}{9.9})
& 82.5 (\textcolor{blue}{2.7}) 
& 71.3 (\textcolor{blue}{16.2})
& 32.7 (\textcolor{blue}{18.7})
& \textbf{82.2 (\textcolor{blue}{5.1})} \\

& \textbf{CU}
& \textbf{77.5 (\textcolor{blue}{2.0})}
& 40.0 (\textcolor{blue}{9.7})
& 82.9 (\textcolor{blue}{3.1}) 
& \textbf{70.6 (\textcolor{blue}{15.5})}
& 29.9 (\textcolor{blue}{15.9})
& 82.5 (\textcolor{blue}{5.4}) \\

& \textbf{SCAR}
& 80.5 (\textcolor{blue}{5.0})
& 42.4 (\textcolor{blue}{12.1})
& 83.6 (\textcolor{blue}{3.8}) 
& 77.3 (\textcolor{blue}{22.2})
& 37.5 (\textcolor{blue}{23.5})
& 83.7 (\textcolor{blue}{6.6}) \\

& \textbf{SCRUB}
& 78.7 (\textcolor{blue}{3.2})
& 39.4 (\textcolor{blue}{9.1})
& 82.9 (\textcolor{blue}{3.1}) 
& 13.8 (\textcolor{blue}{41.3})
& 5.6 (\textcolor{blue}{8.4})
& 65.4 (\textcolor{blue}{11.7}) \\

\bottomrule
\end{tabular}
}
\caption{Comparison of $k$-NN performance across different unlearning subsets. Each block (Random, Office-Home Top, CUB Top, DomainNet Top) indicates the subset of classes removed. For each subset, we report $k$-NN Accuracy on three downstream tasks (Office-Home, CUB, and DomainNet-126) to measure transferability. The \textcolor{blue}{blue numbers} indicate the gap relative to the retrained model. The \textbf{bold numbers} represent optimal values.}
\label{tab:knn}
\end{table}

\begin{table}[t]
\centering
\resizebox{\textwidth}{!}{
\begin{tabular}{lcccccccc}
\toprule
\multicolumn{2}{c}{}
& \multicolumn{4}{c}{\textbf{Random}} 
& \multicolumn{1}{c}{\textbf{Office-Home Top}} 
& \multicolumn{1}{c}{\textbf{CUB Top}} 
& \multicolumn{1}{c}{\textbf{DomainNet Top}} 
\\
\cmidrule(lr){3-6}
\cmidrule(lr){7-7}
\cmidrule(lr){8-8}
\cmidrule(lr){9-9}
\textbf{Classes}
& \textbf{Method}
& \textbf{Office-Home CKA} & \textbf{CUB CKA} & \textbf{DomainNet CKA} & \textbf{Avg.}
& \textbf{Office-Home CKA}
& \textbf{CUB CKA}
& \textbf{DomainNet CKA}
\\
\midrule
\multirow{11}{*}{\textbf{100}} 
& \textbf{Original}
& 89.8 (\textcolor{blue}{10.2})
& 82.1 (\textcolor{blue}{17.9})
& 81.8 (\textcolor{blue}{18.2}) 
& \textcolor{blue}{84.6} 
& 83.0 (\textcolor{blue}{17.0})
& 69.7 (\textcolor{blue}{30.3})
& 80.4 (\textcolor{blue}{19.6})\\

& \textbf{Retrained}
& 100.0 (\textcolor{blue}{0.0})
& 100.0 (\textcolor{blue}{0.0})
& 100.0 (\textcolor{blue}{0.0})
& \textcolor{blue}{100.0} 
& 100.0 (\textcolor{blue}{0.0})
& 100.0 (\textcolor{blue}{0.0})
& 100.0 (\textcolor{blue}{0.0})\\

& \textbf{FT}
& 87.6 (\textcolor{blue}{12.4})
& 79.0 (\textcolor{blue}{21.0})
& 79.9 (\textcolor{blue}{20.1})
& \textcolor{blue}{82.2}
& 82.4 (\textcolor{blue}{17.6})
& 66.9 (\textcolor{blue}{33.1})
& 78.9 (\textcolor{blue}{21.1})
\\

& \textbf{GA}
& 8.3 (\textcolor{blue}{91.7})
& 9.8 (\textcolor{blue}{90.2})
& 10.2 (\textcolor{blue}{89.8})
& \textcolor{blue}{9.4}
& 17.5 (\textcolor{blue}{82.5})
& 20.1 (\textcolor{blue}{79.9})
& 20.0 (\textcolor{blue}{80.0}) \\

& \textbf{RL}
& 6.3 (\textcolor{blue}{93.7})
& 5.8 (\textcolor{blue}{94.2})
& 4.7 (\textcolor{blue}{95.3})
& \textcolor{blue}{5.6}
& 24.4 (\textcolor{blue}{75.6})
& 17.8 (\textcolor{blue}{82.2})
& 12.3 (\textcolor{blue}{87.7}) \\

& \textbf{PL}
& \textbf{91.6 (\textcolor{blue}{8.4})}
& \textbf{84.7 (\textcolor{blue}{15.3})}
& 84.5 (\textcolor{blue}{15.5})
& \textbf{\textcolor{blue}{86.9}}
& 87.3 (\textcolor{blue}{12.7})
& 75.0 (\textcolor{blue}{25.0})
& 83.5 (\textcolor{blue}{16.5}) \\

& \textbf{SalUn}
& 9.7 (\textcolor{blue}{90.3})
& 8.0 (\textcolor{blue}{92.0})
& 8.5 (\textcolor{blue}{91.5})
& \textcolor{blue}{8.7}
& 11.6 (\textcolor{blue}{88.4})
& 24.9 (\textcolor{blue}{75.1})
& 12.4 (\textcolor{blue}{87.6})\\

& \textbf{DUCK}
& 90.7 (\textcolor{blue}{9.3})
& 83.2 (\textcolor{blue}{16.8})
& \textbf{84.9 (\textcolor{blue}{15.1})}
& \textcolor{blue}{86.3}
& \textbf{88.2 (\textcolor{blue}{11.8})}
& \textbf{78.2 (\textcolor{blue}{21.8})}
& \textbf{87.7 (\textcolor{blue}{12.3})} \\

& \textbf{CU}
& 85.6 (\textcolor{blue}{14.4})
& 75.9 (\textcolor{blue}{24.1})
& 76.1 (\textcolor{blue}{23.9})
& \textcolor{blue}{79.2}
& 10.1 (\textcolor{blue}{89.9})
& 5.2 (\textcolor{blue}{94.8})
& 72.2 (\textcolor{blue}{27.8}) \\

& \textbf{SCAR}
& 74.2 (\textcolor{blue}{25.8})
& 65.4 (\textcolor{blue}{34.6})
& 58.8 (\textcolor{blue}{41.2})
& \textcolor{blue}{66.1}
& 69.0 (\textcolor{blue}{31.0})
& 74.0 (\textcolor{blue}{26.0})
& 83.6 (\textcolor{blue}{16.4})
\\

& \textbf{SCRUB}
& 68.9 (\textcolor{blue}{31.1})
& 63.8 (\textcolor{blue}{36.2})
& 52.8 (\textcolor{blue}{47.2})
& \textcolor{blue}{61.8}
& 5.7 (\textcolor{blue}{94.3})
& 33.1 (\textcolor{blue}{66.9})
& 29.3 (\textcolor{blue}{70.7})
\\
\midrule


\multirow{11}{*}{\textbf{200}}

& \textbf{Original}
& 88.6 (\textcolor{blue}{11.4})
& 80.5 (\textcolor{blue}{19.5})
& 80.9 (\textcolor{blue}{19.1})
& \textcolor{blue}{83.3}
& 79.7 (\textcolor{blue}{20.3})
& 63.4 (\textcolor{blue}{36.6})
& 78.9 (\textcolor{blue}{21.1})

\\

& \textbf{Retrained}
& 100.0 (\textcolor{blue}{0.0})
& 100.0 (\textcolor{blue}{0.0})
& 100.0 (\textcolor{blue}{0.0})
& \textcolor{blue}{100.0}
& 100.0 (\textcolor{blue}{0.0})
& 100.0 (\textcolor{blue}{0.0})
& 100.0 (\textcolor{blue}{0.0})
\\

& \textbf{FT}
& 87.0 (\textcolor{blue}{13.0})
& 77.6 (\textcolor{blue}{22.4})
& 78.9 (\textcolor{blue}{21.1})
& \textcolor{blue}{81.2}
& 82.4 (\textcolor{blue}{17.6})
& 70.7 (\textcolor{blue}{29.3})
& 82.8 (\textcolor{blue}{17.2})

\\

& \textbf{GA}
& 13.2 (\textcolor{blue}{86.8})
& 9.8 (\textcolor{blue}{90.2})
& 17.3 (\textcolor{blue}{82.7})
& \textcolor{blue}{13.4}
& 16.9 (\textcolor{blue}{83.1})
& 24.6 (\textcolor{blue}{75.4})
& 19.6 (\textcolor{blue}{80.4})
\\

& \textbf{RL}
& 12.9 (\textcolor{blue}{87.1})
& 10.7 (\textcolor{blue}{89.3})
& 12.9 (\textcolor{blue}{87.1})
& \textcolor{blue}{12.2}
& 25.8 (\textcolor{blue}{74.2})
& 22.6 (\textcolor{blue}{77.4})
& 12.4 (\textcolor{blue}{87.6})

\\

& \textbf{PL}
& 91.3 (\textcolor{blue}{8.7})
& 83.1 (\textcolor{blue}{16.9})
& 84.6 (\textcolor{blue}{15.4})
& \textcolor{blue}{86.3}
& 83.7 (\textcolor{blue}{16.3})
& 70.1 (\textcolor{blue}{29.9})
& 82.7 (\textcolor{blue}{17.3})
\\

& \textbf{SalUn}
& 9.8 (\textcolor{blue}{90.2})
& 6.3 (\textcolor{blue}{93.7})
& 11.7 (\textcolor{blue}{88.3})
& \textcolor{blue}{9.3}
& 12.7 (\textcolor{blue}{87.3})
& 5.5 (\textcolor{blue}{94.5})
& 9.8 (\textcolor{blue}{90.2})

\\

& \textbf{DUCK}
& \textbf{92.4 (\textcolor{blue}{7.6})}
& \textbf{84.5 (\textcolor{blue}{15.5})}
& \textbf{87.7 (\textcolor{blue}{12.3})}
& \textbf{\textcolor{blue}{88.2}}
& \textbf{85.8 (\textcolor{blue}{14.2})}
& \textbf{74.5 (\textcolor{blue}{25.5})}
& \textbf{86.7 (\textcolor{blue}{13.3})}
\\

& \textbf{CU}
& 91.3 (\textcolor{blue}{8.7})
& 83.9 (\textcolor{blue}{16.1})
& 84.7 (\textcolor{blue}{15.3})
& \textcolor{blue}{86.6}
& 84.6 (\textcolor{blue}{15.4})
& 63.9 (\textcolor{blue}{36.1})
& 74.0 (\textcolor{blue}{26.0})
\\

& \textbf{SCAR}
& 91.1 (\textcolor{blue}{8.9})
& 83.1 (\textcolor{blue}{16.9})
& 85.0 (\textcolor{blue}{15.0})
& \textcolor{blue}{86.4}
& 83.3 (\textcolor{blue}{16.7})
& 70.6 (\textcolor{blue}{29.4})
& 82.7 (\textcolor{blue}{17.3})
\\

& \textbf{SCRUB}
& 75.9 (\textcolor{blue}{24.1})
& 69.6 (\textcolor{blue}{30.4})
& 48.3 (\textcolor{blue}{51.7})
& \textcolor{blue}{64.6}
& 15.9 (\textcolor{blue}{84.1})
& 41.5 (\textcolor{blue}{58.5})
& 18.6 (\textcolor{blue}{81.4})
\\
\midrule


\multirow{11}{*}{\textbf{300}}
& \textbf{Original}
& 86.8 (\textcolor{blue}{13.2})
& 76.8 (\textcolor{blue}{23.2})
& 80.2 (\textcolor{blue}{19.8})
& \textcolor{blue}{81.3}
& 76.7 (\textcolor{blue}{23.3})
& 59.6 (\textcolor{blue}{40.4})
& 77.3 (\textcolor{blue}{22.7})
\\

& \textbf{Retrained}
& 100.0 (\textcolor{blue}{0.0})
& 100.0 (\textcolor{blue}{0.0})
& 100.0 (\textcolor{blue}{0.0})
& \textcolor{blue}{100.0}
& 100.0 (\textcolor{blue}{0.0})
& 100.0 (\textcolor{blue}{0.0})
& 100.0 (\textcolor{blue}{0.0})
\\

& \textbf{FT}
& 86.3 (\textcolor{blue}{13.7})
& 73.4 (\textcolor{blue}{26.6})
& 78.1 (\textcolor{blue}{21.9})
& \textcolor{blue}{79.3}
& 79.7 (\textcolor{blue}{20.3})
& 67.5 (\textcolor{blue}{32.5})
& 80.5 (\textcolor{blue}{19.5})
\\

& \textbf{GA}
& 12.8 (\textcolor{blue}{87.2})
& 10.5 (\textcolor{blue}{89.5})
& 10.7 (\textcolor{blue}{89.3})
& \textcolor{blue}{11.3}
& 16.1 (\textcolor{blue}{83.9})
& 23.6 (\textcolor{blue}{76.4})
& 21.0 (\textcolor{blue}{79.0})

\\

& \textbf{RL}
& 6.8 (\textcolor{blue}{93.2})
& 6.4 (\textcolor{blue}{93.6})
& 4.6 (\textcolor{blue}{95.4})
& \textcolor{blue}{5.9}
& 13.2 (\textcolor{blue}{86.8})
& 12.5 (\textcolor{blue}{87.5})
& 13.2 (\textcolor{blue}{86.8})
\\

& \textbf{PL}
& 90.5 (\textcolor{blue}{9.5})
& 81.1 (\textcolor{blue}{18.9})
& 83.6 (\textcolor{blue}{16.4})
& \textcolor{blue}{85.1}
& 81.6 (\textcolor{blue}{18.4})
& 66.8 (\textcolor{blue}{33.2})
& 81.2 (\textcolor{blue}{18.8})
\\

& \textbf{SalUn}
& 9.5 (\textcolor{blue}{90.5})
& 6.3 (\textcolor{blue}{93.7})
& 12.0 (\textcolor{blue}{88.0})
& \textcolor{blue}{9.3}
& 13.4 (\textcolor{blue}{86.6})
& 5.3 (\textcolor{blue}{94.7})
& 13.3 (\textcolor{blue}{86.7})
\\

& \textbf{DUCK}
& \textbf{91.7 (\textcolor{blue}{8.3})}
& \textbf{82.0 (\textcolor{blue}{18.0})}
& \textbf{87.3 (\textcolor{blue}{12.7})}
& \textbf{\textcolor{blue}{87.0}}
& \textbf{83.5 (\textcolor{blue}{16.5})}
& \textbf{72.2 (\textcolor{blue}{27.8})}
& \textbf{85.3 (\textcolor{blue}{14.7})}
\\

& \textbf{CU}
& 90.5 (\textcolor{blue}{9.5})
& 80.2 (\textcolor{blue}{19.8})
& 84.1 (\textcolor{blue}{15.9})
& \textcolor{blue}{84.9}
& 82.4 (\textcolor{blue}{17.6})
& 72.5 (\textcolor{blue}{27.5})
& 81.8 (\textcolor{blue}{18.2})
\\

& \textbf{SCAR}
& 89.6 (\textcolor{blue}{10.4})
& 79.2 (\textcolor{blue}{20.8})
& 84.5 (\textcolor{blue}{15.5})
& \textcolor{blue}{84.4}
& 80.2 (\textcolor{blue}{19.8})
& 65.5 (\textcolor{blue}{34.5})
& 82.0 (\textcolor{blue}{18.0})
\\

& \textbf{SCRUB}
& 86.6 (\textcolor{blue}{13.4})
& 76.6 (\textcolor{blue}{23.4})
& 78.9 (\textcolor{blue}{21.1})
& \textcolor{blue}{80.7}
& 24.6 (\textcolor{blue}{75.4})
& 43.2 (\textcolor{blue}{56.8})
& 33.5 (\textcolor{blue}{66.5})
\\
\bottomrule
\end{tabular}
}
\caption{Comparison of CKA values between \thetau and \thetar across different unlearning subsets. Each block (Random, Office-Home Top, CUB Top, DomainNet Top) indicates the subset of classes removed. For each subset, we report CKA value on three downstream tasks (Office-Home, CUB, and DomainNet-126) to gauge representational quality. The \textbf{bold numbers} represent optimal values.}
\label{tab:cka}
\end{table}

\begin{table}[ht]
\centering
\resizebox{\textwidth}{!}{
\begin{tabular}{lccccccccccccc}
\toprule
\multicolumn{2}{c}{}
& \multicolumn{3}{c}{\textbf{Random}} 
& \multicolumn{3}{c}{\textbf{Office-Home Top}} 
& \multicolumn{3}{c}{\textbf{CUB Top}} 
& \multicolumn{3}{c}{\textbf{DomainNet Top}} 
\\
\cmidrule(lr){3-5}
\cmidrule(lr){6-8}
\cmidrule(lr){9-11}
\cmidrule(lr){12-14}
\textbf{Classes}
& \textbf{Method}
& \textbf{AGL} & \textbf{AGR} & \textbf{H-LR} 
& \textbf{AGL} & \textbf{AGR} & \textbf{H-LR} 
& \textbf{AGL} & \textbf{AGR} & \textbf{H-LR} 
& \textbf{AGL} & \textbf{AGR} & \textbf{H-LR} 
\\
\midrule
\multirow{10}{*}{\textbf{100}} 


& \textbf{Retrained}
& 1.00
& 1.00
& 1.00

& 1.00
& 1.00
& 1.00

& 1.00
& 1.00
& 1.00

& 1.00
& 1.00
& 1.00  \\

& \textbf{FT}
& 0.78
& 0.81
& 0.79

& 0.82
& 0.77
& 0.79

& 0.52
& 0.60
& 0.56

& 0.76
& 0.79
& 0.77  \\

& \textbf{GA}
& 0.12
& 0.06
& 0.08

& 0.06
& 0.08
& 0.07

& 0.08
& 0.16
& 0.11

& 0.06
& 0.08
& 0.07  \\

& \textbf{RL}
& 0.12
& 0.04
& 0.06

& 0.08
& 0.16
& 0.11

& 0.33
& 0.14
& 0.20

& 0.15
& 0.07
& 0.10   \\

& \textbf{PL}
& 0.94
& 0.84
& 0.89

& 0.92
& 0.78
& 0.84

& \textbf{0.94}
& 0.65
& \textbf{0.77}

& 0.94
& 0.81
& 0.87   \\

& \textbf{SalUn}
& 0.15
& 0.06
& 0.09

& 0.08
& 0.07
& 0.07

& 0.26
& 0.20
& 0.23

& 0.14
& 0.07
& 0.09  \\

& \textbf{DUCK}
& \textbf{0.96}
& 0.85
& \textbf{0.90}

& \textbf{0.93}
& \textbf{0.82}
& \textbf{0.87}

& 0.85
& \textbf{0.71}
& \textbf{0.77}

& \textbf{0.96}
& \textbf{0.86}
& \textbf{0.91}  \\

& \textbf{CU}
& 0.92
& 0.78
& 0.84

& 0.69
& 0.09
& 0.16

& 0.49
& 0.05
& 0.09

& 0.90
& 0.70
& 0.79  \\

& \textbf{SCAR}
& 0.87
& \textbf{0.94}
& 0.74

& 0.83
& 0.61
& 0.70

& 0.47
& 0.63
& 0.54

& 0.75
& 0.82
& 0.78  \\

& \textbf{SCRUB}
& 0.80
& 0.60
& 0.69

& 0.06
& 0.02
& 0.03

& 0.09
& 0.27
& 0.14

& 0.07
& 0.18
& 0.10  \\
\bottomrule
\end{tabular}
}
\caption{Comparison of the proposed metrics across different unlearning subsets. Each block (Random, Office-Home Top, CUB Top, DomainNet Top) indicates the subset of classes removed. The evaluated metrics include AGL, AGR, and the unified metric H-LR. Higher AGL, AGR, and H-LR scores indicate better preservation of feature representations and overall unlearning effectiveness. The \textbf{bold numbers} represent optimal values.}
\label{tab:all_h-lr}
\end{table}






\clearpage
\bibliographystyle{elsarticle-harv}
\bibliography{elsarticle}

\end{document}